\documentclass{article}

\usepackage{microtype}
\usepackage{graphicx}
\usepackage{subcaption}
\usepackage{booktabs} 
\usepackage{tikz}
\usetikzlibrary{arrows.meta,calc}
\usepackage{pgfplots}
\pgfplotsset{compat=1.18}

\usepackage{hyperref}

\usepackage[preprint]{icml2026}

\usepackage{amsmath}
\usepackage{amssymb}
\usepackage{mathtools}
\usepackage{amsthm}

\usepackage[capitalize,noabbrev]{cleveref}

\theoremstyle{plain}
\newtheorem{theorem}{Theorem}[section]
\newtheorem{proposition}[theorem]{Proposition}
\newtheorem{lemma}[theorem]{Lemma}
\newtheorem{corollary}[theorem]{Corollary}
\theoremstyle{definition}

\theoremstyle{remark}

\usepackage[textsize=tiny]{todonotes}

\icmltitlerunning{The Information Limits of Neural CDEs Under Sparse Observations}

\begin{document}

\twocolumn[
\icmltitle{The Information Limits of Neural Controlled Differential Equations\\Under Sparse Observations}



\icmlsetsymbol{equal}{*}

\begin{icmlauthorlist}
\icmlauthor{Rostislav Gusev}{yyy}
\icmlauthor{Evgenia Romanenkova}{yyy}
\icmlauthor{Alexey Zaytsev}{yyy}
\end{icmlauthorlist}

\icmlaffiliation{yyy}{Applied AI Institute}

\icmlcorrespondingauthor{Rostislav Gusev}{gavkav05@gmail.com}

\icmlkeywords{Machine Learning, ICML}

\vskip 0.3in
]



\printAffiliationsAndNotice{}  

\begin{abstract}
Neural controlled differential equations (Neural CDEs) evolve hidden states along continuous paths constructed by interpolating discrete observations. Sparse measurements leave the input between observation times undetermined, and admissible continuous paths consistent with the same data can produce different terminal states. We characterize this ambiguity for planar drivers with bounded speed. The set of compatible terminal states admits a leading-order zonotopic approximation determined by feasible signed areas and transported Lie brackets. Its projection onto a scalar prediction yields a computable, target-specific ambiguity certificate. We also study the number of measurements needed to reduce the remaining uncertainty. For nondegenerate planar CDEs, the optimal worst-case radius over deterministic adaptive sampling policies scales as $N^{-1}$ when speed is bounded and $N^{-2}$ when both speed and acceleration are bounded, where $N$ is the number of observation intervals. Synthetic experiments assess the certificate using admissible continuous completions and examine the role of path regularity.
\end{abstract}

\section{Introduction}
\label{sec:intro}

Neural controlled differential equations (Neural CDEs) process irregularly sampled time series by evolving hidden states along continuous control paths constructed from discrete observations \citep{kidger2020ncde,morrill2022interpolation}. Interpolation specifies the control between sampled times. When the observations come from an underlying continuous signal, its evolution between measurements remains unknown. Different continuous paths can agree with every recorded value and still drive the same CDE to different terminal states.

These differences arise from path geometry that point observations do not capture. The observed endpoints determine the increments of a planar control, while the signed areas traced between them remain undetermined. In a CDE with noncommuting vector fields, these areas influence the hidden state, and their effects propagate through the subsequent dynamics. The resulting variation in the terminal response depends on the geometry permitted by the observations and the sensitivity of the CDE to that geometry.

The size of this ambiguity matters when a Neural CDE is used to model a continuously driven system. Observation schedules with the same number of measurements can constrain a prediction to different degrees, depending on where the missing path information affects the dynamics. An interpolated control provides a prediction, while the range of responses across compatible continuous paths describes how much of that prediction is determined by the available measurements. Quantifying this range can help assess the adequacy of a sampling schedule and identify intervals where additional observations would be informative.

Previous analyses of Neural CDEs have examined interpolation schemes and sampling-dependent approximation and generalization \citep{morrill2022interpolation,bleistein2024generalization}. Higher-order signatures and Lie brackets have also been used to understand and approximate CDE dynamics \citep{walker2024logncde}. We study how the information missing from a discrete observation record affects the responses of a fixed CDE. The admissible inputs are continuous paths satisfying the observed values and prescribed regularity constraints. Their terminal responses form the identified set associated with the observation record.

Our main results are summarized below.

\begin{itemize}
    \item \textbf{Geometry of partial identification.}
    We characterize the leading terminal-state identified set for planar CDEs with bounded-speed inputs. Feasible signed areas and transported Lie brackets determine a zonotopic approximation with a uniform Hausdorff error bound.

    \item \textbf{Target-specific ambiguity certificate.}
    We derive a computable certificate for scalar predictions that combines the unresolved path geometry on each observation interval with the sensitivity of the terminal output to that geometry.

    \item \textbf{Minimax limits of observation.}
    We establish minimax rates of $N^{-1}$ under bounded speed and $N^{-2}$ under bounded speed and acceleration for planar CDEs with nonvanishing transported Lie-bracket sensitivity. The bounds hold over deterministic adaptive observation policies, where $N$ denotes the number of observation intervals.

    \item \textbf{Empirical evaluation.}
    We evaluate the certificate on trained planar Neural CDEs using admissible continuous completions of sparse observations. The experiments examine certificate accuracy, the effect of path regularity, and the selection of additional measurements.
\end{itemize}

\section{Related Work}
\label{sec:related_work}

\paragraph{Neural CDEs and sparse observations.}
\citet{kidger2020ncde} introduced Neural CDEs for irregularly sampled and partially observed time series, using continuous control paths constructed from discrete measurements. The choice of interpolation affects the information available to the model, particularly in online prediction \citep{morrill2022interpolation}. More recent work develops injective embeddings of irregular data without reconstructing the observed variables \citep{walker2026faithful}. Theoretical analyses by \citet{bleistein2024generalization} establish sampling-dependent generalization bounds and characterize sampling-induced approximation bias using continuity properties of CDE solutions. These results describe how sampling and continuous representations affect Neural CDE predictions.

\begin{figure*}[t]
    \centering
    \includegraphics[width=\textwidth]{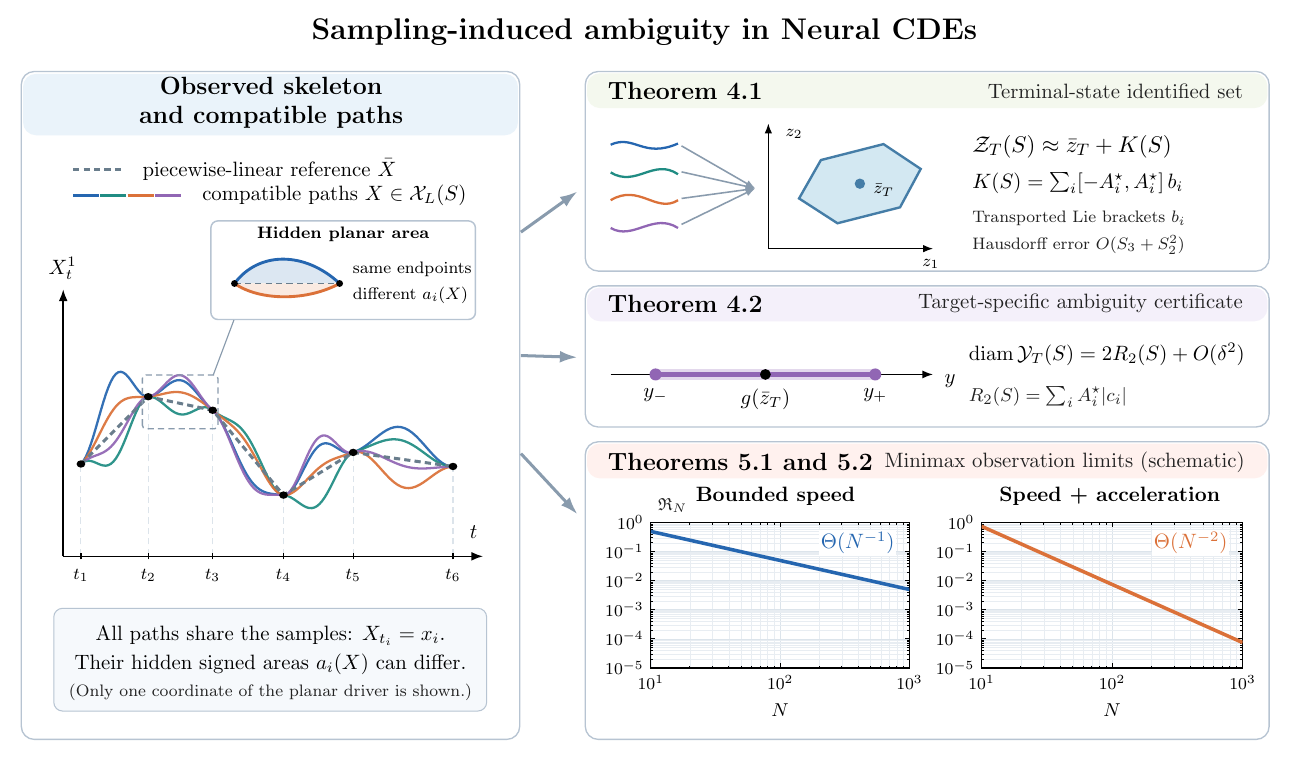}

    \caption{\textbf{Sampling-induced ambiguity in Neural CDEs.}
A discrete observation record $S=\{(t_i,x_i)\}$ admits multiple bounded-speed continuous paths $X\in\mathcal X_L(S)$ (colored curves). The dashed line represents the piecewise-linear reference $\bar X$. Only the first coordinate of the planar input is shown. Paths sharing the same observations may have different signed areas $a_i(X)$ between measurements. Theorem~\ref{thm:partial_state_zonotope} approximates the terminal-state identified set by $\bar z_T+K(S)$, where $\bar z_T=Z_T(\bar X)$ and $K(S)$ is the zonotope generated by feasible signed areas and transported Lie brackets, defined in~\eqref{eq:partial_state_zonotope}. The Hausdorff approximation error is $O(S_3+S_2^2)$. Theorem~\ref{thm:partial_target_certificate} gives a leading-order approximation to the scalar response identified set by the interval with endpoints $y_\pm=g(\bar z_T)\pm R_2(S)$ ($\delta=\max_i(t_{i+1}-t_i)$), where the target-specific radius $R_2(S)$ is defined in~\eqref{eq:partial_target_radius}. Theorems~\ref{thm:minimax_speed} and~\ref{thm:minimax_acceleration} establish minimax information-radius rates $\Theta(N^{-1})$ and $\Theta(N^{-2})$ under bounded speed and bounded speed plus acceleration, respectively. Here $N$ is the number of final observation intervals, and the minimax radius $\mathfrak R_N$ is defined in~\eqref{eq:minimax_information_radius}. The rates hold under the nondegeneracy assumptions of the respective theorems. The plotted curves are schematic.}
    \label{fig:overview}
\end{figure*}

\paragraph{Path geometry and uncertain dynamics.}
Higher-order path information influences controlled dynamics through iterated integrals and Lie brackets. Log-NCDEs exploit this structure to approximate CDE solutions using higher-order log-signatures \citep{walker2024logncde}. In stochastic settings, \citet{malham2014levy} study the conditional distribution of Levy area for a Brownian path with prescribed endpoints. Inverse problems for rough differential equations investigate recovery of the driving signal from observed responses \citep{bailleul2015inverse,morrish2026inverse}. Reachability analysis studies the sets of states generated by admissible inputs \citep{althoff2021reachability}. Discrete observations of the driver impose a different constraint: the compatible input paths must agree at the measured times, and their unresolved geometry determines the range of possible CDE responses.

\paragraph{Optimal recovery and information limits.}
Optimal recovery theory establishes worst-case limits on reconstructing functions from finitely many observations \citep{micchelli1976optimal}. Information-based complexity provides a general framework for studying the information required to approximate a functional \citep{traub1988ibc}. Related questions arise for stochastic differential equations. \citet{mullergronbach2004optimal} derive minimal approximation errors from sequential observations of the driving Brownian motion and construct asymptotically optimal adaptive methods. The information radius considered here measures the worst-case variation of a CDE output over observation-compatible drivers satisfying deterministic regularity constraints. Its dependence on transported Lie-bracket sensitivity connects these information limits to the dynamics and the prediction target.

\paragraph{Adaptive computation and observation design.}
Adaptive numerical methods use estimates of local error to allocate computational effort. \citet{moon2003convergence} establish convergence and complexity results for adaptive ODE approximation, while \citet{bayer2023adaptive} develop an adaptive RDE solver based on local error contributions to the global solution error. Active sensing concerns the acquisition of additional measurements. \citet{qin2023ras} develop a Neural-CDE-based sensing policy that selects observation times and measured variables under prediction-risk and acquisition-cost objectives. In the present setting, measurement locations are evaluated through the ambiguity remaining among admissible continuous drivers, linking observation design to the radius of information.

\section{Problem Formulation}
\label{sec:problem}

\paragraph{Controlled dynamics and observations.}
Let $T>0$, let $d,m\ge 1$, and let $X:[0,T]\to\mathbb R^d$ be an unknown continuous input. Fix an initial state $z_0\in\mathbb R^m$ and vector fields $V_j:\mathbb R^m\to\mathbb R^m$, $j=1,\ldots,d$. The input drives the controlled differential equation
\begin{equation}
\label{eq:cde_model}
\mathrm{d}Z_t=\sum_{j=1}^d V_j(Z_t)\,\mathrm{d}X_t^j,
\qquad Z_0=z_0.
\end{equation}
For a given input $X$, the solution is denoted by $Z_t(X)\in\mathbb R^m$. We consider a scalar terminal output $F(X)=g(Z_T(X))$, where $g:\mathbb R^m\to\mathbb R$ is a fixed readout. The regularity assumptions on the vector fields and the readout are specified in the corresponding theorems.

The input is observed at times $0=t_0<\cdots<t_n=T$, giving $S=\{(t_i,x_i)\}_{i=0}^n$, where $x_i=X_{t_i}\in\mathbb R^d$. These observations constrain the values of $X$ at the sampled times, leaving its evolution between measurements unspecified. To characterize the possible continuous paths, we impose regularity bounds on the input.

For $k=1,2$, let $W^{k,\infty}([0,T];\mathbb R^d)$ denote the Sobolev space of paths whose weak derivatives up to order $k$ are essentially bounded. We use their continuous representatives, so the values $X_{t_i}$ are well defined. For prescribed bounds $L,M>0$, define the classes of continuous paths compatible with the observations:
\[
\begin{aligned}
\mathcal X_L(S)
&=\bigl\{X\in W^{1,\infty}([0,T];\mathbb R^d):\\
&\qquad X_{t_i}=x_i,\quad i=0,\ldots,n,\\
&\qquad \|\dot X\|_{L^\infty}\le L\bigr\},\\[2mm]
\mathcal X_{L,M}(S)
&=\bigl\{X\in\mathcal X_L(S)\cap W^{2,\infty}:\\
&\qquad \|\ddot X\|_{L^\infty}\le M\bigr\}.
\end{aligned}
\]
All derivative bounds hold almost everywhere and use the Euclidean norm. The speed-bounded class $\mathcal X_L(S)$ is nonempty exactly when $\|x_{i+1}-x_i\|\le L(t_{i+1}-t_i)$ on every observation interval, since the piecewise-linear interpolation is then admissible. The observations therefore determine the minimum feasible speed bound, but place no finite upper bound on the speed of an unobserved completion. Both $L$ and $M$ are prescribed regularity assumptions. If $\mathcal X_{L,M}(S)=\varnothing$, those bounds are inconsistent with the observed record.

\paragraph{Response identified sets.}
Let $\mathcal C(S)$ denote either $\mathcal X_L(S)$ or $\mathcal X_{L,M}(S)$. The terminal states generated by paths in this class form the identified set
\[
\mathcal Z_T(S;\mathcal C)
=\{Z_T(X):X\in\mathcal C(S)\}
\subseteq\mathbb R^m.
\]
The corresponding scalar outputs form
\begin{equation}
\label{eq:response_identified_set}
\mathcal Y_T(S;\mathcal C)
=\{F(X):X\in\mathcal C(S)\}
\subseteq\mathbb R.
\end{equation}

\paragraph{Adaptive observations and information limits.}
The identified set quantifies the ambiguity remaining after a particular set of measurements. We also consider how small this ambiguity can become when observation times are selected to reduce it.

For the minimax problem, we specialize to planar drivers ($d=2$). For the fixed-initial-state CDE in~\eqref{eq:cde_model}, translating the driver by $-X_0$ leaves the solution unchanged, since only the increments of $X$ enter the equation. We may therefore take $X_0=0$ without loss of generality. Let $\mathcal X_L$ and $\mathcal X_{L,M}$ denote the corresponding regularity classes with $X_0=0$ but no prescribed observations at later times; the policy produces the observation record. A deterministic policy $\pi\in\Pi_N$ observes $X_0$ and $X_T$ without cost and may request at most $N-1$ additional values of $X$. Each measurement time may depend on all previous queries and their observed values, and the queries need not be ordered chronologically. Thus $N$ denotes the maximum number of resulting observation intervals. The complete sequence of queries and observations, including the stopping decision, is denoted by $\mathcal O_\pi(X)$.

Two admissible paths are indistinguishable to the policy if they produce the same observation transcript. Given a path $X\in\mathcal C$, where $\mathcal C\in\{\mathcal X_L,\mathcal X_{L,M}\}$, the paths remaining compatible with its transcript form
\[
\mathcal F_\pi(X;\mathcal C)
=\{\widetilde X\in\mathcal C:
\mathcal O_\pi(\widetilde X)=\mathcal O_\pi(X)\}.
\]
For a set of drivers $B$, write $F(B)=\{F(Y):Y\in B\}$. The policy therefore leaves a range of possible outputs $F(\mathcal F_\pi(X;\mathcal C))$.

We seek the observation policy that minimizes the largest possible residual output radius. For any nonempty bounded set $A\subseteq\mathbb R$, define
$\operatorname{rad}(A)=\operatorname{diam}(A)/2=(\sup A-\inf A)/2$. This leads to the deterministic minimax radius of information,
\begin{equation}
\label{eq:minimax_information_radius}
\mathfrak R_N(F;\mathcal C)
:=\inf_{\pi\in\Pi_N}\sup_{X\in\mathcal C}
\operatorname{rad}\!\left(
F\bigl(\mathcal F_\pi(X;\mathcal C)\bigr)
\right).
\end{equation}
For a fixed policy, the supremum measures its worst-case ambiguity over admissible input paths. The infimum then selects the policy with the smallest such worst-case radius. Hence $\mathfrak R_N(F;\mathcal C)$ quantifies the information about the terminal response that cannot be resolved by any deterministic adaptive sampling strategy with the given observation budget.

\section{Geometry of Partial Identification}
\label{sec:partial_geometry}

\subsection{Unobserved area and terminal states}
\label{subsec:partial_geometry_state}

We specialize to planar drivers ($d=2$). Fix an observed record $S$ and a prescribed speed bound $L>0$ such that $\mathcal X_L(S)\ne\varnothing$. Let $\bar X$ be the piecewise-linear interpolation of the observed values, defined on each interval $[t_i,t_{i+1}]$ by
\[
\bar X_t=x_i+\frac{t-t_i}{t_{i+1}-t_i}(x_{i+1}-x_i).
\]
The feasibility condition $\|x_{i+1}-x_i\|\le L(t_{i+1}-t_i)$ ensures that $\bar X\in\mathcal X_L(S)$. We use this path as a reference for the CDE dynamics and write $\bar z_t=Z_t(\bar X)$ for the corresponding hidden state, with $\bar z_i=\bar z_{t_i}$.

The observations determine the displacement $x_{i+1}-x_i$ on each interval $[t_i,t_{i+1}]$. The signed area traced by the path relative to its chord is
\[
a_i(X)=\frac12\int_{t_i}^{t_{i+1}}
\left[
(X_t^1-x_i^1)\,\mathrm dX_t^2
-(X_t^2-x_i^2)\,\mathrm dX_t^1
\right].
\]
Let $A_\star(r,\ell)$ denote the maximum absolute signed area enclosed by a planar path of length at most $\ell$ and the chord joining its endpoints, whose distance is $r$. The fixed-endpoint Dido theorem gives the exact range of admissible areas on observation interval $i$ as $[-A_i^\star,A_i^\star]$, where $A_i^\star=A_\star(\|x_{i+1}-x_i\|,L(t_{i+1}-t_i))$. Under the speed constraint, these areas can be chosen independently across observation intervals. The geometric result and the characterization of $A_\star$ are given in Appendix~\ref{app:geometry}.

To describe how the unobserved signed area affects the CDE, let $\Phi_{s,t}^{X}(z)$ denote the state at time $t$ obtained by solving the CDE along $X$, starting from $z\in\mathbb R^m$ at time $s$. We compare the evolutions along $X$ and $\bar X$ on interval $[t_i,t_{i+1}]$, both starting from the reference state $\bar z_i$. Since the paths have the same endpoints, their leading-order difference is
\[
\begin{aligned}
&\Phi_{t_i,t_{i+1}}^X(\bar z_i)
-\Phi_{t_i,t_{i+1}}^{\bar X}(\bar z_i)\\
&\quad=a_i(X)[V_1,V_2](\bar z_i)
+O\!\left(L^3(t_{i+1}-t_i)^3\right),
\end{aligned}
\]
where $[V_1,V_2]=DV_2\,V_1-DV_1\,V_2$. The remainder is uniform over admissible path segments. The local expansion is proved in Appendix~\ref{app:identified_set_proof}.

The area-induced change appears in the state at $t_{i+1}$ and affects the terminal state through the subsequent dynamics. Along the reference path, the derivative of the state evolution on interval $i$ with respect to its initial state is $
J_i=D_z\Phi_{t_i,t_{i+1}}^{\bar X}(\bar z_i) \in\mathbb R^{m\times m}$. It maps a small change in the state at $t_i$ to its first-order effect at $t_{i+1}$.

A change in the state at $t_{i+1}$ is subsequently transported by $J_{i+1}$, then by $J_{i+2}J_{i+1}$, and so on. The complete linear transport from $t_{i+1}$ to $T$ is
\[
P_i=J_{n-1}\cdots J_{i+1}
\in\mathbb R^{m\times m},
\qquad P_{n-1}=I.
\]
The terminal-state direction associated with one unit of signed area on interval $i$ is therefore $b_i=P_i[V_1,V_2](\bar z_i)\in\mathbb R^m$.

Combining the feasible areas over all observation intervals gives the zonotope
\begin{equation}
\label{eq:partial_state_zonotope}
K(S)=
\left\{
\sum_{i=0}^{n-1}a_i b_i:
|a_i|\le A_i^\star\ \text{for all }i
\right\}
\subseteq\mathbb R^m.
\end{equation}
The coefficients $a_i$ in the definition of $K(S)$ vary independently over the intervals $[-A_i^\star,A_i^\star]$. For a particular admissible path $X$, they take the values $a_i(X)$. Thus the zonotope contains all combinations of the leading area-induced changes permitted by the observations.

To state the approximation error, define the sums of squared and cubed variation bounds over the observation intervals:
\[
\begin{aligned}
S_2&=\sum_{i=0}^{n-1}L^2(t_{i+1}-t_i)^2,\quad
S_3=\sum_{i=0}^{n-1}L^3(t_{i+1}-t_i)^3.
\end{aligned}
\]

To compare the terminal-state identified set with its zonotopic approximation $\bar z_T+K(S)$, we use the Hausdorff distance between nonempty compact sets $A,B\subseteq\mathbb R^m$, defined by
\[
d_{\mathrm H}(A,B)
=
\max\left\{
\sup_{a\in A}\inf_{b\in B}\|a-b\|,
\sup_{b\in B}\inf_{a\in A}\|b-a\|
\right\}.
\]
The two terms measure how far each set extends beyond the other. A bound on $d_{\mathrm H}$ therefore controls both the approximation of admissible terminal states by the zonotope and the approximation of points in the zonotope by admissible terminal states.

\begin{theorem}[Planar response identified set]
\label{thm:partial_state_zonotope}
Let $T,L>0$, $z_0\in\mathbb R^m$, and $V_1,V_2\in C_b^4(\mathbb R^m;\mathbb R^m)$. For an observation record $S=\{(t_i,x_i)\}_{i=0}^n$ with $0=t_0<\cdots<t_n=T$ and $x_i\in\mathbb R^2$, let $\mathcal X_L(S)$ consist of all paths $X\in W^{1,\infty}([0,T];\mathbb R^2)$ satisfying $X_{t_i}=x_i$ for every $i$ and $\|\dot X\|_{L^\infty}\le L$.

There exists a constant $C_Z=C_Z(V_1,V_2,L,T)>0$ such that, for every observation record $S$ with $\mathcal X_L(S)\neq\varnothing$, the terminal-state identified set $\mathcal Z_T(S;\mathcal X_L)$ satisfies
\begin{equation}
\label{eq:partial_state_hausdorff}
d_{\mathrm H}\!\left(
\mathcal Z_T(S;\mathcal X_L),
\bar z_T+K(S)
\right)
\le C_Z(S_3+S_2^2).
\end{equation}
The same constant applies to all such observation records, independently of the number and placement of measurements. Its explicit value and the complete proof are given in Appendix~\ref{app:identified_set_proof}.
\end{theorem}

Each observation interval contributes to the leading terminal-state ambiguity through the segment $[-A_i^\star,A_i^\star]b_i$, whose extent is determined by the unresolved signed area and whose direction is determined by the transported Lie bracket.

Since $K(S)$ is centrally symmetric, the terminal-state diameter satisfies
\[
\begin{aligned}
\operatorname{diam}\mathcal Z_T(S;\mathcal X_L)
&=2\max_{\|u\|=1}
\sum_{i=0}^{n-1}A_i^\star|u^\top b_i|
+O(S_3+S_2^2).
\end{aligned}
\]
Writing $\delta=\max_i(t_{i+1}-t_i)$, we have $S_2\le L^2T\delta$ and $S_3\le L^3T\delta^2$. The approximation error is therefore $O(\delta^2)$ for fixed dynamics, $L$, and $T$. The leading ambiguity is determined by the transported brackets and may vanish when the dynamics are insensitive to the unresolved areas.

\subsection{Target-specific ambiguity certificate}
\label{subsec:partial_geometry_certificate}

The state identified set describes variation in the full hidden state. For a scalar prediction $F(X)=g(Z_T(X))$, the effect of a small change in the terminal state is determined at first order by the gradient of $g$. Let $g\in C^2(\mathbb R^m;\mathbb R)$ have bounded first and second derivatives. Applying this gradient to the area-induced direction $b_i$ gives the scalar sensitivity $c_i=\nabla g(\bar z_T)^\top b_i\in\mathbb R$.

The resulting target-specific ambiguity radius is
\begin{equation}
\label{eq:partial_target_radius}
R_2(S)=\sum_{i=0}^{n-1}A_i^\star|c_i|.
\end{equation}
Each term combines the area permitted by the observations with the sensitivity of the prediction to the corresponding terminal-state direction.

\begin{theorem}[Target-specific ambiguity certificate]
\label{thm:partial_target_certificate}
Under the assumptions of Theorem~\ref{thm:partial_state_zonotope}, let $g\in C^2(\mathbb R^m;\mathbb R)$ have bounded first and second derivatives. There exists a constant $C_Y=C_Y(V_1,V_2,g,L,T)>0$ such that, for every observation record $S$ with $\mathcal X_L(S)\neq\varnothing$, the scalar response identified set $\mathcal Y_T(S;\mathcal X_L)$ satisfies
\begin{equation}
\label{eq:partial_target_hausdorff}
\begin{aligned}
d_{\mathrm H}\bigl(
\mathcal Y_T(S;\mathcal X_L),
&[F(\bar X)-R_2(S),\\
&F(\bar X)+R_2(S)]
\bigr)
\le C_Y(S_3+S_2^2).
\end{aligned}
\end{equation}
Consequently,
\[
\left|
\operatorname{diam}\mathcal Y_T(S;\mathcal X_L)
-2R_2(S)
\right|
\le 2C_Y(S_3+S_2^2).
\]
The same constant applies to all such observation records, independently of the number and placement of measurements. Its explicit value and the complete proof are given in Appendix~\ref{app:certificate_proof}.
\end{theorem}

The certificate approximates the half-width of the scalar response identified set. An interval with $c_i=0$ contributes no leading-order ambiguity to the prediction, even if its feasible area is nonzero. When all coefficients vanish, the remaining output variation is controlled by the higher-order remainder.

The coefficients $c_i$ can be computed without explicitly forming the transport products $P_i$. Let $\lambda_i\in\mathbb R^m$ denote the sensitivity of the terminal scalar output to the reference state at time $t_i$, with the subsequent control fixed to $\bar X$. At the terminal time this sensitivity is $\lambda_n=\nabla g(\bar z_T)$. Propagating it backward through the derivatives $J_i$ gives
\[
\lambda_i=J_i^\top\lambda_{i+1},
\qquad i=n-1,\ldots,0.
\]
Since $\lambda_{i+1}=P_i^\top\nabla g(\bar z_T)$, we obtain
\[
c_i=\lambda_{i+1}^\top[V_1,V_2](\bar z_i).
\]
The adjoint computation is justified in Appendix~\ref{app:certificate_proof}.

\section{Minimax Limits of Observation}
\label{sec:minimax_limits}

For a fixed set of observations, the identified set describes the remaining uncertainty in the CDE response. We study how this uncertainty decreases as more measurements become available. The minimax radius $\mathfrak R_N$, defined in~\eqref{eq:minimax_information_radius}, gives the smallest worst-case radius attainable by a deterministic adaptive policy. Throughout this section, the driver is planar, the endpoints $X_0$ and $X_T$ are observed without cost, and the policy may request at most $N-1$ additional values.

\subsection{Information limits for nondegenerate planar CDEs}
\label{subsec:minimax_general}

The minimax bounds depend on how strongly the terminal prediction responds to unobserved path geometry. To quantify this dependence, fix a unit vector $u\in\mathbb R^2$ and a speed $0<v<L$. Consider the straight reference driver $\bar X_t=vtu$ and its CDE solution $\bar z_t=Z_t(\bar X)$. The sensitivity of the terminal output to the state at time $t$ is described by the adjoint $\lambda_t\in\mathbb R^m$, which satisfies
\[
\begin{aligned}
\dot\lambda_t
&=-vD(u_1V_1+u_2V_2)(\bar z_t)^\top\lambda_t,\\
\lambda_T&=\nabla g(\bar z_T).
\end{aligned}
\]
The corresponding sensitivity to signed area is
\[
\beta_{v,u}(t)
=\bigl|\lambda_t^\top[V_1,V_2](\bar z_t)\bigr|.
\]
This is the continuous-time counterpart of the target-specific coefficients $c_i$ from Section~\ref{sec:partial_geometry}.

We assume that the prediction is sensitive to signed area on at least one interval $I=[a,b]\subset(0,T)$, with $a<b$. Specifically, $\beta_{v,u}(t)\ge\beta_0>0$ for every $t\in I$. This condition allows the influence of unobserved path geometry to be bounded from below.

Recall the maximal-area function $A_\star(r,\ell)$ from Section~\ref{sec:partial_geometry}. On a reference interval of duration $s$, the endpoint displacement has length $vs$ and the available path length is $Ls$. The scaling of planar area gives $A_\star(vs,Ls)=s^2A_\star(v,L)$. Since $v<L$, the coefficient $A_\star(v,L)$ is strictly positive.

\begin{theorem}[Bounded-speed information limit]
\label{thm:minimax_speed}
Let $T,L>0$, $V_1,V_2\in C_b^4(\mathbb R^m;\mathbb R^m)$, and $g\in C_b^2(\mathbb R^m;\mathbb R)$. Consider the global class $\mathcal X_L$ of planar drivers $X\in W^{1,\infty}([0,T];\mathbb R^2)$ satisfying $X_0=0$ and $\|\dot X\|_{L^\infty}\le L$. Suppose there exist a unit vector $u\in\mathbb R^2$, a speed $0<v<L$, and an interval $I\subset(0,T)$ such that $\beta_{v,u}(t)\ge\beta_0>0$ for all $t\in I$. Then
\begin{equation}
\label{eq:minimax_speed_lower}
\begin{aligned}
&\liminf_{N\to\infty}N\mathfrak R_N(F;\mathcal X_L)\\
&\qquad\ge A_\star(v,L)
\left(\int_I\sqrt{\beta_{v,u}(t)}\,\mathrm dt\right)^2.
\end{aligned}
\end{equation}
Uniform sampling provides a matching worst-case upper bound of order $N^{-1}$. Consequently, $\mathfrak R_N(F;\mathcal X_L)=\Theta(N^{-1})$.
\end{theorem}

The positive lower constant depends on the feasible hidden area and the sensitivity of the terminal prediction to that area. Positivity of $\beta_{v,u}$ on a single interval is sufficient to establish the $N^{-1}$ minimax rate over all deterministic adaptive observation policies. The constant is independent of the observation budget and the policy. The complete proof is given in Appendix~\ref{app:minimax_speed_proof}.

\begin{theorem}[Bounded-acceleration information limit]
\label{thm:minimax_acceleration}
Let $T,L,M>0$, $V_1,V_2\in C_b^4(\mathbb R^m;\mathbb R^m)$, and $g\in C_b^3(\mathbb R^m;\mathbb R)$. Consider the global class $\mathcal X_{L,M}$ of planar drivers $X\in W^{2,\infty}([0,T];\mathbb R^2)$ satisfying $X_0=0$, $\|\dot X\|_{L^\infty}\le L$, and $\|\ddot X\|_{L^\infty}\le M$. Suppose there exist a unit vector $u\in\mathbb R^2$, a speed $0<v<L$, and an interval $I\subset(0,T)$ such that $\beta_{v,u}(t)\ge\beta_0>0$ for all $t\in I$. Then
\begin{equation}
\label{eq:minimax_acceleration_lower}
\begin{aligned}
&\liminf_{N\to\infty}N^2\mathfrak R_N(F;\mathcal X_{L,M})\\
&\qquad\ge\frac{vM}{32}
\left(\int_I\beta_{v,u}(t)^{1/3}\,\mathrm dt\right)^3.
\end{aligned}
\end{equation}
Uniform sampling provides a matching worst-case upper bound of order $N^{-2}$. Hence $\mathfrak R_N(F;\mathcal X_{L,M})=\Theta(N^{-2})$.
\end{theorem}

Bounding acceleration restricts how far a compatible continuous path can deviate from the piecewise-linear reference between observation times. Under the same nondegeneracy condition, the minimax radius decreases at rate $N^{-2}$, compared with $N^{-1}$ under the speed constraint alone. The lower bound retains the dependence on the prediction target through $\beta_{v,u}$ and applies to every deterministic adaptive policy. The complete proof is given in Appendix~\ref{app:minimax_acceleration_proof}.

\paragraph{Sharp canonical case.}
An exact finite-budget result is available for the CDE that computes signed area. Consider the three-dimensional system with initial state $Z_0=0$, vector fields $V_1(z)=(1,0,-z_2/2)^\top$ and $V_2(z)=(0,1,z_1/2)^\top$, and readout $g(z)=z_3$. Its output equals the signed area of the planar input path relative to its starting point. Although these vector fields and the readout are unbounded on $\mathbb R^3$, for fixed $L,T$ they admit smooth bounded extensions that agree with them on a bounded neighborhood of the states reachable from $Z_0=0$ under $\mathcal X_L$. These extensions leave the output unchanged on $\mathcal X_L$.

\begin{proposition}[Exact minimax radius for the area CDE]
\label{thm:minimax_area_exact}
Let $T,L>0$ and $N\ge1$. For the bounded-speed class $\mathcal X_L$, the area-sensitive CDE satisfies
\begin{equation}
\label{eq:minimax_area_exact}
\mathfrak R_N(F;\mathcal X_L)
=\frac{L^2T^2}{2\pi N}.
\end{equation}
A uniform nonadaptive observation grid attains this radius among all deterministic adaptive policies with at most $N$ final observation intervals.
\end{proposition}

The exact equality holds for every finite observation budget. For this CDE, adaptive selection of measurement times does not improve the worst-case information radius achieved by uniform sampling. The complete proof is given in Appendix~\ref{app:exact_minimax_area}.

\section{Experiments}
\label{sec:experiments}

We train Neural CDEs on densely observed planar trajectories and evaluate them using progressively sparser observations, keeping the model parameters fixed. The experiments include two path-dependent regression targets, signed area and the output of a nonlinear controlled system, together with an endpoint-dependent negative control. All reported diameters and output spreads are normalized by the training-set standard deviation of the corresponding target. The experimental procedures and additional results are provided in Appendices~\ref{app:experimental_protocol} and~\ref{app:additional_experiments}.

\subsection{Certificate accuracy for admissible completions}
\label{subsec:exp_certificate}

We evaluate the certificate by comparing its predicted diameter $2R_2(S)$ with the output variation of two admissible continuous paths sharing the same observations. On each observation interval, circular-arc completions attain the maximal positive and negative signed areas permitted by the speed bound. The first path chooses the sign of each area according to the corresponding coefficient $c_i$, and the second uses the opposite signs. Both paths agree at every observation time and satisfy the prescribed speed constraint. Their terminal output difference provides a feasible lower bound on the identified-set diameter.

\begin{figure*}[t]
\centering
\includegraphics[width=.48\textwidth]{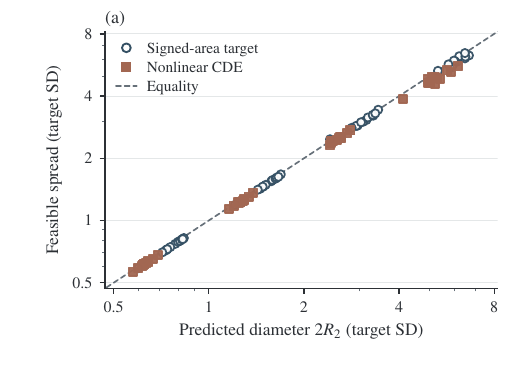}\hfill
\includegraphics[width=.48\textwidth]{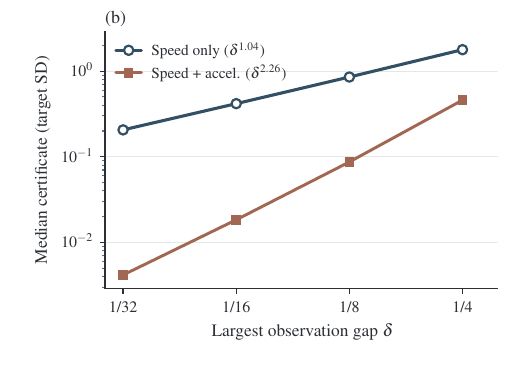}
\caption{\textbf{Certificate accuracy and path regularity.} (a) Predicted diameter $2R_2(S)$ against the nonlinear output spread between admissible circular-arc completions. The dashed line indicates equality. (b) Median leading certificate diameters under bounded-speed and bounded-acceleration assumptions as the maximum observation gap decreases. All widths are normalized by the corresponding target standard deviation.}
\label{fig:exp_geometry_regularity}
\end{figure*}

Across 96 evaluations of the path-dependent targets, the predicted diameter and the feasible output spread have a Spearman correlation of $0.9987$. The median ratio of feasible to predicted diameter is $0.981$. As the number of observations increases from five to 33, the median relative discrepancy decreases from $5.49\%$ to $1.69\%$ (Figure~\ref{fig:exp_geometry_regularity}, left).

The feasible spread decreases with a fitted log--log slope of $1.01$ against the maximum observation gap, while the absolute discrepancy decreases with a slope of $1.55$. For the endpoint-dependent negative control, the median certificate diameter is only $0.07\%$ of the corresponding scale for path-dependent targets.

\subsection{Effect of path regularity}
\label{subsec:exp_regularity}

We compare the bounded-speed certificate with the acceleration-aware leading outer approximation on smooth input paths satisfying both velocity and acceleration constraints. Across 72 evaluations of path-dependent targets, the acceleration-aware approximation has a median width equal to $7.11\%$ of the bounded-speed certificate. The fitted sampling exponents are $1.04$ and $2.26$, respectively (Figure~\ref{fig:exp_geometry_regularity}, right).

To assess the acceleration-aware approximation, we optimize over globally $C^1$ admissible completions whose velocities agree at observation times. The resulting feasible output spread has a Spearman correlation of $0.975$ with the outer approximation, but its median magnitude is only $9.11\%$ of that approximation.

The outer approximation combines bounds obtained separately on individual observation intervals. Globally $C^1$ paths must also satisfy velocity compatibility across neighboring intervals, and the numerical search explores a restricted family of admissible completions. The observed difference therefore does not determine how much conservatism arises from each source. The fitted exponents describe the finite-resolution certificates; the minimax results in Section~\ref{sec:minimax_limits} concern worst-case ambiguity over adaptive observation policies.

\subsection{Certificate-guided observation design}
\label{subsec:exp_acquisition}

We evaluate sequential measurement selection starting from five uniformly or irregularly spaced observations. Each policy acquires six additional measurements, selecting the midpoint of one observation interval at each step. The trained CDE remains fixed, and its ambiguity certificate is recomputed after every acquisition.

We compare the rule that selects the largest local contribution $A_i^\star|c_i|$ with a closed-form midpoint-reduction score, longest-gap selection, geometry-only and sensitivity-only scores, and random selection. A one-step look-ahead oracle evaluates the underlying signal at every candidate midpoint and selects the measurement producing the smallest recomputed certificate radius.

We measure acquisition performance by the area under the certificate-radius curve, normalized by its initial value. Lower values indicate faster reduction of the certificate over the six acquisitions.

Across 24 initial observation records, the median normalized area is $3.633$ for both largest-local-ambiguity and midpoint-reduction selection. The one-step oracle achieves $3.626$, compared with $3.823$ for longest-gap selection and $4.544$ for random selection.

The two certificate-guided rules choose different initial intervals in four of the 24 cases. The midpoint-reduction score produces a smaller recomputed radius in one of those four cases. On this benchmark, selecting the largest local ambiguity achieves nearly the same certificate reduction as the one-step oracle, with no additional improvement from the midpoint-reduction score.

\section{Discussion and Limitations}
\label{sec:discussion}

The identified set characterizes the terminal responses of a fixed CDE across continuous paths compatible with the observations and prescribed regularity bounds. When the driver represents a physical signal, its width quantifies the effect of unobserved evolution between measurements. If the continuous path serves only as a computational representation of discrete data, the certificate instead approximates the variation in model outputs across the prescribed class of alternative continuous reconstructions. Predictive accuracy also depends on the learned dynamics and the relationship between the CDE output and the target variable.

The admissible path class determines the range of possible responses. A bounded-speed assumption gives an exact feasible signed-area interval on each observation gap. Bounding acceleration further restricts this range and requires velocities to agree across neighboring intervals. The acceleration-aware certificate consequently provides an outer approximation to the leading ambiguity. Its practical interpretation requires regularity bounds that reflect the underlying signal, since bounds that exclude the true path invalidate the corresponding identification guarantee.

The present geometric characterization applies to planar drivers. The minimax rates additionally require nonvanishing transported Lie-bracket sensitivity and concern deterministic adaptive observation policies. For higher-dimensional inputs, the unresolved second-level geometry contains coupled area components whose joint feasibility must be characterized. Establishing a corresponding identified-set approximation would extend the certificate beyond planar controls. The certificate-guided acquisition rule also raises the question of how closely its sequential performance approaches the optimal adaptive information radius.

Our experiments evaluate the geometric certificate and sequential measurement selection on synthetic planar signals. Assessing their use for physical sensing requires densely recorded continuous processes, where additional measurements can be acquired and their effect on predictive performance evaluated.

\section{Conclusion}
\label{sec:conclusion}

Discrete observations of a continuous driver can leave the terminal response of a fixed CDE undetermined. For planar bounded-speed inputs, we characterize the leading geometry of the response identified set through feasible signed areas and transported Lie brackets, and derive a computable certificate for scalar predictions. The certificate quantifies how the unobserved geometry on each observation interval contributes to uncertainty in a given output. For nondegenerate planar CDEs, the minimax residual radius under deterministic adaptive sampling scales as $N^{-1}$ with bounded speed and as $N^{-2}$ when acceleration is also bounded. The resulting characterization connects the information missing from a given observation record with the rate at which further measurements can reduce the ambiguity.

\bibliography{example_paper}
\bibliographystyle{icml2026}

\newpage
\appendix
\onecolumn

\section{Mathematical Preliminaries and Geometric Results}
\label{app:geometry}

\subsection{Conventions and compatibility with observations}
\label{app:geometry_conventions}

Throughout the proofs, vector norms are Euclidean and norms of linear or multilinear maps are induced operator norms. Since the drivers are absolutely continuous, every integral against $\mathrm dX_t$ is evaluated as an integral against $\dot X_t\,\mathrm dt$. For planar vectors we write $u\wedge v=u^1v^2-u^2v^1$; in particular, $|u\wedge v|\le\|u\|\|v\|$. Our convention for the Lie bracket is $[V_1,V_2]=DV_2\,V_1-DV_1\,V_2$.

The constants used below depend only on the parameters specified at their definitions. They are independent of the observation partition and admissible driver; constants in the minimax bounds are also independent of the observation policy and budget $N$. Any constant used to select a sufficiently small interval or absorb a remainder is fixed before that choice.

The piecewise-linear path through the observed points is denoted by $\bar X$. On $[t_i,t_{i+1}]$ it is given by
\[
\bar X_t=x_i+\frac{t-t_i}{t_{i+1}-t_i}(x_{i+1}-x_i),\qquad \dot{\bar X}_t=\frac{x_{i+1}-x_i}{t_{i+1}-t_i}.
\]

\begin{lemma}[Feasibility under a speed bound]
\label{lem:app_speed_feasibility}
Let $0=t_0<\cdots<t_n=T$ and $L>0$. The completion class $\mathcal X_L(S)$ defined in Section~\ref{sec:problem} is nonempty if and only if $\|x_{i+1}-x_i\|\le L(t_{i+1}-t_i)$ for every $i=0,\ldots,n-1$.
\end{lemma}
\begin{proof}
If $X\in\mathcal X_L(S)$, the fundamental theorem of calculus and the speed constraint yield
\[
\begin{aligned}
\|x_{i+1}-x_i\|
&=\left\|\int_{t_i}^{t_{i+1}}\dot X_t\,\mathrm dt\right\|\le\int_{t_i}^{t_{i+1}}\|\dot X_t\|\,\mathrm dt\le L(t_{i+1}-t_i).
\end{aligned}
\]
Conversely, if all these inequalities hold, then on each open observation interval
\[
\|\dot{\bar X}_t\|=\frac{\|x_{i+1}-x_i\|}{t_{i+1}-t_i}\le L.
\]
The segments join continuously at the observations. Therefore $\bar X\in W^{1,\infty}([0,T];\mathbb R^2)$, agrees with $S$, and belongs to $\mathcal X_L(S)$. No matching of velocities across the observation times is needed in $W^{1,\infty}$.
\end{proof}

The same chord inequalities are necessary for $\mathcal X_{L,M}(S)$, but not sufficient: a globally $W^{2,\infty}$ path must also have compatible velocities at neighboring observation intervals.

\subsection{Signed areas and the Dido fiber}
\label{app:geometry_dido}

On an observation interval, the second-level iterated integrals are
\[
\mathbb X_i^{jk}=\int_{t_i}^{t_{i+1}}(X_t^j-x_i^j)\,\mathrm dX_t^k,\qquad j,k\in\{1,2\}.
\]
Their antisymmetric part is the signed area relative to the straight chord:
\[
\begin{aligned}
a_i(X)
&=\frac12\bigl(\mathbb X_i^{12}-\mathbb X_i^{21}\bigr)=\frac12\int_{t_i}^{t_{i+1}}\bigl[(X_t^1-x_i^1)\,\mathrm dX_t^2-(X_t^2-x_i^2)\,\mathrm dX_t^1\bigr].
\end{aligned}
\]
To see which part the endpoints determine, apply the product rule:
\[
\begin{aligned}
\mathbb X_i^{12}+\mathbb X_i^{21}
&=\int_{t_i}^{t_{i+1}}\mathrm d\bigl[(X_t^1-x_i^1)(X_t^2-x_i^2)\bigr]=(x_{i+1}^1-x_i^1)(x_{i+1}^2-x_i^2).
\end{aligned}
\]
In particular, the symmetric part is fixed by the observed increment while $a_i(X)$ need not be. For the straight path, $\bar X_t-x_i$ and $\dot{\bar X}_t$ are parallel, so $a_i(\bar X)=0$.

\begin{lemma}[Fixed-endpoint Dido area fiber]
\label{lem:app_dido_fiber}
Fix two endpoints in $\mathbb R^2$ at distance $r$, and a positive path-length budget $\ell\ge r$. Among absolutely continuous curves between the endpoints whose lengths do not exceed $\ell$, the attainable signed areas relative to the chord form exactly $[-A_\star(r,\ell),A_\star(r,\ell)]$, where
\begin{equation}
\label{eq:app_dido_area}
\begin{aligned}
A_\star(r,\ell)&=\frac{\ell^2}{2\theta^2}(\theta-\sin\theta),\quad
\frac r\ell=\frac{2\sin(\theta/2)}{\theta},\quad 0\le\theta\le2\pi.
\end{aligned}
\end{equation}
At the endpoints, the expression is interpreted by continuity: $A_\star(\ell,\ell)=0$ and $A_\star(0,\ell)=\ell^2/(4\pi)$.
\end{lemma}
\begin{proof}
The extremality of a circular arc for the fixed-endpoint, fixed-length area problem is the classical Dido isoperimetric theorem \citep{juillet2010dido}. We use this theorem for maximality and calculate the maximal area and the complete feasible interval directly.

Let $R$ be the radius and $\theta$ the central angle of the maximizing arc. Its length is $\ell=R\theta$, while its chord has length $r=2R\sin(\theta/2)$. Thus
\[
\begin{aligned}
R&=\frac\ell\theta,\quad
\frac r\ell=\frac{2(\ell/\theta)\sin(\theta/2)}\ell
=\frac{2\sin(\theta/2)}\theta.
\end{aligned}
\]
The area bounded by the arc and its chord is the signed sector area minus the signed triangle area. Substituting the radius gives
\[
\begin{aligned}
A_\star(r,\ell)
&=\frac{R^2\theta}{2}-\frac{R^2\sin\theta}{2}
=\frac{R^2}{2}(\theta-\sin\theta)
=\frac{\ell^2}{2\theta^2}(\theta-\sin\theta).
\end{aligned}
\]
When $\theta\downarrow0$, the expansions $\sin\theta=\theta-\theta^3/6+O(\theta^5)$ and $2\sin(\theta/2)/\theta\to1$ give $A_\star(\ell,\ell)=0$. When $\theta=2\pi$, the chord vanishes and
\[
A_\star(0,\ell)=\frac{\ell^2}{2(2\pi)^2}(2\pi-\sin2\pi)=\frac{\ell^2}{4\pi}.
\]

It remains to show that every area between the extreme values is attained under the same speed bound. Fix an interval $[t_i,t_{i+1}]$, parameterize a maximizing arc $X^\star$ at constant speed $\ell/(t_{i+1}-t_i)$, and let $\bar X$ be the straight path between the same endpoints. For $0\le\alpha\le1$ set $X_t^{(\alpha)}=(1-\alpha)\bar X_t+\alpha X_t^\star$. The endpoints remain fixed, and for almost every $t$,
\[
\begin{aligned}
\|\dot X_t^{(\alpha)}\|
&=\|(1-\alpha)\dot{\bar X}_t+\alpha\dot X_t^\star\|
\le(1-\alpha)\frac r{t_{i+1}-t_i}+\alpha\frac\ell{t_{i+1}-t_i}
\le\frac\ell{t_{i+1}-t_i}.
\end{aligned}
\]
The area of this path is continuous in $\alpha$. Indeed, substituting the interpolation into its definition gives
\[
\begin{aligned}
2a_i(X^{(\alpha)})
={}&(1-\alpha)^2\int_{t_i}^{t_{i+1}}(\bar X_t-x_i)\wedge\mathrm d\bar X_t\\
&+\alpha^2\int_{t_i}^{t_{i+1}}(X_t^\star-x_i)\wedge\mathrm dX_t^\star\\
&+\alpha(1-\alpha)\int_{t_i}^{t_{i+1}}(\bar X_t-x_i)\wedge\mathrm dX_t^\star\\
&+\alpha(1-\alpha)\int_{t_i}^{t_{i+1}}(X_t^\star-x_i)\wedge\mathrm d\bar X_t.
\end{aligned}
\]
The four integrals are independent of $\alpha$; the area is therefore a polynomial in $\alpha$, with values $0$ at $\alpha=0$ and $A_\star(r,\ell)$ at $\alpha=1$. The intermediate value theorem gives every area in $[0,A_\star(r,\ell)]$. Reflection across the chord preserves endpoints and length and reverses signed area; when the endpoints coincide, any reflection axis through their common point serves the same purpose. Hence the full interval $[-A_\star(r,\ell),A_\star(r,\ell)]$ is attainable.
\end{proof}

For a speed bound $L$, the length on $[t_i,t_{i+1}]$ does not exceed $L(t_{i+1}-t_i)$. We therefore use the established notation $A_i^\star=A_\star(\|x_{i+1}-x_i\|,L(t_{i+1}-t_i))$. Formula~\eqref{eq:app_dido_area} also shows the homogeneity $A_\star(\rho r,\rho\ell)=\rho^2A_\star(r,\ell)$ for every $\rho>0$.

\begin{corollary}[Independent areas under a speed bound]
\label{cor:app_independent_areas}
Fix the observed planar record $S$ and a speed bound $L$ such that $\mathcal X_L(S)\ne\varnothing$. For any numbers $a_0,\ldots,a_{n-1}$ satisfying $|a_i|\le A_i^\star$, there exists $X\in\mathcal X_L(S)$ with $a_i(X)=a_i$ simultaneously for all $i$.
\end{corollary}
\begin{proof}
For each $i$, Lemma~\ref{lem:app_dido_fiber} supplies an absolutely continuous path on $[t_i,t_{i+1}]$ with the prescribed endpoints, signed area $a_i$, and speed at most $L$. Concatenate the paths in increasing order of their intervals. They agree at each shared endpoint $x_i$ and are uniformly $L$-Lipschitz, so the concatenation is a globally absolutely continuous path with $\|\dot X\|_{L^\infty}\le L$. It matches $S$ and realizes all the specified areas. The construction does not require its one-sided velocities to agree at the observation times.
\end{proof}

\begin{lemma}[Maximum of the Dido profile]
\label{lem:app_dido_max}
For every $\ell>0$ and $0\le r\le\ell$,
\begin{equation}
\label{eq:app_dido_uniform}
A_\star(r,\ell)\le\frac{\ell^2}{2\pi},\qquad
\text{with equality at }r=\frac{2\ell}{\pi}.
\end{equation}
\end{lemma}
\begin{proof}
For fixed $\ell$, differentiate the Dido expression directly in the angle $\theta$:
\[
\begin{aligned}
\frac{\mathrm d}{\mathrm d\theta}\frac{\theta-\sin\theta}{2\theta^2}
&=\frac{\theta(1-\cos\theta)-2(\theta-\sin\theta)}{2\theta^3}
=\frac{2\sin\theta-\theta(1+\cos\theta)}{2\theta^3}
=\frac{\cos(\theta/2)}{\theta^3}\bigl(2\sin(\theta/2)-\theta\cos(\theta/2)\bigr).
\end{aligned}
\]
The factor in parentheses is positive for $0<\theta<2\pi$, because it is zero at $\theta=0$ and its derivative is
\[
\begin{aligned}
\frac{\mathrm d}{\mathrm d\theta}\bigl(2\sin(\theta/2)-\theta\cos(\theta/2)\bigr)
&=\frac\theta2\sin(\theta/2)>0.
\end{aligned}
\]
The sign of the area derivative is therefore the sign of $\cos(\theta/2)$: positive for $0<\theta<\pi$ and negative for $\pi<\theta<2\pi$. At $\theta=\pi$,
\[
\begin{aligned}
\frac r\ell&=\frac{2\sin(\pi/2)}\pi=\frac2\pi,\\
A_\star(r,\ell)&=\frac{\ell^2}{2\pi^2}(\pi-\sin\pi)=\frac{\ell^2}{2\pi}.
\end{aligned}
\]
This proves the claimed sharp bound. The maximizing curve is a semicircle.
\end{proof}

\subsection{One-gap geometry under bounded acceleration}
\label{app:geometry_acceleration}

Fix one observation interval $[t_i,t_{i+1}]$ and assume $X\in W^{2,\infty}([t_i,t_{i+1}];\mathbb R^2)$ with $X_{t_i}=x_i$, $X_{t_{i+1}}=x_{i+1}$, $\|\dot X\|_{L^\infty}\le L$, and $\|\ddot X\|_{L^\infty}\le M$.

The signed area has the exact decomposition
\[
\begin{aligned}
a_i(X)
={}&\int_{t_i}^{t_{i+1}}(X_t-\bar X_t)\wedge\frac{x_{i+1}-x_i}{t_{i+1}-t_i}\,\mathrm dt
+\frac12\int_{t_i}^{t_{i+1}}(X_t-\bar X_t)\wedge\left(\dot X_t-\frac{x_{i+1}-x_i}{t_{i+1}-t_i}\right)\mathrm dt.
\end{aligned}
\]
To derive this identity, expand the defining area integral using $X_t-x_i=(\bar X_t-x_i)+(X_t-\bar X_t)$ and $\dot X_t=\dot{\bar X}_t+(\dot X_t-\dot{\bar X}_t)$:
\[
\begin{aligned}
2a_i(X)
={}&\int_{t_i}^{t_{i+1}}(\bar X_t-x_i)\wedge\dot{\bar X}_t\,\mathrm dt\\
&+\int_{t_i}^{t_{i+1}}(\bar X_t-x_i)\wedge(\dot X_t-\dot{\bar X}_t)\,\mathrm dt\\
&+\int_{t_i}^{t_{i+1}}(X_t-\bar X_t)\wedge\dot{\bar X}_t\,\mathrm dt\\
&+\int_{t_i}^{t_{i+1}}(X_t-\bar X_t)\wedge(\dot X_t-\dot{\bar X}_t)\,\mathrm dt.
\end{aligned}
\]
The first integral vanishes because its vectors are parallel. Integrating the second by parts and using $X_{t_i}=\bar X_{t_i}$ and $X_{t_{i+1}}=\bar X_{t_{i+1}}$ gives
\[
\begin{aligned}
\int_{t_i}^{t_{i+1}}(\bar X_t-x_i)\wedge(\dot X_t-\dot{\bar X}_t)\,\mathrm dt&=\left[(\bar X_t-x_i)\wedge(X_t-\bar X_t)\right]_{t_i}^{t_{i+1}}-\int_{t_i}^{t_{i+1}}\dot{\bar X}_t\wedge(X_t-\bar X_t)\,\mathrm dt\\
&=\int_{t_i}^{t_{i+1}}(X_t-\bar X_t)\wedge\frac{x_{i+1}-x_i}{t_{i+1}-t_i}\,\mathrm dt.
\end{aligned}
\]
The second and third integrals in the expansion are therefore equal. Dividing by two gives the stated decomposition.

\begin{lemma}[Deviation from the chord]
\label{lem:app_acc_deviation}
If $X\in W^{2,\infty}([t_i,t_{i+1}];\mathbb R^2)$ has the prescribed endpoints and $\|\ddot X\|_{L^\infty}\le M$, then, for every $t\in[t_i,t_{i+1}]$,
\[
\begin{aligned}
\|X_t-\bar X_t\|&\le\frac M2(t-t_i)(t_{i+1}-t),\\
\left\|\dot X_t-\frac{x_{i+1}-x_i}{t_{i+1}-t_i}\right\|
&\le\frac M{2(t_{i+1}-t_i)}\bigl[(t-t_i)^2+(t_{i+1}-t)^2\bigr].
\end{aligned}
\]
\end{lemma}
\begin{proof}
Integrating the acceleration twice from $t_i$ yields
\[
\begin{aligned}
X_t-x_i
&=\int_{t_i}^t\dot X_\tau\,\mathrm d\tau=(t-t_i)\dot X_{t_i}+\int_{t_i}^t\int_{t_i}^\tau\ddot X_s\,\mathrm ds\,\mathrm d\tau=(t-t_i)\dot X_{t_i}+\int_{t_i}^t(t-s)\ddot X_s\,\mathrm ds.
\end{aligned}
\]
The last equality follows by reversing the order of integration on $\{(s,\tau):t_i\le s\le\tau\le t\}$: for fixed $s$, the outer time $\tau$ ranges from $s$ to $t$. Setting $t=t_{i+1}$ gives
\[
\begin{aligned}
x_{i+1}-x_i
={}&(t_{i+1}-t_i)\dot X_{t_i}
+\int_{t_i}^{t_{i+1}}(t_{i+1}-s)\ddot X_s\,\mathrm ds.
\end{aligned}
\]
Solving this equality for $\dot X_{t_i}$ and substituting back into the earlier expression yields
\[
\begin{aligned}
\dot X_{t_i}
={}&\frac{x_{i+1}-x_i}{t_{i+1}-t_i}
-\frac1{t_{i+1}-t_i}\int_{t_i}^{t_{i+1}}(t_{i+1}-s)\ddot X_s\,\mathrm ds,\\[2pt]
X_t-\bar X_t
={}&\int_{t_i}^t(t-s)\ddot X_s\,\mathrm ds
-\frac{t-t_i}{t_{i+1}-t_i}\int_{t_i}^{t_{i+1}}(t_{i+1}-s)\ddot X_s\,\mathrm ds.
\end{aligned}
\]
Split the final integral at $t$ and collect the coefficients of $\ddot X_s$ on each part:
\[
\begin{aligned}
X_t-\bar X_t
={}&\int_{t_i}^t\left[(t-s)-\frac{(t-t_i)(t_{i+1}-s)}{t_{i+1}-t_i}\right]\ddot X_s\,\mathrm ds
-\frac{t-t_i}{t_{i+1}-t_i}\int_t^{t_{i+1}}(t_{i+1}-s)\ddot X_s\,\mathrm ds\\
={}&-\frac{t_{i+1}-t}{t_{i+1}-t_i}\int_{t_i}^t(s-t_i)\ddot X_s\,\mathrm ds
-\frac{t-t_i}{t_{i+1}-t_i}\int_t^{t_{i+1}}(t_{i+1}-s)\ddot X_s\,\mathrm ds.
\end{aligned}
\]
Applying $\|\ddot X_s\|\le M$ and evaluating both integrals explicitly gives
\[
\begin{aligned}
\|X_t-\bar X_t\|
&\le\frac{M(t_{i+1}-t)}{t_{i+1}-t_i}\int_{t_i}^t(s-t_i)\,\mathrm ds+\frac{M(t-t_i)}{t_{i+1}-t_i}\int_t^{t_{i+1}}(t_{i+1}-s)\,\mathrm ds\\
&=\frac{M(t_{i+1}-t)(t-t_i)^2}{2(t_{i+1}-t_i)}+\frac{M(t-t_i)(t_{i+1}-t)^2}{2(t_{i+1}-t_i)}\\
&=\frac M2(t-t_i)(t_{i+1}-t).
\end{aligned}
\]

For the velocity estimate, the endpoint condition gives $x_{i+1}-x_i=\int_{t_i}^{t_{i+1}}\dot X_s\,\mathrm ds$. Consequently,
\[
\begin{aligned}
\dot X_t-\frac{x_{i+1}-x_i}{t_{i+1}-t_i}
&=\frac1{t_{i+1}-t_i}\int_{t_i}^{t_{i+1}}(\dot X_t-\dot X_s)\,\mathrm ds.
\end{aligned}
\]
Since $\dot X_t-\dot X_s=\int_s^t\ddot X_\tau\,\mathrm d\tau$, we have $\|\dot X_t-\dot X_s\|\le M|t-s|$. Splitting the integral at $t$ then gives
\[
\begin{aligned}
\left\|\dot X_t-\frac{x_{i+1}-x_i}{t_{i+1}-t_i}\right\|&\le\frac M{t_{i+1}-t_i}\left[\int_{t_i}^t(t-s)\,\mathrm ds+\int_t^{t_{i+1}}(s-t)\,\mathrm ds\right]\\
&=\frac M{t_{i+1}-t_i}\left[\frac{(t-t_i)^2}{2}+\frac{(t_{i+1}-t)^2}{2}\right]\\
&=\frac M{2(t_{i+1}-t_i)}\bigl[(t-t_i)^2+(t_{i+1}-t)^2\bigr].
\end{aligned}
\]
This proves both estimates with the explicit constant $M$ from the acceleration assumption and no dependence on the number of observation intervals.
\end{proof}

\begin{proposition}[Acceleration-aware area bounds]
\label{prop:app_acc_area_bounds}
Let $A^\star_{L,M}(\|x_{i+1}-x_i\|,t_{i+1}-t_i)$ be the supremum of $|a_i(X)|$ over paths on $[t_i,t_{i+1}]$ with these endpoints, $\|\dot X\|_{L^\infty}\le L$ and $\|\ddot X\|_{L^\infty}\le M$. For every feasible such interval,
\begin{equation}
\label{eq:app_acc_area_bound}
\begin{aligned}
&A^\star_{L,M}(\|x_{i+1}-x_i\|,t_{i+1}-t_i)\\
&\ \le\min\Biggl\{ A_\star\bigl(\|x_{i+1}-x_i\|,L(t_{i+1}-t_i)\bigr),\frac{\|x_{i+1}-x_i\|M(t_{i+1}-t_i)^2}{12}+\frac{M^2(t_{i+1}-t_i)^4}{80}\Biggr\}.
\end{aligned}
\end{equation}
For any unit vector $\nu$ perpendicular to $x_{i+1}-x_i$ and any $\mu\in[0,M]$ satisfying
\[
\frac{\|x_{i+1}-x_i\|^2}{(t_{i+1}-t_i)^2}+\frac{\mu^2(t_{i+1}-t_i)^2}{4}\le L^2,
\]
the parabolic path
\[
X_t^{\mathrm{par}}=\bar X_t+\frac\mu2(t-t_i)(t_{i+1}-t)\nu
\]
is feasible on this interval and has $|a_i(X^{\mathrm{par}})|=\|x_{i+1}-x_i\|\mu(t_{i+1}-t_i)^2/12$.
\end{proposition}
\begin{proof}
By the exact area decomposition and the first estimate of Lemma~\ref{lem:app_acc_deviation}, the chord--deviation contribution satisfies
\[
\begin{aligned}
\left|\int_{t_i}^{t_{i+1}}(X_t-\bar X_t)\wedge\frac{x_{i+1}-x_i}{t_{i+1}-t_i}\,\mathrm dt\right|&\le\frac{\|x_{i+1}-x_i\|}{t_{i+1}-t_i}\int_{t_i}^{t_{i+1}}\|X_t-\bar X_t\|\,\mathrm dt
\le\frac{\|x_{i+1}-x_i\|M}{2(t_{i+1}-t_i)}\int_{t_i}^{t_{i+1}}(t-t_i)(t_{i+1}-t)\,\mathrm dt.
\end{aligned}
\]
The integral equals
\[
\begin{aligned}
\int_{t_i}^{t_{i+1}}(t-t_i)(t_{i+1}-t)\,\mathrm dt&=\left[\frac{(t_{i+1}-t_i)(t-t_i)^2}{2}-\frac{(t-t_i)^3}{3}\right]_{t_i}^{t_{i+1}}
=(t_{i+1}-t_i)^3\left(\frac12-\frac13\right)
=\frac{(t_{i+1}-t_i)^3}{6}.
\end{aligned}
\]
Substitution gives the first contribution $\|x_{i+1}-x_i\|M(t_{i+1}-t_i)^2/12$.

For the second contribution, use both bounds of Lemma~\ref{lem:app_acc_deviation}:
\[
\begin{aligned}
\frac12\left|\int_{t_i}^{t_{i+1}}(X_t-\bar X_t)\wedge\left(\dot X_t-\frac{x_{i+1}-x_i}{t_{i+1}-t_i}\right)\mathrm dt\right|&\le\frac12\int_{t_i}^{t_{i+1}}\|X_t-\bar X_t\|\left\|\dot X_t-\frac{x_{i+1}-x_i}{t_{i+1}-t_i}\right\|\mathrm dt\\
&\quad\le\frac{M^2}{8(t_{i+1}-t_i)}\int_{t_i}^{t_{i+1}}(t-t_i)(t_{i+1}-t)\times\bigl[(t-t_i)^2+(t_{i+1}-t)^2\bigr]\,\mathrm dt.
\end{aligned}
\]
Expanding the integrand gives
\[
\begin{aligned}
(t-t_i)(t_{i+1}-t)\bigl[(t-t_i)^2+(t_{i+1}-t)^2\bigr]&=-2(t-t_i)^4+4(t_{i+1}-t_i)(t-t_i)^3\\
&\quad-3(t_{i+1}-t_i)^2(t-t_i)^2+(t_{i+1}-t_i)^3(t-t_i).
\end{aligned}
\]
Its integral is
\[
\begin{aligned}
&\int_{t_i}^{t_{i+1}}(t-t_i)(t_{i+1}-t)\bigl[(t-t_i)^2+(t_{i+1}-t)^2\bigr]\,\mathrm dt\\
&=\Biggl[-\frac{2(t-t_i)^5}{5}+(t_{i+1}-t_i)(t-t_i)^4-(t_{i+1}-t_i)^2(t-t_i)^3+\frac{(t_{i+1}-t_i)^3(t-t_i)^2}{2}\Biggr]_{t_i}^{t_{i+1}}\\
&=(t_{i+1}-t_i)^5\left(-\frac25+1-1+\frac12\right)\\
&=\frac{(t_{i+1}-t_i)^5}{10}.
\end{aligned}
\]
Multiplying by $M^2/[8(t_{i+1}-t_i)]$ yields $M^2(t_{i+1}-t_i)^4/80$. Adding the two contributions gives the acceleration-based upper estimate. The length bound $\int_{t_i}^{t_{i+1}}\|\dot X_t\|\,\mathrm dt\le L(t_{i+1}-t_i)$ independently yields the Dido estimate; taking their minimum proves~\eqref{eq:app_acc_area_bound}.

To prove the lower witness statement, consider the displayed parabolic path. Differentiating gives
\[
\begin{aligned}
\dot X_t^{\mathrm{par}}
&=\frac{x_{i+1}-x_i}{t_{i+1}-t_i}+\frac\mu2(t_{i+1}+t_i-2t)\nu,\\
\ddot X_t^{\mathrm{par}}&=-\mu\nu.
\end{aligned}
\]
Since $\nu\perp(x_{i+1}-x_i)$ and $|t_{i+1}+t_i-2t|\le t_{i+1}-t_i$,
\[
\begin{aligned}
\|\dot X_t^{\mathrm{par}}\|^2
&=\frac{\|x_{i+1}-x_i\|^2}{(t_{i+1}-t_i)^2}+\frac{\mu^2(t_{i+1}+t_i-2t)^2}{4}\le\frac{\|x_{i+1}-x_i\|^2}{(t_{i+1}-t_i)^2}+\frac{\mu^2(t_{i+1}-t_i)^2}{4}
\le L^2,\\
\|\ddot X_t^{\mathrm{par}}\|&=\mu\le M.
\end{aligned}
\]
The deviation $X_t^{\mathrm{par}}-\bar X_t$ and its derivative are both parallel to $\nu$, so their wedge product is zero. Using the first term in the exact area decomposition and the integral already evaluated above,
\[
\begin{aligned}
a_i(X^{\mathrm{par}})
&=\frac\mu{2(t_{i+1}-t_i)}\bigl(\nu\wedge(x_{i+1}-x_i)\bigr)\times\int_{t_i}^{t_{i+1}}(t-t_i)(t_{i+1}-t)\,\mathrm dt
=\frac{\mu(t_{i+1}-t_i)^2}{12}\bigl(\nu\wedge(x_{i+1}-x_i)\bigr).
\end{aligned}
\]
Orthogonality and $\|\nu\|=1$ give $|\nu\wedge(x_{i+1}-x_i)|=\|x_{i+1}-x_i\|$, completing the proof.
\end{proof}

For the observed record $S$, assume that the prescribed speed and acceleration bounds admit a compatible path, so $\mathcal X_{L,M}(S)\ne\varnothing$. The resulting bound $\|x_{i+1}-x_i\|\le L(t_{i+1}-t_i)$ implies
\[
\begin{aligned}
&A^\star_{L,M}(\|x_{i+1}-x_i\|,t_{i+1}-t_i)\le\frac{LM}{12}(t_{i+1}-t_i)^3+\frac{M^2}{80}(t_{i+1}-t_i)^4.
\end{aligned}
\]
If $\delta=\max_i(t_{i+1}-t_i)$, summing these one-gap bounds and using $\sum_i(t_{i+1}-t_i)=T$ yields
\[
\begin{aligned}
\sum_{i=0}^{n-1}A^\star_{L,M}(\|x_{i+1}-x_i\|,t_{i+1}-t_i)
&\le\frac{LM}{12}\sum_i(t_{i+1}-t_i)^3+\frac{M^2}{80}\sum_i(t_{i+1}-t_i)^4
\le\frac{LMT}{12}\delta^2+\frac{M^2T}{80}\delta^3.
\end{aligned}
\]
For comparison, Lemma~\ref{lem:app_dido_max} gives $\sum_i A_i^\star\le L^2T\delta/(2\pi)$ under the speed-only constraint. These are upper bounds on the sums of gap-wise area budgets. In particular, under a global $W^{2,\infty}$ constraint, velocities must agree at shared observation times, so the one-gap areas cannot generally be selected independently. The proposition does not claim a sharp global acceleration-constrained identified set.

\section{Proof of the Response Identified-Set Theorem}
\label{app:identified_set_proof}

For $s\le t$, let $\Phi_{s,t}^{X}(z)$ denote the state at time $t$ of the CDE driven by $X$ on $[s,t]$, starting from state $z$ at time $s$. Then $Z_t(X)=\Phi_{0,t}^{X}(z_0)$ and $\bar z_t=\Phi_{0,t}^{\bar X}(z_0)$. Uniqueness gives $\Phi_{s,t}^{X}=\Phi_{r,t}^{X}\circ\Phi_{s,r}^{X}$ for $s\le r\le t$.

\subsection{The local CDE expansion}
\label{app:state_local_expansion}

The assumption $V_1,V_2\in C_b^4(\mathbb R^m;\mathbb R^m)$ makes the following field norms finite:
\[
M_k=\sup_{z\in\mathbb R^m}
\left(\sum_{j=1}^2\|D^kV_j(z)\|^2\right)^{1/2},
\qquad k=0,1,2,
\]
where $D^0V_j=V_j$. Thus $M_0$ bounds the vector fields, while $M_1$ and $M_2$ bound their first and second derivatives. These bounds depend only on the fixed vector fields.

\begin{lemma}[Uniform one-interval area expansion]
\label{lem:app_uniform_local_area}
Fix any observation interval $[t_i,t_{i+1}]$ and any admissible path $X$ on it. Compare the state at $t_{i+1}$ under $X$ with that under its straight chord $\bar X$, starting both solutions from the same arbitrary state $z$ at $t_i$. Then
\begin{equation}
\label{eq:app_uniform_local_area}
\begin{aligned}
&\bigl\|\Phi_{t_i,t_{i+1}}^{X}(z)-\Phi_{t_i,t_{i+1}}^{\bar X}(z)
-a_i(X)[V_1,V_2](z)\bigr\|
\le C_{\mathrm{loc}}L^3(t_{i+1}-t_i)^3.
\end{aligned}
\end{equation}
The fixed coefficient $C_{\mathrm{loc}}$ is defined at the end of the proof in terms of $M_0,M_1,M_2$. It is independent of the initial state, the interval, the skeleton, and the admissible completion.
\end{lemma}

\begin{proof}
Set $Z_t=\Phi_{t_i,t}^{X}(z)$, with $Z_{t_i}=z$. Its integral equation is
\[
Z_t-z=\sum_{j=1}^2\int_{t_i}^tV_j(Z_\tau)\dot X_\tau^j\,\mathrm d\tau.
\]
By the triangle inequality and Cauchy--Schwarz, for every $t\in[t_i,t_{i+1}]$,
\[
\begin{aligned}
\|Z_t-z\|
&\le\int_{t_i}^t\left\|\sum_{j=1}^2V_j(Z_\tau)\dot X_\tau^j\right\|\mathrm d\tau
\le\int_{t_i}^t\left(\sum_{j=1}^2\|V_j(Z_\tau)\|^2\right)^{1/2}\|\dot X_\tau\|\,\mathrm d\tau
\le M_0\int_{t_i}^t\|\dot X_\tau\|\,\mathrm d\tau.
\end{aligned}
\]
Subtract the first-order expression obtained by freezing the fields at $z$. Since $\int_{t_i}^t\dot X_\tau^j\,\mathrm d\tau=X_t^j-x_i^j$, the difference is exactly
\[
\begin{aligned}
&Z_t-z-\sum_{k=1}^2V_k(z)(X_t^k-x_i^k)
=\sum_{j=1}^2\int_{t_i}^t\bigl(V_j(Z_\tau)-V_j(z)\bigr)\dot X_\tau^j\,\mathrm d\tau.
\end{aligned}
\]
The mean-value formula, the definition of $M_1$, and the preceding state estimate give
\[
\begin{aligned}
&\left\|Z_t-z-\sum_{k=1}^2V_k(z)(X_t^k-x_i^k)\right\|
\le M_1\int_{t_i}^t\|Z_\tau-z\|\,\|\dot X_\tau\|\,\mathrm d\tau\\
&\quad\le M_0M_1\int_{t_i}^t\left(\int_{t_i}^\tau\|\dot X_s\|\,\mathrm ds\right)\|\dot X_\tau\|\,\mathrm d\tau
=\frac{M_0M_1}{2}\left(\int_{t_i}^t\|\dot X_\tau\|\,\mathrm d\tau\right)^2.
\end{aligned}
\]
The last equality is obtained by differentiating $\frac12(\int_{t_i}^\tau\|\dot X_s\|\,\mathrm ds)^2$ with respect to $\tau$ and evaluating the endpoints.

For each field, Taylor's formula at $z$ gives
\[
\begin{aligned}
V_j(Z_t)={}&V_j(z)+DV_j(z)(Z_t-z)
+\int_0^1(1-\theta)
D^2V_j\bigl(z+\theta(Z_t-z)\bigr)
[Z_t-z,Z_t-z] \,\mathrm d\theta.
\end{aligned}
\]
Substituting the first-order decomposition of $Z_t-z$ gives
\[
\begin{aligned}
&V_j(Z_t)-V_j(z)-\sum_{k=1}^2DV_j(z)V_k(z)(X_t^k-x_i^k)\\
&\quad=DV_j(z)\left(Z_t-z-\sum_{k=1}^2V_k(z)(X_t^k-x_i^k)\right)
+\int_0^1(1-\theta)D^2V_j\bigl(z+\theta(Z_t-z)\bigr)[Z_t-z,Z_t-z]
\mathrm d\theta.
\end{aligned}
\]
Take the Euclidean norm over the two field indices. The first term is controlled by the previous first-order error and $M_1$; the second uses $M_2$ and $\int_0^1(1-\theta)\,\mathrm d\theta=1/2$. Thus
\[
\begin{aligned}
&\Biggl(\sum_{j=1}^2\Biggl\|V_j(Z_t)-V_j(z)
-\sum_{k=1}^2DV_j(z)V_k(z)
(X_t^k-x_i^k)\Biggr\|^2\Biggr)^{1/2}\\
&\quad\le \frac{M_0M_1^2}{2}
\left(\int_{t_i}^t\|\dot X_\tau\|\,\mathrm d\tau\right)^2
+\frac{M_0^2M_2}{2}
\left(\int_{t_i}^t\|\dot X_\tau\|\,\mathrm d\tau\right)^2\\
&\quad=\frac{M_0M_1^2+M_0^2M_2}{2}
\left(\int_{t_i}^t\|\dot X_\tau\|\,\mathrm d\tau\right)^2.
\end{aligned}
\]

Return to the integral equation at $t_{i+1}$ and substitute the expanded fields term by term:
\[
\begin{aligned}
\Phi_{t_i,t_{i+1}}^{X}(z)
={}&z+\sum_{j=1}^2V_j(z)\int_{t_i}^{t_{i+1}}\mathrm dX_t^j
+\sum_{j,k=1}^2DV_j(z)V_k(z)\int_{t_i}^{t_{i+1}}(X_t^k-x_i^k)\,\mathrm dX_t^j
+\mathcal R_X(z)\\
={}&z+\sum_{j=1}^2V_j(z)(x_{i+1}^j-x_i^j)
+\sum_{j,k=1}^2DV_j(z)V_k(z)\mathbb X_i^{kj}+\mathcal R_X(z),
\end{aligned}
\]
where $\mathbb X_i^{kj}=\int_{t_i}^{t_{i+1}}(X_t^k-x_i^k)\,\mathrm dX_t^j$ is the second-level integral from Appendix~\ref{app:geometry}. The error $\mathcal R_X(z)$ is precisely the integral of the two-field difference bounded above. Consequently,
\[
\begin{aligned}
\|\mathcal R_X(z)\|
&\le\frac{M_0M_1^2+M_0^2M_2}{2}
\times\int_{t_i}^{t_{i+1}}
\left(\int_{t_i}^t\|\dot X_s\|\,\mathrm ds\right)^2
\|\dot X_t\|\,\mathrm dt\\
&=\frac{M_0M_1^2+M_0^2M_2}{6}
\times\left(\int_{t_i}^{t_{i+1}}
\|\dot X_t\|\,\mathrm dt\right)^3\\
&\le\frac{M_0M_1^2+M_0^2M_2}{6}
L^3(t_{i+1}-t_i)^3.
\end{aligned}
\]
To check the equality, differentiate $\frac13(\int_{t_i}^t\|\dot X_s\|\,\mathrm ds)^3$ and integrate the result over $[t_i,t_{i+1}]$.

The product rule identifies the second-level terms fixed by the endpoints:
\[
\begin{aligned}
\mathbb X_i^{11}
&=\left[\frac12(X_t^1-x_i^1)^2\right]_{t_i}^{t_{i+1}}=\frac12(x_{i+1}^1-x_i^1)^2,\\
\mathbb X_i^{22}
&=\left[\frac12(X_t^2-x_i^2)^2\right]_{t_i}^{t_{i+1}}=\frac12(x_{i+1}^2-x_i^2)^2,\\
\mathbb X_i^{12}+\mathbb X_i^{21}
&=\int_{t_i}^{t_{i+1}}\mathrm d\bigl[(X_t^1-x_i^1)(X_t^2-x_i^2)\bigr]
=(x_{i+1}^1-x_i^1)(x_{i+1}^2-x_i^2),\\
\mathbb X_i^{12}-\mathbb X_i^{21}
&=\int_{t_i}^{t_{i+1}}\bigl[(X_t^1-x_i^1)\,\mathrm dX_t^2-(X_t^2-x_i^2)\,\mathrm dX_t^1\bigr]
=2a_i(X).
\end{aligned}
\]
Adding and subtracting the last two identities gives
\[
\begin{aligned}
\mathbb X_i^{12}&=\frac12(x_{i+1}^1-x_i^1)(x_{i+1}^2-x_i^2)+a_i(X),\\
\mathbb X_i^{21}&=\frac12(x_{i+1}^1-x_i^1)(x_{i+1}^2-x_i^2)-a_i(X).
\end{aligned}
\]
For the straight path, the corresponding integrals are
\[
\begin{aligned}
\bar X_t^k-x_i^k
&=\frac{t-t_i}{t_{i+1}-t_i}(x_{i+1}^k-x_i^k),\\
\mathrm d\bar X_t^j
&=\frac{x_{i+1}^j-x_i^j}{t_{i+1}-t_i}\,\mathrm dt,\\
\int_{t_i}^{t_{i+1}}(t-t_i)\,\mathrm dt
&=\left[\frac{(t-t_i)^2}{2}\right]_{t_i}^{t_{i+1}}
=\frac{(t_{i+1}-t_i)^2}{2},\\
\int_{t_i}^{t_{i+1}}(\bar X_t^k-x_i^k)
\,\mathrm d\bar X_t^j
&=\frac{x_{i+1}^k-x_i^k}{t_{i+1}-t_i}
\frac{x_{i+1}^j-x_i^j}{t_{i+1}-t_i}
\times\int_{t_i}^{t_{i+1}}(t-t_i)\,\mathrm dt
=\frac12(x_{i+1}^k-x_i^k)
(x_{i+1}^j-x_i^j).
\end{aligned}
\]
Thus the first-level terms and the diagonal second-level terms in the expansions of $\Phi_{t_i,t_{i+1}}^{X}(z)$ and $\Phi_{t_i,t_{i+1}}^{\bar X}(z)$ agree; their two mixed terms differ by $a_i(X)$ and $-a_i(X)$. Subtracting the two expansions gives
\[
\begin{aligned}
\Phi_{t_i,t_{i+1}}^{X}(z)-\Phi_{t_i,t_{i+1}}^{\bar X}(z)
={}&a_i(X)DV_2(z)V_1(z)
-a_i(X)DV_1(z)V_2(z)
+\mathcal R_X(z)-\mathcal R_{\bar X}(z)\\
={}&a_i(X)[V_1,V_2](z)
+\mathcal R_X(z)-\mathcal R_{\bar X}(z).
\end{aligned}
\]
Both $X$ and $\bar X$ have variation at most $L(t_{i+1}-t_i)$. Applying the already established error bound separately to them yields
\[
\begin{aligned}
\bigl\|\Phi_{t_i,t_{i+1}}^{X}(z)-\Phi_{t_i,t_{i+1}}^{\bar X}(z)
-a_i(X)[V_1,V_2](z)\bigr\|
&\le\|\mathcal R_X(z)\|
+\|\mathcal R_{\bar X}(z)\|\\
&\le\frac{M_0M_1^2+M_0^2M_2}{6}L^3(t_{i+1}-t_i)^3
+\frac{M_0M_1^2+M_0^2M_2}{6}L^3(t_{i+1}-t_i)^3\\
&=\frac{M_0M_1^2+M_0^2M_2}{3}L^3(t_{i+1}-t_i)^3.
\end{aligned}
\]
We therefore fix
\[
C_{\mathrm{loc}}:=\frac{M_0M_1^2+M_0^2M_2}{3}.
\]
This proves the lemma with a coefficient independent of the interval and the path.
\end{proof}

\subsection{Derivatives and transport along the reference path}
\label{app:state_flow_bounds}

To control propagation of a state perturbation along the reference chord, set $Z_t=\Phi_{t_i,t}^{\bar X}(z)$ on $[t_i,t_{i+1}]$. The derivatives $D_zZ_t$ and $D_z^2Z_t$ are taken with respect to the initial state $Z_{t_i}=z$. At $z=\bar z_i$ this is the reference trajectory, and the state satisfies
\[
\frac{\mathrm d}{\mathrm dt}Z_t
=\sum_{j=1}^2\frac{x_{i+1}^j-x_i^j}{t_{i+1}-t_i}V_j(Z_t).
\]
Differentiating with respect to $z$ gives the first variational equation and its initial condition:
\[
\begin{aligned}
\frac{\mathrm d}{\mathrm dt}D_zZ_t
=\sum_{j=1}^2\frac{x_{i+1}^j-x_i^j}{t_{i+1}-t_i}DV_j(Z_t)D_zZ_t, \quad
D_zZ_{t_i}&=I.
\end{aligned}
\]
Integrating this equation and applying $M_1$ gives
\[
\begin{aligned}
\|D_zZ_t\|
&\le1+M_1\frac{\|x_{i+1}-x_i\|}{t_{i+1}-t_i}\int_{t_i}^t\|D_zZ_\tau\|\,\mathrm d\tau.
\end{aligned}
\]
Integral Grönwall, applied to this displayed inequality with the constant coefficient $M_1\|x_{i+1}-x_i\|/(t_{i+1}-t_i)$, yields
\[
\begin{aligned}
\|D_zZ_t\|
&\le\exp\!\left(M_1\frac{\|x_{i+1}-x_i\|}{t_{i+1}-t_i}(t-t_i)\right),\\
\|D_z\Phi_{t_i,t_{i+1}}^{\bar X}(z)\|&\le e^{M_1\|x_{i+1}-x_i\|}.
\end{aligned}
\]
The last estimate also implies a bound linear in the length of the observation interval: since $e^r-1=\int_0^re^s\,\mathrm ds\le re^r$ for $r\ge0$ and $\|x_{i+1}-x_i\|\le L(t_{i+1}-t_i)\le LT$,
\[
\begin{aligned}
\|D_z\Phi_{t_i,t_{i+1}}^{\bar X}(z)\|
&\le1+M_1\|x_{i+1}-x_i\|e^{M_1\|x_{i+1}-x_i\|}\\
&\le1+M_1e^{M_1LT}\|x_{i+1}-x_i\|\\
&\le1+M_1e^{M_1LT}L(t_{i+1}-t_i).
\end{aligned}
\]

The second derivative must be controlled separately. For arbitrary unit vectors $u,v\in\mathbb R^m$, differentiating the first variational equation once more gives
\[
\begin{aligned}
&\frac{\mathrm d}{\mathrm dt}
D_z^2Z_t[u,v]
=\sum_{j=1}^2
\frac{x_{i+1}^j-x_i^j}{t_{i+1}-t_i}
\times DV_j(Z_t)
D_z^2Z_t[u,v]
+\sum_{j=1}^2
\frac{x_{i+1}^j-x_i^j}{t_{i+1}-t_i}
\times D^2V_j(Z_t)
[D_zZ_tu,
D_zZ_tv],\\
&D_z^2Z_{t_i}[u,v]=0.
\end{aligned}
\]
The homogeneous part of this equation is governed by the same coefficient matrix as the first variational equation. Its propagator from $\tau$ to $t$ has norm at most $\exp(M_1\|x_{i+1}-x_i\|(t-\tau)/(t_{i+1}-t_i))$. Variation of constants, $M_2$, and the first-derivative estimate give
\[
\begin{aligned}
\|D_z^2Z_t\|
&\le M_2\frac{\|x_{i+1}-x_i\|}{t_{i+1}-t_i}
\times\int_{t_i}^t
\exp\!\left(\frac{M_1\|x_{i+1}-x_i\|}{t_{i+1}-t_i}
(t-\tau)\right)
\times\|D_zZ_\tau\|^2\,\mathrm d\tau\\
&\le M_2\frac{\|x_{i+1}-x_i\|}{t_{i+1}-t_i}
\times\int_{t_i}^t
\exp\!\left(\frac{M_1\|x_{i+1}-x_i\|}{t_{i+1}-t_i}
(t-t_i+\tau-t_i)\right)\,\mathrm d\tau\\
&\le M_2\|x_{i+1}-x_i\|
\frac{t-t_i}{t_{i+1}-t_i}
\times\exp\!\left(2M_1\|x_{i+1}-x_i\|
\frac{t-t_i}{t_{i+1}-t_i}\right).
\end{aligned}
\]
The last inequality uses $t-\tau+2(\tau-t_i)=t-t_i+\tau-t_i\le2(t-t_i)$. In particular,
\[
\begin{aligned}
\|D_z^2\Phi_{t_i,t_{i+1}}^{\bar X}(z)\|
&\le M_2\|x_{i+1}-x_i\|e^{2M_1\|x_{i+1}-x_i\|}\\
&\le M_2e^{2M_1LT}L(t_{i+1}-t_i).
\end{aligned}
\]

For every $e\in\mathbb R^m$, the mean-value and integral Taylor formulas now imply
\[
\begin{aligned}
\Phi_{t_i,t_{i+1}}^{\bar X}(z+e)-\Phi_{t_i,t_{i+1}}^{\bar X}(z)
&=\int_0^1D_z\Phi_{t_i,t_{i+1}}^{\bar X}(z+\theta e)e\,\mathrm d\theta,\\
\|\Phi_{t_i,t_{i+1}}^{\bar X}(z+e)-\Phi_{t_i,t_{i+1}}^{\bar X}(z)\|
&\le e^{M_1\|x_{i+1}-x_i\|}\|e\|,\\
\Phi_{t_i,t_{i+1}}^{\bar X}(z+e)-\Phi_{t_i,t_{i+1}}^{\bar X}(z)-D_z\Phi_{t_i,t_{i+1}}^{\bar X}(z)e
&=\int_0^1(1-\theta)
D_z^2\Phi_{t_i,t_{i+1}}^{\bar X}(z+\theta e)[e,e]\,\mathrm d\theta,\\
\bigl\|\Phi_{t_i,t_{i+1}}^{\bar X}(z+e)-\Phi_{t_i,t_{i+1}}^{\bar X}(z)
-D_z\Phi_{t_i,t_{i+1}}^{\bar X}(z)e\bigr\|
&\le\frac{M_2e^{2M_1LT}}{2}
L(t_{i+1}-t_i)\|e\|^2.
\end{aligned}
\]

At the reference starting state, the definition from Section~\ref{sec:partial_geometry} is $J_i=D_z\Phi_{t_i,t_{i+1}}^{\bar X}(\bar z_i)$. Thus the transport matrices $P_i=J_{n-1}\cdots J_{i+1}$ are products of these derivatives. The first-derivative estimate controls the whole product:
\[
\begin{aligned}
\|P_i\|
\le\prod_{j=i+1}^{n-1}e^{M_1\|x_{j+1}-x_j\|}
&=\exp\!\left(M_1\sum_{j=i+1}^{n-1}\|x_{j+1}-x_j\|\right)\\
&\le\exp\!\left(M_1L\sum_{j=i+1}^{n-1}(t_{j+1}-t_j)\right)\\
&=e^{M_1L(T-t_{i+1})}\le e^{M_1LT}.
\end{aligned}
\]
The telescoping sum of time intervals is what makes this estimate uniform in their number.

\subsection{Global expansion and uniform remainder}
\label{app:state_global_expansion}

Fix any admissible completion $X\in\mathcal X_L(S)$ and set $e_i=Z_{t_i}(X)-\bar z_i$. Since the initial state is fixed, $e_0=0$. The one-interval lemma, applied with initial state $Z_{t_i}(X)=\bar z_i+e_i$, gives
\[
\begin{aligned}
Z_{t_{i+1}}(X)
&=\Phi_{t_i,t_{i+1}}^{\bar X}(\bar z_i+e_i)
+a_i(X)[V_1,V_2](\bar z_i+e_i)
+\mathcal R_i^{\mathrm{loc}},\\
\|\mathcal R_i^{\mathrm{loc}}\|
&\le C_{\mathrm{loc}}L^3(t_{i+1}-t_i)^3.
\end{aligned}
\]
Subtracting $\bar z_{i+1}=\Phi_{t_i,t_{i+1}}^{\bar X}(\bar z_i)$ yields the exact recurrence
\[
\begin{aligned}
e_{i+1}={}&\Phi_{t_i,t_{i+1}}^{\bar X}(\bar z_i+e_i)-\Phi_{t_i,t_{i+1}}^{\bar X}(\bar z_i)
+a_i(X)[V_1,V_2](\bar z_i+e_i)+\mathcal R_i^{\mathrm{loc}}.
\end{aligned}
\]
Before linearizing the recurrence, the field bounds give
\[
\begin{aligned}
\|[V_1,V_2](z)\|
&=\|DV_2(z)V_1(z)-DV_1(z)V_2(z)\|\\
&\le\|DV_2(z)\|\,\|V_1(z)\|+\|DV_1(z)\|\,\|V_2(z)\|\\
&\le2M_0M_1.
\end{aligned}
\]
The Dido maximum, Lemma~\ref{lem:app_dido_max}, implies $|a_i(X)|\le L^2(t_{i+1}-t_i)^2/(2\pi)$. Together with the flow Lipschitz bound, this gives
\[
\begin{aligned}
\|e_{i+1}\|
&\le e^{M_1\|x_{i+1}-x_i\|}\|e_i\|
+\frac{M_0M_1}{\pi}L^2(t_{i+1}-t_i)^2
+C_{\mathrm{loc}}L^3(t_{i+1}-t_i)^3\\
&\le e^{M_1\|x_{i+1}-x_i\|}\|e_i\|
+\left(\frac{M_0M_1}{\pi}+C_{\mathrm{loc}}LT\right)
\times L^2(t_{i+1}-t_i)^2.
\end{aligned}
\]
The last step uses $L(t_{i+1}-t_i)\le LT$ and is uniform over all observation gaps. Set
\[
C_{\mathrm{step}}:=\frac{M_0M_1}{\pi}+C_{\mathrm{loc}}LT.
\]
Since $e_0=0$, the first two iterations of the recurrence are
\[
\begin{aligned}
\|e_1\|&\le C_{\mathrm{step}}L^2(t_1-t_0)^2,\\
\|e_2\|&\le e^{M_1\|x_2-x_1\|}C_{\mathrm{step}}L^2(t_1-t_0)^2
+C_{\mathrm{step}}L^2(t_2-t_1)^2.
\end{aligned}
\]
Continuing the same substitution to time $t_k$ gives
\[
\begin{aligned}
\|e_k\|
&\le C_{\mathrm{step}}\sum_{i=0}^{k-1}L^2(t_{i+1}-t_i)^2
\prod_{j=i+1}^{k-1}e^{M_1\|x_{j+1}-x_j\|}.
\end{aligned}
\]
The product is bounded uniformly because its exponent is controlled by the total elapsed time. Feasibility and telescoping of the observation intervals give
\[
\begin{aligned}
\sum_{j=i+1}^{k-1}\|x_{j+1}-x_j\|
&\le L\sum_{j=i+1}^{k-1}(t_{j+1}-t_j)
=L(t_k-t_{i+1})\le LT,\\
\prod_{j=i+1}^{k-1}e^{M_1\|x_{j+1}-x_j\|}
&=\exp\!\left(M_1\sum_{j=i+1}^{k-1}\|x_{j+1}-x_j\|\right)
\le e^{M_1L(t_k-t_{i+1})}\le e^{M_1LT}.
\end{aligned}
\]
Substituting this estimate into the preceding sum yields
\[
\begin{aligned}
\|e_k\|
&\le e^{M_1LT}C_{\mathrm{step}}\sum_{i=0}^{k-1}L^2(t_{i+1}-t_i)^2
\le e^{M_1LT}C_{\mathrm{step}}S_2.
\end{aligned}
\]
Set
\[
E:=e^{M_1LT}C_{\mathrm{step}},\qquad
\max_{0\le i\le n}\|e_i\|\le ES_2.
\]
In particular, $E$ depends only on the fields, $L$, and $T$, and not on the number or location of observations.

The derivative of the Lie bracket in a direction $v$ is
\[
\begin{aligned}
D[V_1,V_2](z)[v]
=D^2V_2(z)[v,V_1(z)]
+DV_2(z)DV_1(z)v
-D^2V_1(z)[v,V_2(z)]
-DV_1(z)DV_2(z)v.
\end{aligned}
\]
Using $M_0,M_1,M_2$ on the four terms, then integrating the derivative along the segment between two states, yields
\[
\begin{aligned}
\|D[V_1,V_2](z)\|&\le2(M_0M_2+M_1^2),\\
[V_1,V_2](z)-[V_1,V_2](z')
&=\int_0^1D[V_1,V_2]\bigl(z'+\theta(z-z')\bigr)[z-z']\,\mathrm d\theta,\\
\|[V_1,V_2](z)-[V_1,V_2](z')\|&\le2(M_0M_2+M_1^2)\|z-z'\|.
\end{aligned}
\]
Taylor-expand the flow at $\bar z_i$ and subtract the bracket at the same point from its value at $\bar z_i+e_i$. The exact recurrence becomes
\[
e_{i+1}=J_ie_i+a_i(X)[V_1,V_2](\bar z_i)+\mathcal R_i,
\]
where $\mathcal R_i$ is the sum of exactly three differences:
\[
\begin{aligned}
\mathcal R_i
={}&\underbrace{\mathcal R_i^{\mathrm{loc}}}_{\text{local CDE expansion}}
+\underbrace{a_i(X)\bigl([V_1,V_2](\bar z_i+e_i)-[V_1,V_2](\bar z_i)\bigr)}_{\text{bracket variation}}
+\underbrace{\Phi_{t_i,t_{i+1}}^{\bar X}(\bar z_i+e_i)-\Phi_{t_i,t_{i+1}}^{\bar X}(\bar z_i)-J_ie_i}_{\text{flow Taylor remainder}}.
\end{aligned}
\]
The three terms satisfy, respectively,
\[
\begin{aligned}
\|\mathcal R_i^{\mathrm{loc}}\|
&\le C_{\mathrm{loc}}L^3(t_{i+1}-t_i)^3,\\
|a_i(X)|\,
\|[V_1,V_2](\bar z_i+e_i)
-[V_1,V_2](\bar z_i)\|
&\le\frac{M_0M_2+M_1^2}{\pi}
L^2(t_{i+1}-t_i)^2\|e_i\|,\\
\|\Phi_{t_i,t_{i+1}}^{\bar X}(\bar z_i+e_i)-\Phi_{t_i,t_{i+1}}^{\bar X}(\bar z_i)
-J_ie_i\|
&\le\frac{M_2e^{2M_1LT}}2
L(t_{i+1}-t_i)\|e_i\|^2.
\end{aligned}
\]
Sum these bounds over all intervals and substitute $\max_i\|e_i\|\le ES_2$. Since $\sum_iL(t_{i+1}-t_i)=LT$, the steps are
\[
\begin{aligned}
\sum_{i=0}^{n-1}\|\mathcal R_i\|
&\le C_{\mathrm{loc}}S_3
+\frac{M_0M_2+M_1^2}{\pi}S_2\max_i\|e_i\|
+\frac{M_2e^{2M_1LT}}2LT
\bigl(\max_i\|e_i\|\bigr)^2\\
&\le C_{\mathrm{loc}}S_3
+\frac{(M_0M_2+M_1^2)E}{\pi}S_2^2
+\frac{M_2e^{2M_1LT}}2LT\,E^2S_2^2.
\end{aligned}
\]

To unfold the transport, first use $e_0=0$ and write the first two steps of the linear recurrence:
\[
\begin{aligned}
e_1&=a_0(X)[V_1,V_2](\bar z_0)+\mathcal R_0,\\
e_2&=J_1a_0(X)[V_1,V_2](\bar z_0)+a_1(X)[V_1,V_2](\bar z_1)
+J_1\mathcal R_0+\mathcal R_1.
\end{aligned}
\]
The same substitution through the final interval gives
\[
\begin{aligned}
e_n
&=\sum_{i=0}^{n-1}a_i(X)J_{n-1}\cdots J_{i+1}[V_1,V_2](\bar z_i)
+\sum_{i=0}^{n-1}J_{n-1}\cdots J_{i+1}\mathcal R_i\\
&=\sum_{i=0}^{n-1}a_i(X)b_i+\sum_{i=0}^{n-1}P_i\mathcal R_i.
\end{aligned}
\]
Using $\|P_i\|\le e^{M_1LT}$ from the flow estimates, the difference between the exact terminal state and the leading area sum satisfies
\[
\begin{aligned}
\left\|Z_T(X)-\bar z_T
-\sum_{i=0}^{n-1}a_i(X)b_i\right\|
&\le\sum_{i=0}^{n-1}
\|P_i\|\,\|\mathcal R_i\|\\
&\le e^{M_1LT}\sum_{i=0}^{n-1}\|\mathcal R_i\|\\
&\le e^{M_1LT}C_{\mathrm{loc}}S_3
+e^{M_1LT}
\frac{(M_0M_2+M_1^2)E}{\pi}S_2^2
+e^{M_1LT}
\frac{M_2e^{2M_1LT}}2LT\,E^2S_2^2.
\end{aligned}
\]
Set
\[
\begin{aligned}
C_Z:=e^{M_1LT}\biggl(&C_{\mathrm{loc}}
+\frac{(M_0M_2+M_1^2)E}{\pi}
+\frac{M_2e^{2M_1LT}}2LT\,E^2\biggr).
\end{aligned}
\]
The resulting bound holds uniformly over all records $S$ with $\mathcal X_L(S)\ne\varnothing$ and all paths in these completion classes:
\begin{equation}
\label{eq:app_state_one_sided}
\left\|Z_T(X)-\bar z_T-\sum_{i=0}^{n-1}a_i(X)b_i\right\|\le C_Z\bigl(S_3+S_2^2\bigr).
\end{equation}
The definitions of $M_0,M_1,M_2$, $C_{\mathrm{loc}}$, $C_{\mathrm{step}}$, $E$, and $C_Z$ show that these bounds depend only on $V_1,V_2,L,T$, independently of the skeleton, its number of intervals, and the admissible completion.

\subsection{The two Hausdorff directions and diameter}
\label{app:state_hausdorff}

The pointwise estimate yields directed distance bounds between $\mathcal Z_T(S;\mathcal X_L)$ and $\bar z_T+K(S)$.

For every $X\in\mathcal X_L(S)$, the Dido bound gives $|a_i(X)|\le A_i^\star$ for all $i$. By the definition of $K(S)$, the point $\bar z_T+\sum_i a_i(X)b_i$ therefore belongs to $\bar z_T+K(S)$. Applying~\eqref{eq:app_state_one_sided} and taking the appropriate supremum and infimum gives the first directed Hausdorff bound:
\[
\sup_{z\in\mathcal Z_T(S;\mathcal X_L)}\inf_{y\in\bar z_T+K(S)}\|z-y\|\le C_Z(S_3+S_2^2).
\]

For the opposite direction, choose any $y\in\bar z_T+K(S)$. There exist $a_0,\ldots,a_{n-1}$ with $|a_i|\le A_i^\star$ and $y=\bar z_T+\sum_i a_ib_i$. The exact Dido fiber, Lemma~\ref{lem:app_dido_fiber}, provides on each $[t_i,t_{i+1}]$ a path with the prescribed endpoints, speed at most $L$, and signed area $a_i$. Concatenating these paths preserves continuity and the speed bound; the class $W^{1,\infty}$ does not require matching derivatives at the observation times. Hence the concatenation is an $X\in\mathcal X_L(S)$ satisfying $a_i(X)=a_i$ for every $i$. By~\eqref{eq:app_state_one_sided},
\[
\begin{aligned}
\|Z_T(X)-y\|
&=\left\|Z_T(X)-\bar z_T-\sum_{i=0}^{n-1}a_i(X)b_i\right\|
\le C_Z(S_3+S_2^2).
\end{aligned}
\]
Therefore the reverse directed bound is
\[
\sup_{y\in\bar z_T+K(S)}\inf_{z\in\mathcal Z_T(S;\mathcal X_L)}\|y-z\|\le C_Z(S_3+S_2^2).
\]
By definition, $d_{\mathrm H}$ is the maximum of these two directed distances. Their bounds prove Theorem~\ref{thm:partial_state_zonotope}.

We also verify the diameter formula used in Section~\ref{sec:partial_geometry}. Since $K(S)$ is centrally symmetric ($a_i$ and $-a_i$ are both feasible),
\[
\begin{aligned}
\operatorname{diam}K(S)
&=2\max_{v\in K(S)}\|v\|
=2\max_{\|u\|=1}\max_{|a_i|\le A_i^\star}\sum_{i=0}^{n-1}a_i\,u^\top b_i
=2\max_{\|u\|=1}\sum_{i=0}^{n-1}A_i^\star|u^\top b_i|.
\end{aligned}
\]
The final equality follows by selecting each $a_i$ to have the sign of $u^\top b_i$. To pass from the zonotope diameter to the true identified-set diameter, take $z,z'\in\mathcal Z_T(S;\mathcal X_L)$. For any $\eta>0$, the first directed Hausdorff estimate supplies $y,y'\in\bar z_T+K(S)$ with $\|z-y\|,\|z'-y'\|\le C_Z(S_3+S_2^2)+\eta$. Therefore
\[
\begin{aligned}
\|z-z'\|
&\le\|z-y\|+\|y-y'\|+\|y'-z'\|
\le\operatorname{diam}K(S)+2C_Z(S_3+S_2^2)+2\eta.
\end{aligned}
\]
Take the supremum over $z,z'$ and let $\eta\downarrow0$. Repeating the same argument with the two sets exchanged gives the reverse inequality. Consequently,
\[
\begin{aligned}
&\biggl|\operatorname{diam}
\mathcal Z_T(S;\mathcal X_L)
-2\max_{\|u\|=1}\sum_{i=0}^{n-1}
A_i^\star|u^\top b_i|\biggr|
\le2C_Z(S_3+S_2^2).
\end{aligned}
\]
Finally, set $\delta=\max_i(t_{i+1}-t_i)$. The identity $\sum_i(t_{i+1}-t_i)=T$ gives
\[
\begin{aligned}
S_2&=L^2\sum_{i=0}^{n-1}(t_{i+1}-t_i)^2
\le L^2\delta\sum_{i=0}^{n-1}(t_{i+1}-t_i)=L^2T\delta,\\
S_3&=L^3\sum_{i=0}^{n-1}(t_{i+1}-t_i)^3
\le L^3\delta^2\sum_{i=0}^{n-1}(t_{i+1}-t_i)=L^3T\delta^2.
\end{aligned}
\]
Thus the Hausdorff error is bounded by $C_Z(L^3T+L^4T^2)\delta^2$, with no dependence on the number of observation intervals. This is an upper bound on the approximation error.

\section{Proof of the Target-Specific Certificate}
\label{app:certificate_proof}

Fix the observed planar record $S$ and assume that the prescribed speed bound satisfies $\mathcal X_L(S)\ne\varnothing$. Let $E,C_Z$ be the uniform constants from Appendix~\ref{app:state_global_expansion}. The remaining notation is as in Theorem~\ref{thm:partial_target_certificate}.

\subsection{Scalar expansion and its remainder}
\label{app:certificate_scalar_expansion}

\begin{lemma}[Uniform scalar expansion]
\label{lem:app_scalar_expansion}
Under the assumptions of Theorem~\ref{thm:partial_target_certificate}, every $X\in\mathcal X_L(S)$ satisfies
\begin{equation}
\label{eq:app_scalar_pointwise}
\left|F(X)-F(\bar X)-\sum_{i=0}^{n-1}c_i a_i(X)\right|\le C_Y(S_3+S_2^2),
\end{equation}
where the constant can be chosen as
\[
C_Y:=\|\nabla g\|_\infty C_Z+\frac12\|D^2g\|_\infty E^2.
\]
It depends only on $V_1,V_2,g,L,T$.
\end{lemma}

\begin{proof}
The global state estimates established in Appendix~\ref{app:state_global_expansion} hold for every admissible completion and give
\[
\begin{aligned}
\|Z_T(X)-\bar z_T\|&\le ES_2,\\
\left\|Z_T(X)-\bar z_T-\sum_{i=0}^{n-1}a_i(X)b_i\right\|&\le C_Z(S_3+S_2^2).
\end{aligned}
\]
Taylor's formula along the segment from $\bar z_T$ to $Z_T(X)$ gives
\[
\begin{aligned}
&g(Z_T(X))-g(\bar z_T)
=\nabla g(\bar z_T)^\top (Z_T(X)-\bar z_T)\\
&\qquad+\int_0^1(1-\theta)\,D^2g\bigl(\bar z_T+\theta(Z_T(X)-\bar z_T)\bigr)
[Z_T(X)-\bar z_T,Z_T(X)-\bar z_T]\mathrm d\theta.
\end{aligned}
\]
Because $D^2g$ is bounded, the integral remainder satisfies
\[
\begin{aligned}
\bigl|g(Z_T(X))-g(\bar z_T)
-\nabla g(\bar z_T)^\top (Z_T(X)-\bar z_T)\bigr|
&\le\int_0^1(1-\theta)\,\|D^2g\|_\infty\|Z_T(X)-\bar z_T\|^2\,\mathrm d\theta\\
&=\frac12\|D^2g\|_\infty\|Z_T(X)-\bar z_T\|^2\\
&\le\frac12\|D^2g\|_\infty E^2S_2^2.
\end{aligned}
\]
The equality uses $\int_0^1(1-\theta)\,\mathrm d\theta=1/2$; the final inequality uses the state-deviation bound above.

Recall $F(X)=g(Z_T(X))$, $F(\bar X)=g(\bar z_T)$, and $c_i=\nabla g(\bar z_T)^\top b_i$. Subtract the proposed leading scalar correction and add and subtract the first-order Taylor term. This produces an exact decomposition into the readout remainder and the state-approximation remainder:
\[
\begin{aligned}
F(X)-F(\bar X)-\sum_{i=0}^{n-1}c_i a_i(X)
\;&=\bigl[g(Z_T(X))-g(\bar z_T)-\nabla g(\bar z_T)^\top(Z_T(X)-\bar z_T)\bigr]\\
&+\nabla g(\bar z_T)^\top\biggl[Z_T(X)-\bar z_T-\sum_{i=0}^{n-1}a_i(X)b_i\biggr].
\end{aligned}
\]
Bound the first bracket by the preceding Taylor estimate and the second by Cauchy--Schwarz and the state estimate:
\[
\begin{aligned}
\left|F(X)-F(\bar X)-\sum_{i=0}^{n-1}c_i a_i(X)\right|
&\le\frac12\|D^2g\|_\infty E^2 S_2^2
+\|\nabla g(\bar z_T)\|\left\|Z_T(X)-\bar z_T-\sum_{i=0}^{n-1}a_i(X)b_i\right\|\\
&\quad\le\frac12\|D^2g\|_\infty E^2 S_2^2
+\|\nabla g\|_\infty C_Z(S_3+S_2^2)\\
&\quad\le\left(\|\nabla g\|_\infty C_Z+\frac12\|D^2g\|_\infty E^2\right)(S_3+S_2^2).
\end{aligned}
\]
The last step uses $S_3\ge0$. This is exactly~\eqref{eq:app_scalar_pointwise} with the stated fixed constant $C_Y$.
\end{proof}

\subsection{Scalar projection of the area zonotope}
\label{app:certificate_projection}

The projection of $K(S)$ under $k\mapsto\nabla g(\bar z_T)^\top k$ is determined by the coefficients $c_i$. From~\eqref{eq:partial_state_zonotope},
\[
\begin{aligned}
&\bigl\{\nabla g(\bar z_T)^\top k:\ k\in K(S)\bigr\}
=\left\{\nabla g(\bar z_T)^\top\sum_{i=0}^{n-1}a_i b_i:\ |a_i|\le A_i^\star\right\}
=\left\{\sum_{i=0}^{n-1}c_i a_i:\ |a_i|\le A_i^\star\right\}.
\end{aligned}
\]
For any admissible choice of $a_i$, the triangle inequality gives
\[
\begin{aligned}
\left|\sum_{i=0}^{n-1}c_i a_i\right|
&\le\sum_{i=0}^{n-1}|c_i|\,|a_i|
\le\sum_{i=0}^{n-1}|c_i|A_i^\star=R_2(S).
\end{aligned}
\]
To see that no value in this interval is missing, first suppose $R_2(S)>0$. For any number $s\in[-R_2(S),R_2(S)]$, choose
\[
a_i=\frac{s}{R_2(S)}A_i^\star\operatorname{sgn}(c_i),
\qquad \operatorname{sgn}(0)=0.
\]
Every chosen area satisfies $|a_i|\le A_i^\star$. Substitution into the scalar sum yields
\[
\begin{aligned}
\sum_{i=0}^{n-1}c_i a_i
&=\frac{s}{R_2(S)}\sum_{i=0}^{n-1}c_i A_i^\star\operatorname{sgn}(c_i)
=\frac{s}{R_2(S)}\sum_{i=0}^{n-1}|c_i|A_i^\star
=s.
\end{aligned}
\]
If $R_2(S)=0$, every summand $|c_i|A_i^\star$ vanishes, so every admissible scalar sum equals zero. In either case, we have proved
\[
\left\{\sum_{i=0}^{n-1}c_i a_i:\ |a_i|\le A_i^\star\right\}=[-R_2(S),R_2(S)].
\]
The Dido fiber in Lemma~\ref{lem:app_dido_fiber} realizes each prescribed $a_i$ on its observation interval. Corollary~\ref{cor:app_independent_areas} then realizes all these areas simultaneously with a single driver $X\in\mathcal X_L(S)$. Consequently, every number in the projected interval is the leading scalar correction of an admissible completion.

\subsection{Both Hausdorff directions and the diameter}
\label{app:certificate_hausdorff}

We compare the actual scalar identified set $\mathcal Y_T(S;\mathcal X_L)$ with $[F(\bar X)-R_2(S),F(\bar X)+R_2(S)]$. The pointwise estimate~\eqref{eq:app_scalar_pointwise} and the exact area projection give both directions of the Hausdorff bound.

First, let $X\in\mathcal X_L(S)$. Since $|a_i(X)|\le A_i^\star$, the projection result gives
\[
F(\bar X)+\sum_{i=0}^{n-1}c_i a_i(X)\in[F(\bar X)-R_2(S),F(\bar X)+R_2(S)].
\]
Its distance from the true output $F(X)$ is at most $C_Y(S_3+S_2^2)$ by~\eqref{eq:app_scalar_pointwise}. Taking the infimum over the comparison interval and then the supremum over all admissible $X$ establishes the first directed bound:
\[
\begin{aligned}
&\sup_{y\in\mathcal Y_T(S;\mathcal X_L)}\ \inf_{v\in[F(\bar X)-R_2(S),\,F(\bar X)+R_2(S)]}|y-v|
\le C_Y(S_3+S_2^2).
\end{aligned}
\]
For the reverse direction, choose any $v\in[F(\bar X)-R_2(S),F(\bar X)+R_2(S)]$. By the preceding subsection, there are areas $a_i\in[-A_i^\star,A_i^\star]$ such that
\[
v=F(\bar X)+\sum_{i=0}^{n-1}c_i a_i.
\]
By Corollary~\ref{cor:app_independent_areas}, one admissible completion $X$ satisfies $a_i(X)=a_i$ for every interval. Applying~\eqref{eq:app_scalar_pointwise} to this completion gives
\[
\begin{aligned}
|v-F(X)|
&=\left|F(\bar X)+\sum_{i=0}^{n-1}c_i a_i(X)-F(X)\right|
\le C_Y(S_3+S_2^2).
\end{aligned}
\]
Taking the corresponding infimum over the actual identified set and supremum over the comparison interval establishes the reverse directed bound:
\[
\begin{aligned}
&\sup_{v\in[F(\bar X)-R_2(S),\,F(\bar X)+R_2(S)]}\ \inf_{y\in\mathcal Y_T(S;\mathcal X_L)}|v-y|
\le C_Y(S_3+S_2^2).
\end{aligned}
\]
The Hausdorff distance is the maximum of the two directed distances. These estimates prove~\eqref{eq:partial_target_hausdorff} with the explicit $C_Y$ from Lemma~\ref{lem:app_scalar_expansion}.

For the diameter, given $X,X'\in\mathcal X_L(S)$, the first directed estimate supplies points $v,v'$ in the comparison interval, each within $C_Y(S_3+S_2^2)$ of its corresponding output. The triangle inequality gives
\[
\begin{aligned}
|F(X)-F(X')|
&\le|F(X)-v|+|v-v'|+|v'-F(X')|
\le2R_2(S)+2C_Y(S_3+S_2^2).
\end{aligned}
\]
Taking the supremum over $X,X'$ proves the upper bound on $\operatorname{diam}\mathcal Y_T(S;\mathcal X_L)$. Conversely, apply the reverse directed estimate to the two endpoints $F(\bar X)+R_2(S)$ and $F(\bar X)-R_2(S)$. It provides admissible completions $X^+,X^-$ satisfying
\[
\begin{aligned}
|F(X^+)-[F(\bar X)+R_2(S)]|&\le C_Y(S_3+S_2^2),\\
|F(X^-)-[F(\bar X)-R_2(S)]|&\le C_Y(S_3+S_2^2).
\end{aligned}
\]
The reverse triangle inequality therefore gives
\[
\begin{aligned}
|F(X^+)-F(X^-)|
&\ge2R_2(S)-|F(X^+)-[F(\bar X)+R_2(S)]|
-|F(X^-)-[F(\bar X)-R_2(S)]|\\
&\ge2R_2(S)-2C_Y(S_3+S_2^2).
\end{aligned}
\]
The diameter of the actual identified set is at least this separation. Combining upper and lower bounds yields
\[
\left|\operatorname{diam}\mathcal Y_T(S;\mathcal X_L)-2R_2(S)\right|\le2C_Y(S_3+S_2^2).
\]
No lower bound on $R_2(S)$ is assumed; in particular, $R_2(S)$ can vanish for a target insensitive to all leading area directions. Finally, Appendix~\ref{app:state_hausdorff} establishes $S_3+S_2^2\le(L^3T+L^4T^2)\delta^2$ for $\delta=\max_i(t_{i+1}-t_i)$. Thus the target-set approximation error is $O(\delta^2)$ with a constant independent of the number of observation intervals.

\subsection{Adjoint computation of the target sensitivities}
\label{app:certificate_adjoint}

The definition $c_i=\nabla g(\bar z_T)^\top P_i[V_1,V_2](\bar z_i)$ involves the matrix product $P_i=J_{n-1}\cdots J_{i+1}$. The backward adjoint computes these products through the recursion in Section~\ref{subsec:partial_geometry_certificate}.

Starting at the final time, set $\lambda_n=\nabla g(\bar z_T)$ and define $\lambda_i=J_i^\top\lambda_{i+1}$ for $i=n-1,\ldots,0$. The first two backward substitutions give
\[
\begin{aligned}
\lambda_{n-1}&=J_{n-1}^\top\nabla g(\bar z_T),\\
\lambda_{n-2}&=J_{n-2}^\top J_{n-1}^\top\nabla g(\bar z_T).
\end{aligned}
\]
Repeating the same substitution until time $t_{i+1}$ gives
\[
\begin{aligned}
\lambda_{i+1}
&=J_{i+1}^\top\cdots J_{n-1}^\top\nabla g(\bar z_T)
=(J_{n-1}\cdots J_{i+1})^\top\nabla g(\bar z_T)
=P_i^\top\nabla g(\bar z_T).
\end{aligned}
\]
For $i=n-1$, the product is empty and equals the identity, consistently with $P_{n-1}=I$ and $\lambda_n=\nabla g(\bar z_T)$. Substitute the last identity into $c_i$ and transpose the scalar product:
\begin{equation}
\label{eq:app_certificate_adjoint}
\begin{aligned}
c_i
&=\nabla g(\bar z_T)^\top P_i[V_1,V_2](\bar z_i)
=(P_i^\top\nabla g(\bar z_T))^\top[V_1,V_2](\bar z_i)
=\lambda_{i+1}^\top[V_1,V_2](\bar z_i).
\end{aligned}
\end{equation}
The index $i+1$ is essential: the area of interval $i$ produces its leading state perturbation at the end of that interval, after which the perturbation is transported to $T$.

To compute the same recursion without explicitly forming $J_i$, use the reference trajectory $\bar z_t=Z_t(\bar X)$ already defined in Section~\ref{sec:partial_geometry}. On each original interval $[t_i,t_{i+1}]$, the reference driver has constant derivative
\[
\dot{\bar X}_t^j=\frac{x_{i+1}^j-x_i^j}{t_{i+1}-t_i},\qquad j=1,2.
\]
For a perturbation $v\in\mathbb R^m$ of the state at $t_i$, let $\xi_t$ be the corresponding first variation of the state along this reference segment. The variational equation from Appendix~\ref{app:state_flow_bounds} reads
\[
\begin{aligned}
\dot\xi_t
&=\sum_{j=1}^2DV_j(\bar z_t)\xi_t\,\frac{x_{i+1}^j-x_i^j}{t_{i+1}-t_i},
\quad\xi_{t_i}=v,\quad\xi_{t_{i+1}}=J_iv.
\end{aligned}
\]
Define the continuous adjoint $\lambda_t$ backward from $\lambda_T=\nabla g(\bar z_T)$ by
\[
\dot\lambda_t=-\sum_{j=1}^2\frac{x_{i+1}^j-x_i^j}{t_{i+1}-t_i}\,DV_j(\bar z_t)^\top\lambda_t,
\quad t\in(t_i,t_{i+1}).
\]
Both equations are interpreted on the original time intervals, with $\lambda_t$ continuous at the observation times. Differentiating the scalar pairing and substituting the two differential equations gives, almost everywhere on each interval,
\[
\begin{aligned}
\frac{\mathrm d}{\mathrm dt}(\lambda_t^\top\xi_t)
&=\dot\lambda_t^\top\xi_t+\lambda_t^\top\dot\xi_t\\
&=-\sum_{j=1}^2\frac{x_{i+1}^j-x_i^j}{t_{i+1}-t_i}\lambda_t^\top DV_j(\bar z_t)\xi_t
+\sum_{j=1}^2\frac{x_{i+1}^j-x_i^j}{t_{i+1}-t_i}\lambda_t^\top DV_j(\bar z_t)\xi_t\\
&=0.
\end{aligned}
\]
Consequently the pairing is constant from $t_i$ to $t_{i+1}$, and hence
\[
\begin{aligned}
\lambda_{t_i}^\top v
&=\lambda_{t_{i+1}}^\top\xi_{t_{i+1}}
=\lambda_{t_{i+1}}^\top J_i v
=(J_i^\top\lambda_{t_{i+1}})^\top v.
\end{aligned}
\]
Because $v\in\mathbb R^m$ is arbitrary, $\lambda_{t_i}=J_i^\top\lambda_{t_{i+1}}$. With the node notation $\lambda_i:=\lambda_{t_i}$, the continuous adjoint therefore reproduces exactly the discrete recursion $\lambda_i=J_i^\top\lambda_{i+1}$, while requiring no storage of the full matrices $P_i$ or $J_i$. Combining it with~\eqref{eq:app_certificate_adjoint} gives the practical formula
\[
R_2(S)=\sum_{i=0}^{n-1}A_i^\star\left|\lambda_{i+1}^\top[V_1,V_2](\bar z_i)\right|.
\]

The same transport and bracket estimates give a uniform bound needed in the minimax and regularity arguments:
\[
\begin{aligned}
|c_i|
&\le\|\nabla g\|_\infty\|P_i\|\,\|[V_1,V_2](\bar z_i)\|
\le2M_0M_1e^{M_1LT}\|\nabla g\|_\infty
=:C_{\mathrm{sens}}.
\end{aligned}
\]
The coefficient $C_{\mathrm{sens}}$ is independent of the skeleton and the interval index. It bounds the absolute sensitivity.

For a linear readout $g(z)=w^\top z+\beta$, the Hessian term is zero, so the same proof applies with $C_Y=\|w\|C_Z$. If $c_i=0$ on every observation interval, the leading radius vanishes, but the theorem yields only $\operatorname{diam}\mathcal Y_T(S;\mathcal X_L)\le2C_Y(S_3+S_2^2)$; it does not assert exact identifiability.

\section{Exact Minimax Characterization for the Area CDE}
\label{app:exact_minimax_area}

We prove Proposition~\ref{thm:minimax_area_exact} under the point-observation model of Section~\ref{sec:problem}. In particular, every admissible driver starts at $X_0=0$, the values at $0$ and $T$ are available to the policy, and at most $N-1$ additional times may be queried. All policies are deterministic, including their stopping decisions.

\subsection{Exact output of the area CDE}
\label{app:exact_area_output}

Let $Z_0=0$, $g(z)=z_3$, and consider the vector fields
\[
V_1(z)=\begin{pmatrix}1\\0\\-z_2/2\end{pmatrix},\qquad V_2(z)=\begin{pmatrix}0\\1\\z_1/2\end{pmatrix}.
\]

\begin{lemma}[Exact area representation]
\label{lem:app_exact_area_representation}
For every absolutely continuous planar driver $X$ with $X_0=0$, the resulting terminal scalar output satisfies
\begin{equation}
\label{eq:app_exact_area_output}
F(X)=\frac12\int_0^T\bigl(X_t^1\,\mathrm dX_t^2-X_t^2\,\mathrm dX_t^1\bigr).
\end{equation}
The Lie bracket has the constant value $[V_1,V_2]=e_3$, and $\nabla g^\top[V_1,V_2]=1$.
\end{lemma}

\begin{proof}
Writing out the three components of the CDE gives the exact differential equations
\[
\begin{aligned}
\mathrm dZ_t^1&=\mathrm dX_t^1,\qquad \mathrm dZ_t^2=\mathrm dX_t^2,\\
\mathrm dZ_t^3&=-\frac12Z_t^2\,\mathrm dX_t^1+\frac12Z_t^1\,\mathrm dX_t^2.
\end{aligned}
\]
Integrating the first two equations from $0$ to $t$, using $Z_0=0$ and $X_0=0$, yields $Z_t^1=X_t^1$ and $Z_t^2=X_t^2$. Substituting both equalities into the third equation and integrating from $0$ to $T$ gives
\[
\begin{aligned}
Z_T^3-Z_0^3&=\int_0^T\left(-\frac12Z_t^2\,\mathrm dX_t^1+\frac12Z_t^1\,\mathrm dX_t^2\right)
=\frac12\int_0^T\bigl(X_t^1\,\mathrm dX_t^2-X_t^2\,\mathrm dX_t^1\bigr).
\end{aligned}
\]
Since $Z_0^3=0$ and $g(Z_T)=Z_T^3$, this proves~\eqref{eq:app_exact_area_output}. For the bracket, the only nonzero derivatives of the fields are $\partial_{z_2}(V_1)^3=-1/2$ and $\partial_{z_1}(V_2)^3=1/2$. Therefore
\[
\begin{aligned}
DV_2(z)V_1(z)&=\begin{pmatrix}0\\0\\1/2\end{pmatrix},\\
DV_1(z)V_2(z)&=\begin{pmatrix}0\\0\\-1/2\end{pmatrix},\\
[V_1,V_2](z)&=DV_2(z)V_1(z)-DV_1(z)V_2(z)=\begin{pmatrix}0\\0\\1\end{pmatrix}.
\end{aligned}
\]
Finally, $\nabla g=e_3$, and hence $\nabla g^\top[V_1,V_2]=1$.
\end{proof}

The canonical fields and the linear readout are not globally bounded, unlike the fields assumed in Theorems~\ref{thm:minimax_speed} and~\ref{thm:minimax_acceleration}. The next lemma shows that the exact example is unchanged by imposing bounded smooth coefficients on the admissible reachable set.

\begin{lemma}[Bounded smooth realization]
\label{lem:app_exact_area_cutoff}
Fix $L,T>0$. There exist $\widetilde V_1,\widetilde V_2\in C_b^\infty(\mathbb R^3;\mathbb R^3)$ and $\widetilde g\in C_b^\infty(\mathbb R^3;\mathbb R)$ whose terminal outputs agree with the area CDE for every $X\in\mathcal X_L$.
\end{lemma}

\begin{proof}
For every $X\in\mathcal X_L$, the identities proved above give
\[
\begin{aligned}
\sqrt{(Z_t^1)^2+(Z_t^2)^2}&=\|X_t-X_0\|\le\int_0^t\|\dot X_s\|\,\mathrm ds\le LT,\\
|Z_t^3|&\le\frac12\int_0^t\|X_s-X_0\|\,\|\dot X_s\|\,\mathrm ds
\le\frac12(LT)\int_0^t\|\dot X_s\|\,\mathrm ds\le\frac12L^2T^2.
\end{aligned}
\]
Thus all states reached at all times, by any driver in $\mathcal X_L$, lie in the fixed compact set
\[
\left\{z\in\mathbb R^3:\sqrt{z_1^2+z_2^2}\le LT,\quad |z_3|\le\frac12L^2T^2\right\}.
\]
Choose $\chi\in C_c^\infty(\mathbb R^3)$ that equals one on an open neighborhood of this compact set, and set $\widetilde V_j(z)=\chi(z)V_j(z)$ for $j=1,2$ and $\widetilde g(z)=\chi(z)z_3$. These functions are smooth, compactly supported, and therefore globally bounded together with all derivatives. Along every original admissible trajectory they coincide with $V_j$ and $g$. Consequently the original trajectory also solves the modified CDE; uniqueness of solutions for the smooth bounded fields makes the trajectories coincide, and their terminal readouts agree.
\end{proof}

\subsection{The exact output fiber of a fixed skeleton}
\label{app:exact_area_skeleton}

Fix the observed planar record $S=\{(t_i,x_i)\}_{i=0}^n$ and assume $\mathcal X_L(S)\ne\varnothing$. Let $\bar X$ be its piecewise-linear interpolation, as in Section~\ref{sec:partial_geometry}. On each gap, the signed area $a_i(X)$ and its sharp bound $A_i^\star$ are those of Section~\ref{subsec:partial_geometry_state}. For this area CDE, the output fiber is an exact interval.

\begin{lemma}[Exact response interval]
\label{lem:app_exact_area_fiber}
For every observed record $S$ with $\mathcal X_L(S)\ne\varnothing$, the area CDE satisfies
\begin{equation}
\label{eq:app_exact_area_fiber}
\begin{aligned}
\mathcal Y_T(S;\mathcal X_L)=\biggl[&F(\bar X)-\sum_{i=0}^{n-1}A_i^\star,
F(\bar X)+\sum_{i=0}^{n-1}A_i^\star\biggr].
\end{aligned}
\end{equation}
In particular, $\operatorname{rad}\mathcal Y_T(S;\mathcal X_L)=\sum_{i=0}^{n-1}A_i^\star$.
\end{lemma}

\begin{proof}
Split the exact integral~\eqref{eq:app_exact_area_output} over the observation intervals. Because $X_0=x_0=0$, for each $t\in[t_i,t_{i+1}]$ we can write $X_t=(x_i-x_0)+(X_t-x_i)$. Substituting this identity before integrating gives
\[
\begin{aligned}
F(X)&=\frac12\sum_{i=0}^{n-1}\int_{t_i}^{t_{i+1}}(X_t-x_0)\wedge\mathrm dX_t\\
&=\frac12\sum_{i=0}^{n-1}\int_{t_i}^{t_{i+1}}(x_i-x_0)\wedge\mathrm dX_t
+\frac12\sum_{i=0}^{n-1}\int_{t_i}^{t_{i+1}}(X_t-x_i)\wedge\mathrm dX_t\\
&=\frac12\sum_{i=0}^{n-1}(x_i-x_0)\wedge(x_{i+1}-x_i)
+\sum_{i=0}^{n-1}a_i(X).
\end{aligned}
\]
Here $u\wedge v=u^1v^2-u^2v^1$, and the last equality uses $\int_{t_i}^{t_{i+1}}\mathrm dX_t=x_{i+1}-x_i$ and the definition of $a_i(X)$ in Section~\ref{subsec:partial_geometry_state}. For the piecewise-linear path $\bar X$, every $a_i(\bar X)=0$, so applying the same identity to $\bar X$ yields
\[
\begin{aligned}
F(\bar X)&=\frac12\sum_{i=0}^{n-1}(x_i-x_0)\wedge(x_{i+1}-x_i),\\
F(X)&=F(\bar X)+\sum_{i=0}^{n-1}a_i(X).
\end{aligned}
\]
By Lemma~\ref{lem:app_dido_fiber}, each signed area belongs to the full interval $[-A_i^\star,A_i^\star]$. By Corollary~\ref{cor:app_independent_areas}, every independent choice of these areas is attainable by one speed-bounded completion. Therefore the set of sums in the displayed identity is exactly $[-\sum_iA_i^\star,\sum_iA_i^\star]$: the bounds follow from the triangle inequality; conversely, if $\sum_iA_i^\star>0$, each value $y$ in this interval is attained by taking $a_i(X)=yA_i^\star/(\sum_jA_j^\star)$ for every $i$. If $\sum_iA_i^\star=0$, all areas vanish. Adding $F(\bar X)$ establishes~\eqref{eq:app_exact_area_fiber}. Its radius is half the difference of its two endpoints, namely $\sum_iA_i^\star$.
\end{proof}

\subsection{Lower bound for every adaptive policy}
\label{app:exact_area_lower}

Fix an arbitrary deterministic policy $\pi\in\Pi_N$. Choose any unit vector $u\in\mathbb R^2$ and run the policy on the straight reference driver
\[
\bar X_t=\frac{2L}{\pi}\,t\,u,\qquad 0\le t\le T.
\]
Its speed is $2L/\pi<L$, so $\bar X\in\mathcal X_L$. After the policy stops, sort the distinct times it observed, including the free endpoints, as $0=s_0<s_1<\cdots<s_k=T$, with $k\le N$. The reference chord on each interval $[s_i,s_{i+1}]$ satisfies
\[
\begin{aligned}
\|\bar X_{s_{i+1}}-\bar X_{s_i}\|&=\frac{2L}{\pi}(s_{i+1}-s_i),\\
\frac{\|\bar X_{s_{i+1}}-\bar X_{s_i}\|}{L(s_{i+1}-s_i)}&=\frac2\pi.
\end{aligned}
\]
The sharp Dido bound, Lemma~\ref{lem:app_dido_max}, is attained at this ratio. Thus every observed gap admits the full signed-area interval with radius
\[
A_i^\star=\frac{L^2(s_{i+1}-s_i)^2}{2\pi}.
\]

The extremizing paths can be chosen as the two semicircles with the observed chord as their diameter. Their common radius is $\frac12\|\bar X_{s_{i+1}}-\bar X_{s_i}\|=\frac{L(s_{i+1}-s_i)}{\pi}$. Their lengths are exactly $\pi\frac{L(s_{i+1}-s_i)}{\pi}=L(s_{i+1}-s_i)$, so traversing either semicircle at constant speed on the original interval $[s_i,s_{i+1}]$ gives speed $L$. Their respective signed areas relative to the chord are
\[
\pm\frac{\pi}{2}\left(\frac{L(s_{i+1}-s_i)}{\pi}\right)^2=\pm\frac{L^2(s_{i+1}-s_i)^2}{2\pi}.
\]
Choose the positive semicircle on every gap and concatenate them to obtain $X^+$; choose all negative semicircles to obtain $X^-$. The paths are continuous across the observation times, have speed at most $L$ almost everywhere, and begin at zero. Therefore $X^+,X^-\in\mathcal X_L$, and for each time $s_i$ observed on the reference run,
\[
X^+_{s_i}=X^-_{s_i}=\bar X_{s_i}.
\]

We verify that the paths also produce the same adaptive transcript. Before the first internal query, the policy has received the same two endpoint observations from $\bar X$, $X^+$, and $X^-$. Since its query rule is deterministic, the first queried time is the same for all three paths. That time belongs to the set $\{s_0,\ldots,s_k\}$ produced on the reference run, so all three observed values agree. If the first $j$ queried times and values agree, the rule selects the same $(j+1)$-st time, and its observed value again agrees because this time also occurs in the reference transcript. Induction gives the same entire query sequence. The deterministic stopping rule then makes the same decision, proving
\[
\mathcal O_\pi(X^+)=\mathcal O_\pi(X^-)=\mathcal O_\pi(\bar X).
\]
Consequently, both $X^+$ and $X^-$ belong to the information fiber $\mathcal F_\pi(\bar X;\mathcal X_L)$ of Section~\ref{sec:problem}.

The exact area formula and the choice of opposite signs now give
\[
\begin{aligned}
F(X^+)&=F(\bar X)+\sum_{i=0}^{k-1}\frac{L^2(s_{i+1}-s_i)^2}{2\pi},\quad
F(X^-)=F(\bar X)-\sum_{i=0}^{k-1}\frac{L^2(s_{i+1}-s_i)^2}{2\pi}.
\end{aligned}
\]
The radius of an output set containing these two values is at least half their difference. Hence
\[
\begin{aligned}
\operatorname{rad}F\bigl(\mathcal F_\pi(\bar X;\mathcal X_L)\bigr)&\ge\frac12|F(X^+)-F(X^-)|
=\frac{L^2}{2\pi}\sum_{i=0}^{k-1}(s_{i+1}-s_i)^2.
\end{aligned}
\]
By Cauchy--Schwarz and telescoping of the observation times,
\[
\begin{aligned}
\left(\sum_{i=0}^{k-1}(s_{i+1}-s_i)\right)^2&\le k\sum_{i=0}^{k-1}(s_{i+1}-s_i)^2,\\
\sum_{i=0}^{k-1}(s_{i+1}-s_i)&=s_k-s_0=T,\\
\sum_{i=0}^{k-1}(s_{i+1}-s_i)^2&\ge\frac{T^2}{k}\ge\frac{T^2}{N}.
\end{aligned}
\]
Combining the last two displays proves, for every deterministic policy $\pi\in\Pi_N$,
\[
\begin{aligned}
&\sup_{X\in\mathcal X_L}\operatorname{rad}F\bigl(\mathcal F_\pi(X;\mathcal X_L)\bigr)
\ge\operatorname{rad}F\bigl(\mathcal F_\pi(\bar X;\mathcal X_L)\bigr)
\ge\frac{L^2T^2}{2\pi N}.
\end{aligned}
\]
Taking the infimum over $\pi$ establishes the lower bound in Proposition~\ref{thm:minimax_area_exact}. No part of the argument assumes that the policy queries observation times in chronological order.

\subsection{Matching upper bound and exact optimality}
\label{app:exact_area_upper}

Consider the nonadaptive policy observing the uniform grid $t_i=iT/N$, $i=0,\ldots,N$. For an arbitrary $X\in\mathcal X_L$, let $S=\{(t_i,X_{t_i})\}_{i=0}^N$ be its observed skeleton. The corresponding information fiber is precisely $\mathcal X_L(S)$, because the policy's query times are fixed and its transcript records all the values at those times.

By Lemma~\ref{lem:app_exact_area_fiber}, the radius left by this skeleton is the sum of its maximal per-gap signed areas. The uniform Dido estimate~\eqref{eq:app_dido_uniform} yields, on every gap,
\[
\begin{aligned}
A_i^\star&=A_\star\!\bigl(\|X_{t_{i+1}}-X_{t_i}\|,
L(t_{i+1}-t_i)\bigr)
\le\frac{L^2(t_{i+1}-t_i)^2}{2\pi}
=\frac{L^2T^2}{2\pi N^2}.
\end{aligned}
\]
Summing over all $N$ intervals and applying the exact radius formula gives
\[
\begin{aligned}
\operatorname{rad}F\bigl(\mathcal F_\pi(X;\mathcal X_L)\bigr)&=\sum_{i=0}^{N-1}A_i^\star
\le\sum_{i=0}^{N-1}\frac{L^2T^2}{2\pi N^2}=\frac{L^2T^2}{2\pi N}.
\end{aligned}
\]
This bound holds for every $X\in\mathcal X_L$, so the worst-case radius of the uniform policy is at most $L^2T^2/(2\pi N)$. Combined with the lower bound for every deterministic adaptive policy, it gives the exact identity~\eqref{eq:minimax_area_exact} and proves that the uniform nonadaptive grid is minimax-optimal.

\section{General Bounded-Speed Minimax Bounds}
\label{app:minimax_speed_proof}

We prove Theorem~\ref{thm:minimax_speed} for the deterministic observation model of Section~\ref{sec:problem}. Throughout, $V_1,V_2\in C_b^4(\mathbb R^m;\mathbb R^m)$ and $g\in C_b^2(\mathbb R^m;\mathbb R)$. The constants $M_0,M_1,M_2$, $E$, and $C_Z$ retain their definitions in Appendix~\ref{app:identified_set_proof}; $C_Y$ is the fixed constant of Appendix~\ref{app:certificate_proof}. All these constants depend only on the fields, the readout where relevant, $L$, and $T$.

\subsection{A uniform-grid upper bound}
\label{app:minimax_speed_upper}

A uniform nonadaptive grid gives an $O(N^{-1})$ upper bound, which also controls the remainders in the lower-bound argument.

\begin{lemma}[Uniform-grid upper bound]
\label{lem:app_speed_uniform_upper}
For every integer $N\ge1$,
\[
\begin{aligned}
\mathfrak R_N(F;\mathcal X_L)
&\le\frac{C_{\mathrm{sens}}L^2T^2}{2\pi N}
+\frac{C_Y(L^3T^3+L^4T^4)}{N^2}.
\end{aligned}
\]
Here $C_{\mathrm{sens}}$ is the fixed sensitivity bound of Appendix~\ref{app:certificate_adjoint}. In particular, $\mathfrak R_N(F;\mathcal X_L)\le C_{\mathrm{speed}}/N$, where $C_{\mathrm{speed}}$ is explicitly collected in the proof and is independent of the unknown driver and budget.
\end{lemma}

\begin{proof}
Fix the nonadaptive times $t_i=iT/N$ for $i=0,\ldots,N$. For any $X\in\mathcal X_L$, let $S=\{(t_i,X_{t_i})\}_{i=0}^N$ be its observed skeleton and $\bar X$ the straight interpolation between its observations. Every driver producing the same transcript belongs to $\mathcal X_L(S)$, and conversely, because the query times are fixed. Thus the residual output set is $\mathcal Y_T(S;\mathcal X_L)$.

The one-sided scalar expansion~\eqref{eq:app_scalar_pointwise} gives, for every $\widetilde X\in\mathcal X_L(S)$,
\[
\begin{aligned}
|F(\widetilde X)-F(\bar X)|
&\le\sum_{i=0}^{N-1}|c_i|\,|a_i(\widetilde X)|
+C_Y(S_3+S_2^2).
\end{aligned}
\]
For the straight reference path, the transport matrices satisfy $\|P_i\|\le e^{M_1LT}$ by Appendix~\ref{app:state_flow_bounds}. Moreover, the definition of $M_0,M_1$ and the Lie-bracket convention imply
\[
\begin{aligned}
\|[V_1,V_2](z)\|
&\le\|DV_2(z)\|\,\|V_1(z)\|
+\|DV_1(z)\|\,\|V_2(z)\|
\le2M_0M_1.
\end{aligned}
\]
Substitution in $c_i=\nabla g(\bar z_T)^\top P_i[V_1,V_2](\bar z_i)$ therefore yields $|c_i|\le2M_0M_1\|\nabla g\|_\infty e^{M_1LT}=C_{\mathrm{sens}}$, with exactly the coefficient defined in Appendix~\ref{app:certificate_adjoint}. The Dido bound~\eqref{eq:app_dido_uniform} gives
\[
|a_i(\widetilde X)|\le A_i^\star\le\frac{L^2(t_{i+1}-t_i)^2}{2\pi}=\frac{L^2T^2}{2\pi N^2}.
\]
Finally, the uniform grid has
\[
\begin{aligned}
S_2&=\sum_{i=0}^{N-1}L^2(t_{i+1}-t_i)^2=\frac{L^2T^2}{N},\\
S_3&=\sum_{i=0}^{N-1}L^3(t_{i+1}-t_i)^3=\frac{L^3T^3}{N^2},\\
S_2^2&=\frac{L^4T^4}{N^2}.
\end{aligned}
\]
Inserting the last three bounds into the scalar expansion, and summing the $N$ area contributions, proves
\[
\begin{aligned}
|F(\widetilde X)-F(\bar X)|
&\le C_{\mathrm{sens}}N\frac{L^2T^2}{2\pi N^2}
+\frac{C_Y(L^3T^3+L^4T^4)}{N^2}\\
&=\frac{C_{\mathrm{sens}}L^2T^2}{2\pi N}
+\frac{C_Y(L^3T^3+L^4T^4)}{N^2}\\
&\le\frac1N\left[\frac{C_{\mathrm{sens}}L^2T^2}{2\pi}
+C_Y(L^3T^3+L^4T^4)\right].
\end{aligned}
\]
The last step uses $N\ge1$. We fix the combined coefficient as
\[
C_{\mathrm{speed}}:=\frac{C_{\mathrm{sens}}L^2T^2}{2\pi}
+C_Y(L^3T^3+L^4T^4).
\]
Any two outputs in $\mathcal Y_T(S;\mathcal X_L)$ differ by at most twice the right-hand side, so the radius of this set is at most that right-hand side. The bound holds for every observed $S$; taking the supremum over $X$ and then the infimum over observation policies proves the lemma.
\end{proof}

\subsection{Localization and the bracket-sensitivity coefficient}
\label{app:minimax_speed_localization}

Fix the reference path $\bar X_t=vtu$ and interval $I=[a,b]\subset(0,T)$ from Section~\ref{subsec:minimax_general}; thus $0<v<L$ and $\beta_{v,u}(t)\ge\beta_0>0$ on $I$. Its state $\bar z_t$ and adjoint $\lambda_t$ satisfy the equations in Section~\ref{subsec:minimax_general}. In particular, the reference is the same straight path before, during, and after $I$.

For any $r\in[0,T]$, the already defined flow $\Phi_{r,T}^{\bar X}(z)$ follows the reference driver from state $z$ at time $r$ to time $T$. If a candidate driver agrees with $\bar X$ from time $r$ onward, then its final output is $g(\Phi_{r,T}^{\bar X}(Z_r(X)))$. This is an exact consequence of uniqueness of the CDE solution.

The first and second variational estimates in Appendix~\ref{app:state_flow_bounds}, applied to the reference suffix, give uniformly for $0\le r\le T$ and $z\in\mathbb R^m$,
\[
\begin{aligned}
\|D_z\Phi_{r,T}^{\bar X}(z)\|&\le e^{M_1v(T-r)}\le e^{M_1LT},\\
\|D_z^2\Phi_{r,T}^{\bar X}(z)\|&\le M_2v(T-r)e^{2M_1v(T-r)}
\le M_2LT e^{2M_1LT}.
\end{aligned}
\]
The chain rule, written for an arbitrary pair of unit state-space directions $w,w'$, gives
\[
\begin{aligned}
D(g\circ\Phi_{r,T}^{\bar X})(z)[w]
&= Dg(\Phi_{r,T}^{\bar X}(z))D_z\Phi_{r,T}^{\bar X}(z)w,\\
D^2(g\circ\Phi_{r,T}^{\bar X})(z)[w,w']
&=D^2g(\Phi_{r,T}^{\bar X}(z))\bigl[D_z\Phi_{r,T}^{\bar X}(z)w, D_z\Phi_{r,T}^{\bar X}(z)w'\bigr]
+Dg(\Phi_{r,T}^{\bar X}(z))D_z^2\Phi_{r,T}^{\bar X}(z)[w,w'].
\end{aligned}
\]
Consequently, the first two derivatives of every effective readout $g\circ\Phi_{r,T}^{\bar X}$ are bounded by
\[
\begin{aligned}
\|D(g\circ\Phi_{r,T}^{\bar X})\|_\infty
&\le\|\nabla g\|_\infty\|D_z\Phi_{r,T}^{\bar X}\|_\infty
\le e^{M_1LT}\|\nabla g\|_\infty,\\
\|D^2(g\circ\Phi_{r,T}^{\bar X})\|_\infty
&\le\|D^2g\|_\infty\|D_z\Phi_{r,T}^{\bar X}\|_\infty^2
+\|\nabla g\|_\infty\|D_z^2\Phi_{r,T}^{\bar X}\|_\infty\\
&\le e^{2M_1LT}\bigl(\|D^2g\|_\infty+M_2LT\|\nabla g\|_\infty\bigr).
\end{aligned}
\]
The following remainder coefficient is uniform over intervals contained in $I$ and their partitions:
\[
\begin{aligned}
C_I:={}&e^{M_1LT}\|\nabla g\|_\infty C_Z
+\frac12 e^{2M_1LT}\bigl(\|D^2g\|_\infty+M_2LT\|\nabla g\|_\infty\bigr)E^2.
\end{aligned}
\]
Here $E$ and $C_Z$ are the constants already fixed in Appendix B, evaluated with the original horizon $T$; the same bounds remain valid on any shorter interval. Applying the scalar expansion of Appendix C on $[a,b]$ with readout $g\circ\Phi_{b,T}^{\bar X}$ therefore gives, for every driver $X$ equal to $\bar X$ outside $I$ and matching it at the chosen observation times inside $I$,
\begin{equation}
\label{eq:app_speed_local_expansion}
\begin{aligned}
&\Bigl|F(X)-F(\bar X)-\sum_{i=0}^{n-1}c_i a_i(X)\Bigr|\le C_I L^3\sum_{i=0}^{n-1}(t_{i+1}-t_i)^3+C_I L^4\biggl(\sum_{i=0}^{n-1}(t_{i+1}-t_i)^2\biggr)^2,
\end{aligned}
\end{equation}
where $a=t_0<t_1<\cdots<t_n=b$ is any partition of $I$, $a_i(X)$ is the area relative to the corresponding chord, and $c_i$ is the target-specific coefficient defined for this localized CDE and its effective readout. The endpoints $a,b$ need not be actual policy queries. Since the candidate and reference paths have the same state at $a$ and share the driver after $b$, the localized scalar output is exactly the original $F(X)$.

To relate $c_i$ to the global adjoint, differentiate the reference flow from $t_{i+1}$ to $T$:
\[
\begin{aligned}
D_z\Phi_{t_{i+1},T}^{\bar X}(\bar z_{t_{i+1}})
&=D_z\Phi_{b,T}^{\bar X}(\bar z_b)\,J_{n-1}\cdots J_{i+1},\\
\lambda_{t_{i+1}}
&=D_z\Phi_{t_{i+1},T}^{\bar X}(\bar z_{t_{i+1}})^\top\nabla g(\bar z_T).
\end{aligned}
\]
To verify the second identity, differentiate the flow with respect to its starting time along the reference solution. The flow property gives
\[
\begin{aligned}
\frac{\mathrm d}{\mathrm dt}D_z\Phi_{t,T}^{\bar X}(\bar z_t)&=-vD_z\Phi_{t,T}^{\bar X}(\bar z_t)
\times D(u_1V_1+u_2V_2)(\bar z_t),\\
\frac{\mathrm d}{\mathrm dt}\bigl[D_z\Phi_{t,T}^{\bar X}(\bar z_t)^\top\nabla g(\bar z_T)\bigr]&=-vD(u_1V_1+u_2V_2)(\bar z_t)^\top\times D_z\Phi_{t,T}^{\bar X}(\bar z_t)^\top\nabla g(\bar z_T).
\end{aligned}
\]
At $t=T$, the flow derivative is the identity, so the last vector equals $\nabla g(\bar z_T)$. It therefore solves the same terminal-value problem as $\lambda_t$, establishing the identity. Taking transposes in the first displayed formula for the flow and using the definition of the localized coefficient now gives
\[
\begin{aligned}
c_i
&=\nabla g(\bar z_T)^\top D_z\Phi_{b,T}^{\bar X}(\bar z_b)\,J_{n-1}\cdots J_{i+1}[V_1,V_2](\bar z_{t_i})
=\lambda_{t_{i+1}}^\top[V_1,V_2](\bar z_{t_i}).
\end{aligned}
\]
For $i=n-1$, the matrix product is empty and equals the identity, as in Section~\ref{subsec:partial_geometry_state}.

By the adjoint equation, Grönwall's inequality from the terminal time yields
\[
\begin{aligned}
\|\lambda_t\|&\le\|\nabla g\|_\infty e^{M_1v(T-t)}\le\|\nabla g\|_\infty e^{M_1LT},\\
\|\dot\lambda_t\|&\le vM_1\|\lambda_t\|\le vM_1\|\nabla g\|_\infty e^{M_1LT}.
\end{aligned}
\]
Integrating the second inequality from $t_i$ to $t_{i+1}$, and using $\|[V_1,V_2]\|_\infty\le2M_0M_1$, we obtain
\[
\begin{aligned}
\bigl||c_i|-\beta_{v,u}(t_i)\bigr|
&\le\bigl|c_i-\lambda_{t_i}^\top[V_1,V_2](\bar z_{t_i})\bigr|\\
&\le\|\lambda_{t_{i+1}}-\lambda_{t_i}\|\,\|[V_1,V_2](\bar z_{t_i})\|\\
&\le2M_0M_1\int_{t_i}^{t_{i+1}}\|\dot\lambda_t\|\,\mathrm dt\\
&\le2vM_0M_1^2\|\nabla g\|_\infty e^{M_1LT}(t_{i+1}-t_i).
\end{aligned}
\]
Set
\[
H:=2vM_0M_1^2\|\nabla g\|_\infty e^{M_1LT}.
\]
Neither $C_I$ nor $H$ depends on the policy, its observation times, or the number of intervals.

\subsection{Excluding persistent unobserved gaps}
\label{app:minimax_speed_coarse}

Before applying the localized expansion to a sequence of policies, we must show that the intervals inside $I$ become small. A policy is free to concentrate all of its measurements elsewhere; the convergence of the maximal gap inside $I$ does not follow merely from the budget $N$.

\begin{lemma}[No persistent coarse gap]
\label{lem:app_speed_coarse_gap}
Let $\pi_N\in\Pi_N$ be deterministic policies such that $\sup_{X\in\mathcal X_L}\operatorname{rad}F(\mathcal F_{\pi_N}(X;\mathcal X_L))\to0$. Run each policy on $\bar X_t=vtu$. After its queries have been made, add $a,b$ to the query times and sort all times lying in $I$ as $a=t_0<\cdots<t_n=b$. Then
\[
\max_{0\le i<n}(t_{i+1}-t_i)\longrightarrow0.
\]
\end{lemma}

\begin{proof}
Suppose the claim is false. Along a subsequence, there is a gap $[t_i,t_{i+1}]\subset I$ with $t_{i+1}-t_i\ge\varepsilon>0$, containing no actual query time in its interior. Choose a positive length $\eta<\varepsilon/2$, fixed along this subsequence and small enough to satisfy
\[
H\eta\le\frac{\beta_0}{4},\qquad C_I(L^3\eta+L^4\eta^2)\le\frac{A_\star(v,L)\beta_0}{4}.
\]
Such a choice is possible because $A_\star(v,L)>0$, $\beta_0>0$, and $H,C_I$ are fixed. Inside each coarse gap choose a closed interval $[r,r+\eta]$; its location may depend on the policy, but its length does not.

The reference displacement on this interval is $\bar X_{r+\eta}-\bar X_r=v\eta u$. The Dido fiber in Appendix~\ref{app:geometry_dido}, together with the homogeneity of the area bound, supplies two speed-bounded paths joining those endpoints whose signed areas are $+A_\star(v,L)\eta^2$ and $-A_\star(v,L)\eta^2$. Form $X^+$ and $X^-$ by using these paths on $[r,r+\eta]$ and retaining $\bar X$ everywhere else. Both are in $\mathcal X_L$. At every time actually queried on the reference run, both alternatives equal $\bar X$: the only modified interval is contained in a query-free gap.

This agreement persists adaptively. The initial endpoint observations agree. If the first $k$ queried times and their values agree with the reference transcript, determinism makes the next queried time identical. That time belongs to the reference transcript and lies outside $(r,r+\eta)$, where the three drivers agree. Induction proves that all later queries and the stopping decisions agree, so
\[
\mathcal O_{\pi_N}(X^+)=\mathcal O_{\pi_N}(X^-)=\mathcal O_{\pi_N}(\bar X).
\]
In particular, both outputs belong to the same information fiber of $\pi_N$ at $\bar X$.

Apply the localized expansion~\eqref{eq:app_speed_local_expansion} on the single original interval $[r,r+\eta]$, with the unchanged reference path before and after it. Let $c$ denote its one-gap coefficient. The expansion gives
\[
\begin{aligned}
F(X^+)&=F(\bar X)+c A_\star(v,L)\eta^2+R^+,\\
F(X^-)&=F(\bar X)-c A_\star(v,L)\eta^2+R^-,\\
|R^+|,\ |R^-|&\le C_I(L^3\eta^3+L^4\eta^4).
\end{aligned}
\]
The adjoint coefficient estimate applies equally to this one-gap interval: since $r\in I$ and $\beta_{v,u}(r)\ge\beta_0$, it yields $|c|\ge\beta_0-H\eta$. Consequently, the radius of the information fiber satisfies
\[
\begin{aligned}
&\operatorname{rad}F(\mathcal F_{\pi_N}(\bar X;\mathcal X_L))\\
&\quad\ge\frac12|F(X^+)-F(X^-)|\\
&\quad\ge A_\star(v,L)\eta^2|c|-C_I(L^3\eta^3+L^4\eta^4)\\
&\quad\ge\eta^2\bigl[A_\star(v,L)(\beta_0-H\eta)-C_I(L^3\eta+L^4\eta^2)\bigr]\\
&\quad\ge\frac12 A_\star(v,L)\beta_0\eta^2>0.
\end{aligned}
\]
The positive lower bound is independent of $N$, contradicting the assumed convergence of the worst-case radii to zero. Therefore the maximal gap inside $I$ tends to zero.
\end{proof}

\subsection{Proof of the general bounded-speed lower bound}
\label{app:minimax_speed_lower}

\begin{proof}[Proof of Theorem~\ref{thm:minimax_speed}]
For each $N$, choose a deterministic policy $\pi_N\in\Pi_N$ with
\[
\sup_{X\in\mathcal X_L}\operatorname{rad}F(\mathcal F_{\pi_N}(X;\mathcal X_L))\le\mathfrak R_N(F;\mathcal X_L)+N^{-2}.
\]
Such a policy exists by the definition of the infimum. Lemma~\ref{lem:app_speed_uniform_upper} makes the right-hand side $O(N^{-1})$, so Lemma~\ref{lem:app_speed_coarse_gap} applies.

Run $\pi_N$ on the straight reference $\bar X_t=vtu$. Include $a,b$ among the observation times for the purpose of constructing adversarial paths, whether or not the policy requested them, and sort the local times as $a=t_0<t_1<\cdots<t_n=b$. There are at most $N-1$ internal policy queries on all of $[0,T]$, and at most two additional times have been inserted. Therefore $n\le N+2$. Write the maximal length of the local intervals as
\[
\delta_N=\max_{0\le i<n}(t_{i+1}-t_i).
\]
By Lemma~\ref{lem:app_speed_coarse_gap}, $\delta_N\to0$.

Each chord of the reference path has displacement $v(t_{i+1}-t_i)u$. The homogeneity of the Dido bound gives
\[
\begin{aligned}
A_i^\star
&=A_\star\bigl(v(t_{i+1}-t_i),L(t_{i+1}-t_i)\bigr)=A_\star(v,L)(t_{i+1}-t_i)^2.
\end{aligned}
\]
For each $i$, use the full Dido fiber to choose two admissible segments with this chord and opposite signed areas $\pm A_i^\star\operatorname{sgn}(c_i)$, where $\operatorname{sgn}(0)=0$. The coefficients $c_i$ are those of the localized expansion~\eqref{eq:app_speed_local_expansion}. Concatenate the positive-choice segments into $X^+$ and the negative-choice segments into $X^-$, leaving both drivers equal to $\bar X$ outside $I$. The paths are continuous, piecewise Lipschitz, and have speed at most $L$ almost everywhere; therefore they belong to $\mathcal X_L$. They agree with $\bar X$ at all times queried by $\pi_N$, as well as at the inserted times $a,b$.

The same induction used in the proof of Lemma~\ref{lem:app_speed_coarse_gap} shows that the three paths produce exactly the same ordered adaptive transcript. Hence $X^+$ and $X^-$ belong to $\mathcal F_{\pi_N}(\bar X;\mathcal X_L)$. Applying~\eqref{eq:app_speed_local_expansion} to each path and using its chosen areas gives
\[
\begin{aligned}
F(X^+)&=F(\bar X)+\sum_{i=0}^{n-1}A_i^\star|c_i|+R^+,\\
F(X^-)&=F(\bar X)-\sum_{i=0}^{n-1}A_i^\star|c_i|+R^-,\\
|R^+|,\ |R^-|&\le C_I\biggl[L^3\sum_{i=0}^{n-1}(t_{i+1}-t_i)^3+L^4\biggl(\sum_{i=0}^{n-1}(t_{i+1}-t_i)^2\biggr)^2\biggr].
\end{aligned}
\]
Half the difference of these outputs is a lower bound for the radius of their common information fiber. Using the explicit expression for $A_i^\star$, we obtain
\begin{equation}
\label{eq:app_speed_radius_bound}
\begin{aligned}
\operatorname{rad}F(\mathcal F_{\pi_N}(\bar X;\mathcal X_L))&\ge A_\star(v,L)\sum_{i=0}^{n-1}|c_i|(t_{i+1}-t_i)^2
-C_I\biggl[L^3\sum_{i=0}^{n-1}(t_{i+1}-t_i)^3+L^4\biggl(\sum_{i=0}^{n-1}(t_{i+1}-t_i)^2\biggr)^2\biggr].
\end{aligned}
\end{equation}

The preceding coefficient estimate gives $|c_i|\ge\beta_{v,u}(t_i)-H(t_{i+1}-t_i)\ge\beta_0-H(t_{i+1}-t_i)$. Substitution into~\eqref{eq:app_speed_radius_bound} yields
\[
\begin{aligned}
\operatorname{rad}F(\mathcal F_{\pi_N}(\bar X;\mathcal X_L))&\ge A_\star(v,L)\beta_0\sum_{i=0}^{n-1}(t_{i+1}-t_i)^2\\
&-\bigl(A_\star(v,L)H+C_I L^3\bigr)\sum_{i=0}^{n-1}(t_{i+1}-t_i)^3\\
&-C_IL^4\biggl(\sum_{i=0}^{n-1}(t_{i+1}-t_i)^2\biggr)^2.
\end{aligned}
\]
Because every gap is at most $\delta_N$ and the lengths telescope to $b-a$, we have
\[
\begin{aligned}
\sum_{i=0}^{n-1}(t_{i+1}-t_i)^3
&\le\delta_N\sum_{i=0}^{n-1}(t_{i+1}-t_i)^2,\\
\sum_{i=0}^{n-1}(t_{i+1}-t_i)^2
&\le\delta_N\sum_{i=0}^{n-1}(t_{i+1}-t_i)=(b-a)\delta_N.
\end{aligned}
\]
Thus the preceding lower bound can be written without dropping the nonlinear term:
\[
\begin{aligned}
\operatorname{rad}F(\mathcal F_{\pi_N}(\bar X;\mathcal X_L))&\ge\Bigl[A_\star(v,L)\beta_0-\bigl(A_\star(v,L)H+C_I L^3+C_IL^4(b-a)\bigr)\delta_N\Bigr]
\times\sum_{i=0}^{n-1}(t_{i+1}-t_i)^2.
\end{aligned}
\]
Set
\[
C_{\mathrm{abs}}:=A_\star(v,L)H+C_IL^3+C_IL^4(b-a).
\]
Since $\delta_N\to0$, eventually $C_{\mathrm{abs}}\delta_N\le A_\star(v,L)\beta_0/2$. The bracket is then at least $A_\star(v,L)\beta_0/2$, and hence
\[
\frac{A_\star(v,L)\beta_0}{2}\sum_{i=0}^{n-1}(t_{i+1}-t_i)^2\le\operatorname{rad}F(\mathcal F_{\pi_N}(\bar X;\mathcal X_L)).
\]
The definition of the near-minimax policies and Lemma~\ref{lem:app_speed_uniform_upper} bound the right-hand side by $C_{\mathrm{speed}}/N+N^{-2}$. Dividing by the fixed positive factor gives
\[
\begin{aligned}
\sum_{i=0}^{n-1}(t_{i+1}-t_i)^2
&\le\frac{2}{A_\star(v,L)\beta_0}
\left(\frac{C_{\mathrm{speed}}}{N}+\frac1{N^2}\right)\\
&\le\frac{2(C_{\mathrm{speed}}+1)}{A_\star(v,L)\beta_0}\frac1N.
\end{aligned}
\]
For the remaining error bounds, denote this fixed coefficient by $C_{2,I}:=2(C_{\mathrm{speed}}+1)/(A_\star(v,L)\beta_0)$. Thus the squared-gap sum is bounded by $C_{2,I}/N$ and is $O(N^{-1})$, with no dependence of its coefficient on the policy.
Return to the two remainder terms in~\eqref{eq:app_speed_radius_bound}. They now satisfy, respectively,
\[
\begin{aligned}
\sum_{i=0}^{n-1}(t_{i+1}-t_i)^3
&\le\delta_N\sum_{i=0}^{n-1}(t_{i+1}-t_i)^2
\le\frac{C_{2,I}\delta_N}{N}=o(N^{-1}),\\
\biggl(\sum_{i=0}^{n-1}(t_{i+1}-t_i)^2\biggr)^2
&\le\frac{C_{2,I}^2}{N^2}=O(N^{-2})=o(N^{-1}).
\end{aligned}
\]
The error from replacing $|c_i|$ by $\beta_{v,u}(t_i)$ is also at most
\[
\begin{aligned}
&A_\star(v,L)\sum_{i=0}^{n-1}\bigl||c_i|-\beta_{v,u}(t_i)\bigr|(t_{i+1}-t_i)^2\le A_\star(v,L)H\sum_{i=0}^{n-1}(t_{i+1}-t_i)^3=o(N^{-1}).
\end{aligned}
\]
More explicitly, the total subtracted error after replacing $|c_i|$ by $\beta_{v,u}(t_i)$ is at most
\[
\frac{(A_\star(v,L)H+C_IL^3)C_{2,I}\delta_N}{N}
+\frac{C_IL^4C_{2,I}^2}{N^2}.
\]
After multiplication by $N$, both terms tend to zero because their coefficients are fixed and $\delta_N\to0$. Combining these estimates with~\eqref{eq:app_speed_radius_bound}, we have the asymptotic bound
\[
\begin{aligned}
&\operatorname{rad}F(\mathcal F_{\pi_N}(\bar X;\mathcal X_L))\ge A_\star(v,L)\sum_{i=0}^{n-1}\beta_{v,u}(t_i)(t_{i+1}-t_i)^2-o(N^{-1}).
\end{aligned}
\]

Apply Cauchy--Schwarz to the $n$ numbers $\sqrt{\beta_{v,u}(t_i)}(t_{i+1}-t_i)$:
\[
\begin{aligned}
\biggl(\sum_{i=0}^{n-1}\sqrt{\beta_{v,u}(t_i)}(t_{i+1}-t_i)\biggr)^2&\le n\sum_{i=0}^{n-1}\beta_{v,u}(t_i)(t_{i+1}-t_i)^2,\\
\sum_{i=0}^{n-1}\beta_{v,u}(t_i)(t_{i+1}-t_i)^2&\ge\frac1n\biggl(\sum_{i=0}^{n-1}\sqrt{\beta_{v,u}(t_i)}(t_{i+1}-t_i)\biggr)^2.
\end{aligned}
\]
Since $n\le N+2$, replacing $1/n$ by $1/(N+2)$ preserves the lower bound. The function $\sqrt{\beta_{v,u}}$ is continuous on $I$, because the reference flow, the adjoint, and the bracket are continuous and $\beta_{v,u}\ge\beta_0>0$. For each local interval,
\[
\begin{aligned}
&\left|\int_{t_i}^{t_{i+1}}\sqrt{\beta_{v,u}(t)}\,\mathrm dt-\sqrt{\beta_{v,u}(t_i)}(t_{i+1}-t_i)\right|\le\int_{t_i}^{t_{i+1}}\left|\sqrt{\beta_{v,u}(t)}-\sqrt{\beta_{v,u}(t_i)}\right|\,\mathrm dt.
\end{aligned}
\]
Uniform continuity on $I$ and $\delta_N\to0$ make the sum of the right-hand sides tend to zero, since their total integration length is $b-a$. Thus
\[
\sum_{i=0}^{n-1}\sqrt{\beta_{v,u}(t_i)}(t_{i+1}-t_i)\longrightarrow\int_a^b\sqrt{\beta_{v,u}(t)}\,\mathrm dt.
\]
Finally, combine the last three displays with the near-minimax inequality:
\[
\begin{aligned}
&\mathfrak R_N(F;\mathcal X_L)+N^{-2}\ge\operatorname{rad}F(\mathcal F_{\pi_N}(\bar X;\mathcal X_L))\ge\frac{A_\star(v,L)}{N+2}\times\biggl(\sum_{i=0}^{n-1}\sqrt{\beta_{v,u}(t_i)}(t_{i+1}-t_i)\biggr)^2-o(N^{-1}).
\end{aligned}
\]
Multiply by $N$ and take the lower limit. As $N/(N+2)\to1$, $N\cdot N^{-2}\to0$, and $N\cdot o(N^{-1})\to0$, this gives
\[
\liminf_{N\to\infty}N\mathfrak R_N(F;\mathcal X_L)\ge A_\star(v,L)\left(\int_a^b\sqrt{\beta_{v,u}(t)}\,\mathrm dt\right)^2,
\]
which is exactly~\eqref{eq:minimax_speed_lower}. The right-hand side is strictly positive because $A_\star(v,L)>0$ and $\beta_{v,u}\ge\beta_0$ on an interval of positive length. Together with the uniform-grid upper bound from Lemma~\ref{lem:app_speed_uniform_upper}, this proves $\mathfrak R_N(F;\mathcal X_L)=\Theta(N^{-1})$ and completes Theorem~\ref{thm:minimax_speed}.
\end{proof}

\section{Bounded-Acceleration Minimax Bounds}
\label{app:minimax_acceleration_proof}

We prove Theorem~\ref{thm:minimax_acceleration} for the deterministic observation model of Section~\ref{sec:problem}. The driving paths belong to $\mathcal X_{L,M}$, with $T,L,M>0$, and $V_1,V_2\in C_b^4(\mathbb R^m;\mathbb R^m)$ and $g\in C_b^3(\mathbb R^m;\mathbb R)$. Fix the straight driver $\bar X_t=vtu$, where $\|u\|=1$ and $0<v<L$. Let $\bar z_t=Z_t(\bar X)$ and $\lambda_t$ be its state and terminal adjoint, and suppose that on $I=[a,b]\subset(0,T)$ the sensitivity $\beta_{v,u}(t)=|\lambda_t^\top[V_1,V_2](\bar z_t)|$ satisfies $\beta_{v,u}(t)\ge\beta_0>0$. The bounds $M_0,M_1,M_2,C_Y,C_{\mathrm{sens}}$ are as defined above. All additional constants below depend only on these bounds and the reference data, not on the sampling partition, perturbation signs, policy, or budget $N$.

\subsection{A clamped perturbation on an observation gap}
\label{app:acceleration_clamped_bump}

The acceleration bound requires the perturbation and its first derivative to agree at neighboring observation times. We construct a scalar perturbation that satisfies both endpoint conditions and has the largest possible integral under the prescribed bound on its second derivative.

\begin{lemma}[Optimal clamped scalar perturbation]
\label{lem:app_acc_optimal_bump}
Fix $t_i<t_{i+1}$ and $M>0$. If $\eta\in W^{2,\infty}([t_i,t_{i+1}])$ satisfies $\eta(t_i)=\eta(t_{i+1})=\dot\eta(t_i)=\dot\eta(t_{i+1})=0$ and $\|\ddot\eta\|_{L^\infty}\le M$, then
\begin{equation}
\label{eq:app_acc_bump_optimum}
\left|\int_{t_i}^{t_{i+1}}\eta(t)\,\mathrm dt\right|\le\frac{M}{32}(t_{i+1}-t_i)^3.
\end{equation}
The positive maximum is attained by the nonnegative function $\eta_i$ defined on three consecutive pieces. On the first quarter, $t_i\le t\le t_i+(t_{i+1}-t_i)/4$, set
\[
\eta_i(t)=\frac M2(t-t_i)^2.
\]
On the middle half, from $t_i+(t_{i+1}-t_i)/4$ to $t_i+3(t_{i+1}-t_i)/4$, set
\[
\begin{aligned}
\eta_i(t)=M\biggl[&-\frac12(t-t_i)^2
+\frac{t_{i+1}-t_i}{2}(t-t_i)
-\frac{(t_{i+1}-t_i)^2}{16}\biggr].
\end{aligned}
\]
On the final quarter, $t_i+3(t_{i+1}-t_i)/4\le t\le t_{i+1}$, set
\[
\eta_i(t)=\frac M2(t_{i+1}-t)^2.
\]
It also satisfies
\[
\begin{aligned}
\int_{t_i}^{t_{i+1}}\eta_i(t)\,\mathrm dt
=\frac{M}{32}(t_{i+1}-t_i)^3, \quad
\|\dot\eta_i\|_{L^\infty}
=\frac{M}{4}(t_{i+1}-t_i), \quad
\int_{t_i}^{t_{i+1}}|\dot\eta_i(t)|\,\mathrm dt
=\frac{M}{8}(t_{i+1}-t_i)^2.
\end{aligned}
\]
\end{lemma}

\begin{proof}
Integrate the second derivative twice, using the initial endpoint conditions. For $t\in[t_i,t_{i+1}]$,
\[
\begin{aligned}
\dot\eta(t)&=\int_{t_i}^{t}\ddot\eta(s)\,\mathrm ds,\\
\eta(t)&=\int_{t_i}^{t}\int_{t_i}^{\tau}\ddot\eta(s)\,\mathrm ds\,\mathrm d\tau
=\int_{t_i}^{t}(t-s)\ddot\eta(s)\,\mathrm ds.
\end{aligned}
\]
The last equality follows by integrating over $t_i\le s\le\tau\le t$ in the opposite order. Integrating once more and reversing the order on $t_i\le s\le t\le t_{i+1}$ yields
\[
\begin{aligned}
\int_{t_i}^{t_{i+1}}\eta(t)\,\mathrm dt
&=\int_{t_i}^{t_{i+1}}\int_{t_i}^{t}(t-s)\ddot\eta(s)
\mathrm ds\,\mathrm dt
=\int_{t_i}^{t_{i+1}}\int_s^{t_{i+1}}(t-s)\,\mathrm dt\,
\ddot\eta(s)\,\mathrm ds
=\frac12\int_{t_i}^{t_{i+1}}(t_{i+1}-s)^2
\ddot\eta(s)\,\mathrm ds.
\end{aligned}
\]
The two terminal conditions imply two moment identities. Indeed,
\[
\begin{aligned}
\int_{t_i}^{t_{i+1}}\ddot\eta(s)\,\mathrm ds
&=\dot\eta(t_{i+1})-\dot\eta(t_i)=0,\\
\int_{t_i}^{t_{i+1}}(t_{i+1}-s)\ddot\eta(s)\,\mathrm ds
&=\bigl[(t_{i+1}-s)\dot\eta(s)\bigr]_{t_i}^{t_{i+1}}+\int_{t_i}^{t_{i+1}}\dot\eta(s)\,\mathrm ds=\eta(t_{i+1})-\eta(t_i)=0.
\end{aligned}
\]
To exploit these constraints, consider the quadratic polynomial
\[
\begin{aligned}
p_i(s)=\frac12&\left(s-t_i-\frac{t_{i+1}-t_i}{4}\right)
\times\left(s-t_i-\frac{3(t_{i+1}-t_i)}4\right).
\end{aligned}
\]
Expanding both products gives the exact identity
\[
\begin{aligned}
\frac12(t_{i+1}-s)^2-p_i(s)
&=-\frac{3(t_{i+1}-t_i)^2}{32}+\frac{t_{i+1}-t_i}{2}(t_{i+1}-s).
\end{aligned}
\]
Its right-hand side integrates to zero against $\ddot\eta$ by the two preceding moment identities. Consequently,
\[
\begin{aligned}
\int_{t_i}^{t_{i+1}}\eta(t)\,\mathrm dt
&=\int_{t_i}^{t_{i+1}}p_i(s)\ddot\eta(s)\,\mathrm ds,\\
\left|\int_{t_i}^{t_{i+1}}\eta(t)\,\mathrm dt\right|
&\le M\int_{t_i}^{t_{i+1}}|p_i(s)|\,\mathrm ds.
\end{aligned}
\]
An antiderivative of $p_i$ is
\[
\begin{aligned}
&\frac{(s-t_i)^3}{6}
-\frac{t_{i+1}-t_i}{4}(s-t_i)^2+\frac{3(t_{i+1}-t_i)^2}{32}(s-t_i).
\end{aligned}
\]
Its values at $s=t_i$, $s=t_i+(t_{i+1}-t_i)/4$, $s=t_i+3(t_{i+1}-t_i)/4$, and $s=t_{i+1}$ are, respectively,
\[
0,\qquad\frac{(t_{i+1}-t_i)^3}{96},\qquad
0,\qquad\frac{(t_{i+1}-t_i)^3}{96}.
\]
The polynomial $p_i$ is positive on the first and last quarters and negative on the middle half. Splitting its absolute integral at these two roots gives
\[
\begin{aligned}
\int_{t_i}^{t_{i+1}}|p_i(s)|\,\mathrm ds
&=\frac{(t_{i+1}-t_i)^3}{96}
+\frac{(t_{i+1}-t_i)^3}{96}+\frac{(t_{i+1}-t_i)^3}{96}=\frac{(t_{i+1}-t_i)^3}{32}.
\end{aligned}
\]
This proves~\eqref{eq:app_acc_bump_optimum}.

Equality is attained when $\ddot\eta(s)=M\operatorname{sgn}p_i(s)$ almost everywhere. This second derivative equals $M$ on the first quarter, $-M$ on the middle half, and $M$ on the last quarter. Integrating twice with $\eta(t_i)=\dot\eta(t_i)=0$ yields exactly the three displayed pieces of $\eta_i$. At the two joining times, direct substitution shows that the values and first derivatives agree. At $t_{i+1}$, both are zero, so $\eta_i\in W^{2,\infty}$ and satisfies all four clamped endpoint conditions.

The polynomial $\eta_i$ increases until the midpoint and decreases afterwards. Its midpoint value, computed from the middle piece, is
\[
\begin{aligned}
\eta_i\!\left(\frac{t_i+t_{i+1}}2\right)
&=M\biggl[-\frac{(t_{i+1}-t_i)^2}{8}+\frac{(t_{i+1}-t_i)^2}{4}
-\frac{(t_{i+1}-t_i)^2}{16}\biggr]=\frac{M}{16}(t_{i+1}-t_i)^2.
\end{aligned}
\]
Therefore $\int_{t_i}^{t_{i+1}}|\dot\eta_i(t)|\,\mathrm dt=2\eta_i((t_i+t_{i+1})/2)=M(t_{i+1}-t_i)^2/8$. Differentiating the three pieces gives $\|\dot\eta_i\|_{L^\infty}=M(t_{i+1}-t_i)/4$. Finally, the equality case in the integral estimate gives $\int_{t_i}^{t_{i+1}}\eta_i=M(t_{i+1}-t_i)^3/32$.
\end{proof}

\begin{lemma}[Globally admissible transverse perturbations]
\label{lem:app_acc_glue}
Consider a finite partition $a=t_0<t_1<\cdots<t_n=b$ of $I$. Choose a unit vector $\nu\perp u$ and signs $\sigma_i\in\{-1,1\}$. On $[t_i,t_{i+1}]$ define $q_t=\sigma_i\eta_i(t)\nu$, with $\eta_i$ from Lemma~\ref{lem:app_acc_optimal_bump}, and set $q_t=0$ outside $I$. Then $q\in W^{2,\infty}([0,T];\mathbb R^2)$, vanishes at all partition times, and
\[
\begin{aligned}
\|\dot q\|_{L^\infty}&\le\frac M4\max_i(t_{i+1}-t_i),\quad
\|\ddot q\|_{L^\infty}\le M, \quad
\operatorname{Var}(q)=\frac M8\sum_{i=0}^{n-1}(t_{i+1}-t_i)^2.
\end{aligned}
\]
If $v^2+(M\max_i(t_{i+1}-t_i)/4)^2\le L^2$, then $\bar X+q$ and $\bar X-q$ belong to $\mathcal X_{L,M}$. On each interval, their signed areas relative to the straight reference chord are equal in magnitude and opposite in sign, with
\[
|a_i(\bar X\pm q)|=\frac{vM}{32}(t_{i+1}-t_i)^3.
\]
\end{lemma}

\begin{proof}
At each partition time both $\eta_i$ and $\dot\eta_i$ vanish from the left and the right. Thus the piecewise function $q$ and its first derivative are continuous, including at $a,b$ where they are joined to zero. Its second derivative exists almost everywhere and is bounded by $M$. The characterization of $W^{2,\infty}$ by an absolutely continuous first derivative therefore gives the claimed global regularity. Since the intervals are disjoint except at endpoints, taking the maximum of their derivative bounds gives $\|\dot q\|_{L^\infty}\le (M/4)\max_i(t_{i+1}-t_i)$ and $\|\ddot q\|_{L^\infty}\le M$; adding their variations gives the stated exact formula.

Because $\nu\perp u$, differentiation of $\bar X_t\pm q_t$ gives the pointwise identity
\[
\begin{aligned}
\|\dot{\bar X}_t\pm\dot q_t\|^2
&=\|vu\pm\dot q_t\|^2=v^2+\|\dot q_t\|^2\le v^2+\frac{M^2}{16}\max_i(t_{i+1}-t_i)^2.
\end{aligned}
\]
Under the stated condition this is at most $L^2$. Also $\ddot{\bar X}_t=0$, so the acceleration norm is at most $M$. The endpoints remain $X_0=0$ and $X_T=vTu$.

On $[t_i,t_{i+1}]$, write $q_t=\sigma_i\eta_i(t)\nu$. The corresponding perturbed driver obeys $X_t-x_i=v(t-t_i)u+q_t$ and $\mathrm dX_t=vu\,\mathrm dt+\dot q_t\,\mathrm dt$. Expanding the signed-area integral gives
\[
\begin{aligned}
2a_i(X)
&=\int_{t_i}^{t_{i+1}}(X_t-x_i)\wedge\dot X_t\,\mathrm dt\\
&=\int_{t_i}^{t_{i+1}}\bigl[v(t-t_i)u\wedge\dot q_t
+q_t\wedge vu+q_t\wedge\dot q_t\bigr]\,\mathrm dt\\
&=v\sigma_i\det(u,\nu)
\int_{t_i}^{t_{i+1}}
\bigl[(t-t_i)\dot\eta_i(t)-\eta_i(t)\bigr]\,\mathrm dt.
\end{aligned}
\]
Here $q_t\wedge\dot q_t=0$ since both vectors are multiples of $\nu$. Integration by parts, using $\eta_i(t_i)=\eta_i(t_{i+1})=0$, gives
\[
\begin{aligned}
&\int_{t_i}^{t_{i+1}}(t-t_i)\dot\eta_i(t)\,\mathrm dt=\bigl[(t-t_i)\eta_i(t)\bigr]_{t_i}^{t_{i+1}}-\int_{t_i}^{t_{i+1}}\eta_i(t)\,\mathrm dt=-\int_{t_i}^{t_{i+1}}\eta_i(t)\,\mathrm dt.
\end{aligned}
\]
Substitution and Lemma~\ref{lem:app_acc_optimal_bump} yield $a_i(X)=-v\sigma_i\det(u,\nu)M(t_{i+1}-t_i)^3/32$. Reversing $q$ reverses this sign. This proves the area formula and completes the lemma.
\end{proof}

\subsection{The symmetric CDE variation and its remainder}
\label{app:acceleration_symmetric_variation}

A linear expansion of $F$ alone is insufficient here: its quadratic remainder could be as large as the leading $N^{-2}$ effect. Comparing opposite perturbations cancels the quadratic term, leaving a remainder cubic in the variation of the perturbation.

\begin{lemma}[Uniform symmetric variation]
\label{lem:app_acc_symmetric_variation}
Let $q$ be one of the globally glued perturbations from Lemma~\ref{lem:app_acc_glue} on any partition of $I=[a,b]$, allowing $q$ to be zero on any subset of the partition intervals, and set $Q=\operatorname{Var}(q)$. There exists an explicit constant $C_{\mathrm{sym}}$, depending only on $V_1,V_2,g,v,M,T$ and $I$, not on the partition or signs, such that
\begin{equation}
\label{eq:app_acc_symmetric_remainder}
\begin{aligned}
&\bigl|F(\bar X+q)-F(\bar X-q)
-2DF_{\bar X}[q]\bigr|\le\frac{C_{\mathrm{sym}}}{3}Q^3.
\end{aligned}
\end{equation}
\end{lemma}

\begin{proof}
For $|\varepsilon|\le1$ write $Z_t^\varepsilon:=Z_t(\bar X+\varepsilon q)$ for the same CDE solution evaluated at the perturbed driver, and set $f(\varepsilon)=g(Z_T^\varepsilon)$. Lemma~\ref{lem:app_acc_glue} and nonnegativity of the interval lengths give
\[
\begin{aligned}
Q&\le\frac M8\sum_{i=0}^{n-1}(t_{i+1}-t_i)^2\le\frac M8\left(\sum_{i=0}^{n-1}(t_{i+1}-t_i)\right)^2
=\frac M8(b-a)^2.
\end{aligned}
\]
Setting a perturbation to zero on a partition interval preserves the zero value and zero first derivative at both endpoints, so global gluing and the derivative bounds remain valid. If no interval is set to zero, the first inequality is an equality; otherwise the omitted intervals only decrease the total variation. Thus
\[
\int_0^T\|vu+\varepsilon\dot q_t\|\,\mathrm dt
\le vT+Q\le vT+\frac M8(b-a)^2.
\]
Set $B:=vT+M(b-a)^2/8$, which is independent of the partition.
For the third-order variation, extend the field bounds of Appendix~\ref{app:state_local_expansion} by setting
\[
M_3:=\sup_{z\in\mathbb R^m}
\left(\sum_{j=1}^2\|D^3V_j(z)\|^2\right)^{1/2},
\]
which is finite under $C_b^4$. Differentiation of the parameter-dependent Carath\'eodory ODE is legitimate through third order, because the fields have bounded derivatives and the total variation is uniformly bounded. For $k=1,2,3$ let $U_k^\varepsilon(t)=\partial_\varepsilon^k Z_t^\varepsilon$. These derivatives have zero initial conditions. Differentiating the ODE once, twice, and three times gives, respectively,
\[
\begin{aligned}
\dot U_1^\varepsilon
&=\sum_{j=1}^2DV_j(Z^\varepsilon)
U_1^\varepsilon\,(vu^j+\varepsilon\dot q^j)
+\sum_{j=1}^2V_j(Z^\varepsilon)\dot q^j,\\[2pt]
\dot U_2^\varepsilon
&=\sum_{j=1}^2DV_j(Z^\varepsilon)
U_2^\varepsilon\,(vu^j+\varepsilon\dot q^j)+\sum_{j=1}^2D^2V_j(Z^\varepsilon)
[U_1^\varepsilon,U_1^\varepsilon]
(vu^j+\varepsilon\dot q^j)+2\sum_{j=1}^2DV_j(Z^\varepsilon)
U_1^\varepsilon\,\dot q^j,\\[2pt]
\dot U_3^\varepsilon
&=\sum_{j=1}^2DV_j(Z^\varepsilon)
U_3^\varepsilon\,(vu^j+\varepsilon\dot q^j)+3\sum_{j=1}^2D^2V_j(Z^\varepsilon)
[U_1^\varepsilon,U_2^\varepsilon]
(vu^j+\varepsilon\dot q^j)+\sum_{j=1}^2D^3V_j(Z^\varepsilon)
[U_1^\varepsilon,U_1^\varepsilon,U_1^\varepsilon]
(vu^j+\varepsilon\dot q^j)\\
&\quad+3\sum_{j=1}^2DV_j(Z^\varepsilon)
U_2^\varepsilon\,\dot q^j+3\sum_{j=1}^2D^2V_j(Z^\varepsilon)
[U_1^\varepsilon,U_1^\varepsilon]\dot q^j.
\end{aligned}
\]
In all lines the fields and derivatives are evaluated at $Z_t^\varepsilon$ and the input derivatives at $t$. The homogeneous coefficient has norm at most $M_1\|vu+\varepsilon\dot q_t\|$. Integral Gr\"onwall therefore bounds its propagator between any two times by
\[
\exp\!\left(M_1\int_0^T\|vu+\varepsilon\dot q_t\|\,\mathrm dt\right)
\le\exp\!\left(M_1\left[vT+\frac M8(b-a)^2\right]\right)
=e^{M_1B}.
\]
This bound is uniform in the partition and in $|\varepsilon|\le1$.

For the first variation, variation of constants and the definition of $M_0$ yield
\[
\begin{aligned}
\|U_1^\varepsilon(t)\|
&\le e^{M_1B}M_0\int_0^t\|\dot q_s\|\,\mathrm ds
\le M_0e^{M_1B}Q.
\end{aligned}
\]
Set $K_1:=M_0e^{M_1B}$. The second variational equation gives
\[
\begin{aligned}
\|U_2^\varepsilon(t)\|
&\le e^{M_1B}\int_0^t\Bigl[
M_2\|U_1^\varepsilon(s)\|^2\|vu+\varepsilon\dot q_s\|
+2M_1\|U_1^\varepsilon(s)\|\,\|\dot q_s\|\Bigr]\,\mathrm ds\\
&\le e^{M_1B}\int_0^t\Bigl[
M_2K_1^2Q^2\|vu+\varepsilon\dot q_s\|
+2M_1K_1Q\|\dot q_s\|\Bigr]\,\mathrm ds\\
&\le e^{M_1B}\bigl(M_2BK_1^2+2M_1K_1\bigr)Q^2.
\end{aligned}
\]
Set $K_2:=e^{M_1B}(M_2BK_1^2+2M_1K_1)$. The four forcing terms in the third variation give
\[
\begin{aligned}
\|U_3^\varepsilon(t)\|
&\le e^{M_1B}\int_0^t\Bigl[
3M_2\|U_1^\varepsilon(s)\|\,\|U_2^\varepsilon(s)\|\,\|vu+\varepsilon\dot q_s\|
+M_3\|U_1^\varepsilon(s)\|^3\|vu+\varepsilon\dot q_s\|\\
&\hspace{48pt}+3M_1\|U_2^\varepsilon(s)\|\,\|\dot q_s\|
+3M_2\|U_1^\varepsilon(s)\|^2\|\dot q_s\|\Bigr]\,\mathrm ds\\
&\le e^{M_1B}\int_0^t\Bigl[
3M_2K_1K_2Q^3\|vu+\varepsilon\dot q_s\|
+M_3K_1^3Q^3\|vu+\varepsilon\dot q_s\|\\
&\hspace{48pt}+3M_1K_2Q^2\|\dot q_s\|
+3M_2K_1^2Q^2\|\dot q_s\|\Bigr]\,\mathrm ds\\
&\le e^{M_1B}\bigl(3M_2BK_1K_2+M_3BK_1^3
+3M_1K_2+3M_2K_1^2\bigr)Q^3.
\end{aligned}
\]
Set
\[
\begin{aligned}
K_3:=e^{M_1B}\bigl(&3M_2BK_1K_2+M_3BK_1^3
+3M_1K_2+3M_2K_1^2\bigr).
\end{aligned}
\]
Hence, uniformly in $\varepsilon$, $t$, the partition, and the choices of active intervals and signs,
\[
\|U_1^\varepsilon(t)\|\le K_1Q,\qquad
\|U_2^\varepsilon(t)\|\le K_2Q^2,\qquad
\|U_3^\varepsilon(t)\|\le K_3Q^3.
\]
By the chain rule, the third derivative of the scalar output is exactly
\[
\begin{aligned}
f'''(\varepsilon)
&=D^3g(Z_T^\varepsilon)
[U_1^\varepsilon,U_1^\varepsilon,U_1^\varepsilon]+3D^2g(Z_T^\varepsilon)
[U_1^\varepsilon,U_2^\varepsilon]
+Dg(Z_T^\varepsilon)U_3^\varepsilon.
\end{aligned}
\]
The preceding bounds therefore give
\[
\begin{aligned}
|f'''(\varepsilon)|
&\le\|D^3g\|_\infty\|U_1^\varepsilon(T)\|^3
+3\|D^2g\|_\infty\|U_1^\varepsilon(T)\|\,\|U_2^\varepsilon(T)\|
+\|Dg\|_\infty\|U_3^\varepsilon(T)\|\\
&\le\bigl(\|D^3g\|_\infty K_1^3
+3\|D^2g\|_\infty K_1K_2+\|Dg\|_\infty K_3\bigr)Q^3
=C_{\mathrm{sym}}Q^3,
\end{aligned}
\]
Set
\[
C_{\mathrm{sym}}:=\|D^3g\|_\infty K_1^3
+3\|D^2g\|_\infty K_1K_2+\|Dg\|_\infty K_3.
\]
Taylor's formula with an integral remainder at $\varepsilon=0$ gives
\[
\begin{aligned}
f(1)&=f(0)+f'(0)
+\int_0^1(1-s)f''(s)\,\mathrm ds,\\
f(-1)&=f(0)-f'(0)
+\int_0^1(1-s)f''(-s)\,\mathrm ds.
\end{aligned}
\]
Subtract the second identity from the first. The remaining second derivatives satisfy $f''(s)-f''(-s)=\int_{-s}^sf'''(r)\,\mathrm dr$. Consequently,
\[
\begin{aligned}
|f(1)-f(-1)-2f'(0)|&\le\int_0^1(1-s)
\int_{-s}^{s}|f'''(r)|\,\mathrm dr\,\mathrm ds\\
&\quad\le C_{\mathrm{sym}}Q^3
\int_0^12s(1-s)\,\mathrm ds\\
&\quad=C_{\mathrm{sym}}Q^3
\left[s^2-\frac{2s^3}{3}\right]_{0}^{1}\\
&\quad=\frac{C_{\mathrm{sym}}}{3}Q^3.
\end{aligned}
\]
Since $f(1)=F(\bar X+q)$, $f(-1)=F(\bar X-q)$ and $f'(0)=DF_{\bar X}[q]$, this is~\eqref{eq:app_acc_symmetric_remainder}.
\end{proof}

\subsection{The adjoint and Lie-bracket identity}
\label{app:acceleration_adjoint}

\begin{lemma}[First variation along the straight driver]
\label{lem:app_acc_adjoint_bracket}
Let $q_t=\eta(t)\nu$ for a fixed unit vector $\nu\perp u$, where $\eta\in W^{1,\infty}$ and $\eta(0)=\eta(T)=0$. With $\lambda_t$ defined in Section~\ref{subsec:minimax_general},
\begin{equation}
\label{eq:app_acc_adjoint_identity}
DF_{\bar X}[q]
=-v\det(u,\nu)\int_0^T\eta(t)
\lambda_t^\top[V_1,V_2](\bar z_t)\,\mathrm dt.
\end{equation}
\end{lemma}

\begin{proof}
Write $V_u=u_1V_1+u_2V_2$ and $V_\nu=\nu_1V_1+\nu_2V_2$. The unperturbed state satisfies $\dot{\bar z}_t=vV_u(\bar z_t)$ and the adjoint satisfies $\dot\lambda_t=-vDV_u(\bar z_t)^\top\lambda_t$, with $\lambda_T=\nabla g(\bar z_T)$. Differentiating the state equation for $\bar X+\varepsilon q$ with respect to $\varepsilon$ at zero shows that the first state variation $\xi_t$ obeys
\[
\dot\xi_t=vDV_u(\bar z_t)\xi_t
+V_\nu(\bar z_t)\dot\eta(t),
\qquad\xi_0=0.
\]
Differentiate the scalar product of the adjoint and first variation, and substitute these two differential equations:
\[
\begin{aligned}
\frac{\mathrm d}{\mathrm dt}(\lambda_t^\top\xi_t)
&=\dot\lambda_t^\top\xi_t
+\lambda_t^\top\dot\xi_t\\
&=-v\lambda_t^\top DV_u(\bar z_t)\xi_t
+v\lambda_t^\top DV_u(\bar z_t)\xi_t+\lambda_t^\top V_\nu(\bar z_t)\dot\eta(t)\\
&=\lambda_t^\top V_\nu(\bar z_t)\dot\eta(t).
\end{aligned}
\]
Integrate over $[0,T]$, using $\xi_0=0$ and $\lambda_T=\nabla g(\bar z_T)$, to obtain
\[
\begin{aligned}
DF_{\bar X}[q]
&=\nabla g(\bar z_T)^\top\xi_T=\int_0^T\lambda_t^\top V_\nu(\bar z_t)
\dot\eta(t)\,\mathrm dt.
\end{aligned}
\]
Integration by parts is valid because $t\mapsto\lambda_t^\top V_\nu(\bar z_t)$ is continuously differentiable and $\eta$ is absolutely continuous. The endpoint terms vanish, giving
\[
\begin{aligned}
DF_{\bar X}[q]
&=\bigl[\eta(t)\lambda_t^\top V_\nu(\bar z_t)\bigr]_0^T-\int_0^T\eta(t)
\frac{\mathrm d}{\mathrm dt}
\bigl[\lambda_t^\top V_\nu(\bar z_t)\bigr]\,\mathrm dt=-\int_0^T\eta(t)
\frac{\mathrm d}{\mathrm dt}
\bigl[\lambda_t^\top V_\nu(\bar z_t)\bigr]\,\mathrm dt.
\end{aligned}
\]
Using the adjoint and state equations once more, the derivative in the integrand is
\[
\begin{aligned}
\frac{\mathrm d}{\mathrm dt}
\bigl[\lambda_t^\top V_\nu(\bar z_t)\bigr]
&=-v\lambda_t^\top DV_u(\bar z_t)V_\nu(\bar z_t)+v\lambda_t^\top DV_\nu(\bar z_t)V_u(\bar z_t)=v\lambda_t^\top[V_u,V_\nu](\bar z_t).
\end{aligned}
\]
By bilinearity and antisymmetry of the bracket in Appendix~\ref{app:geometry},
\[
\begin{aligned}
[V_u,V_\nu]
&=u_1\nu_2[V_1,V_2]+u_2\nu_1[V_2,V_1]=(u_1\nu_2-u_2\nu_1)[V_1,V_2]=\det(u,\nu)[V_1,V_2].
\end{aligned}
\]
Substituting this identity into the integral proves~\eqref{eq:app_acc_adjoint_identity}.
\end{proof}

\subsection{A uniform upper bound and the absence of large gaps}
\label{app:acceleration_upper_coarse}

\begin{lemma}[Uniform-grid upper bound]
\label{lem:app_acc_uniform_upper}
For every $N\ge1$, a nonadaptive uniform grid with $N$ gaps satisfies
\[
\begin{aligned}
\mathfrak R_N(F;\mathcal X_{L,M})
&\le\frac{1}{N^2}\biggl[
C_{\mathrm{sens}}\left(\frac{LMT^3}{12}
+\frac{M^2T^4}{80N}\right)+C_Y(L^3T^3+L^4T^4)\biggr].
\end{aligned}
\]
In particular, $\mathfrak R_N(F;\mathcal X_{L,M})\le C_{\mathrm{upper}}N^{-2}$ for all $N\ge1$, where the fixed constant is
\[
\begin{aligned}
C_{\mathrm{upper}}
&:=C_{\mathrm{sens}}\left(\frac{LMT^3}{12}+\frac{M^2T^4}{80}\right)+C_Y(L^3T^3+L^4T^4).
\end{aligned}
\]
It is independent of the skeleton, driver, policy, and number of intervals.
\end{lemma}

\begin{proof}
Take $t_i=iT/N$ and observe an arbitrary driver $X\in\mathcal X_{L,M}$ at these times. Let $S$ be its skeleton and $\bar X$ its straight interpolation. Every driver $\widetilde X\in\mathcal X_{L,M}(S)$ also belongs to $\mathcal X_L(S)$, so the scalar expansion of Appendix~\ref{app:certificate_scalar_expansion} gives
\[
\begin{aligned}
|F(\widetilde X)-F(\bar X)|
&\le\sum_{i=0}^{N-1}|c_i|\,|a_i(\widetilde X)|
+C_Y(S_3+S_2^2).
\end{aligned}
\]
The transport estimates in Appendix~\ref{app:state_flow_bounds} and the bracket bound in Appendix~\ref{app:state_global_expansion} imply, uniformly in $i$ and $S$,
\[
\begin{aligned}
|c_i|
&=|\nabla g(\bar z_T)^\top P_i[V_1,V_2](\bar z_i)|\\
&\le\|\nabla g\|_\infty\|P_i\|\,\|[V_1,V_2](\bar z_i)\|\\
&\le2M_0M_1e^{M_1LT}\|\nabla g\|_\infty=C_{\mathrm{sens}}.
\end{aligned}
\]
For every gap, the acceleration-aware area bound of Proposition~\ref{prop:app_acc_area_bounds} and $\|x_{i+1}-x_i\|\le L(t_{i+1}-t_i)$ give
\[
\begin{aligned}
|a_i(\widetilde X)|
&\le\frac{\|x_{i+1}-x_i\|M}{12}(t_{i+1}-t_i)^2
+\frac{M^2}{80}(t_{i+1}-t_i)^4\le\frac{LM}{12}\left(\frac TN\right)^3
+\frac{M^2}{80}\left(\frac TN\right)^4.
\end{aligned}
\]
Summing all $N$ intervals gives
\[
\begin{aligned}
\sum_{i=0}^{N-1}|a_i(\widetilde X)|
&\le\frac{LMT^3}{12N^2}
+\frac{M^2T^4}{80N^3}.
\end{aligned}
\]
Moreover, the quantities defined in Section~\ref{sec:partial_geometry} satisfy, on this grid,
\[
\begin{aligned}
S_3&=\sum_{i=0}^{N-1}L^3(t_{i+1}-t_i)^3
=\frac{L^3T^3}{N^2},\quad
S_2^2=\left(\sum_{i=0}^{N-1}L^2(t_{i+1}-t_i)^2\right)^2
=\frac{L^4T^4}{N^2}.
\end{aligned}
\]
Substituting these expressions into the scalar bound gives
\[
\begin{aligned}
|F(\widetilde X)-F(\bar X)|
&\le C_{\mathrm{sens}}\left(\frac{LMT^3}{12N^2}+\frac{M^2T^4}{80N^3}\right)
+\frac{C_Y(L^3T^3+L^4T^4)}{N^2}\\
&\le\frac{1}{N^2}\left[C_{\mathrm{sens}}\left(\frac{LMT^3}{12}+\frac{M^2T^4}{80}\right)
+C_Y(L^3T^3+L^4T^4)\right]
=\frac{C_{\mathrm{upper}}}{N^2}.
\end{aligned}
\]
The second step uses $N\ge1$ and the fixed coefficient in the lemma. All elements of $\mathcal Y_T(S;\mathcal X_{L,M})$ therefore lie within this bound of the common number $F(\bar X)$. Their radius is therefore no larger than that bound. The grid is a valid deterministic policy, and the bound is uniform over its observed skeletons, proving the lemma.
\end{proof}

\begin{lemma}[Vanishing gaps on a sensitive interval]
\label{lem:app_acc_coarse_gap}
Let $(\pi_N)$ be deterministic policies such that $\sup_{X\in\mathcal X_{L,M}}\operatorname{rad}F(\mathcal F_{\pi_N}(X;\mathcal X_{L,M}))\to0$. Run each policy on $\bar X_t=vtu$, add the two analysis points $a,b$ to its resulting query set, and sort the points in $I=[a,b]$. If $\delta_N$ is the largest resulting gap within $I$, then $\delta_N\to0$.
\end{lemma}

\begin{proof}
Suppose instead that for infinitely many $N$ one of the augmented gaps in $I$ has length at least $\varepsilon>0$. Choose a length $h_0>0$ smaller than $\varepsilon/3$ and a closed interval $J\subset I$ of length $h_0$ inside such a gap. The interval $J$ contains no actual query time in its interior. Apply Lemma~\ref{lem:app_acc_optimal_bump} to $J$, set $q_t=\eta_J(t)\nu$ there and $q_t=0$ elsewhere, and choose $h_0$ so small that
\[
\begin{aligned}
v^2+\frac{M^2h_0^2}{16}&\le L^2,
\frac{C_{\mathrm{sym}}M^3h_0^3}{3072}
\le\frac{vM\beta_0}{64}.
\end{aligned}
\]
These requirements are compatible because $v<L$, $C_{\mathrm{sym}}$ is fixed in Lemma~\ref{lem:app_acc_symmetric_variation}, and $\beta_0>0$. The clamped endpoint conditions and the gluing argument of Lemma~\ref{lem:app_acc_glue} show that the zero extension of this single bump belongs to $W^{2,\infty}$. Only $J$ carries a perturbation, so its derivative bounds are $\|\dot q\|_{L^\infty}\le Mh_0/4$ and $\|\ddot q\|_{L^\infty}\le M$, regardless of the lengths of the other query gaps. Since $\nu\perp u$ and $\ddot{\bar X}=0$, the inequalities
\[
\begin{aligned}
\|\dot{\bar X}_t\pm\dot q_t\|^2
&=v^2+\|\dot q_t\|^2\le v^2+\frac{M^2h_0^2}{16}\le L^2,\\
\|\ddot{\bar X}_t\pm\ddot q_t\|&=\|\ddot q_t\|\le M
\end{aligned}
\]
prove that $\bar X\pm q$ are admissible. Since $q$ vanishes at every actual query time of the reference run and at $0,T$, determinism shows by induction that the policy makes the same sequence of queries, sees the same values, and makes the same stopping decision for $\bar X$, $\bar X+q$, and $\bar X-q$.

Because the continuous function $t\mapsto\lambda_t^\top[V_1,V_2](\bar z_t)$ has no zero on $I$, it has a fixed sign there. The adjoint identity and the nonnegativity of $\eta_J$ give
\[
\begin{aligned}
|DF_{\bar X}[q]|
&=v\left|\int_J\eta_J(t)
\lambda_t^\top[V_1,V_2](\bar z_t)\,\mathrm dt\right|\ge v\beta_0\int_J\eta_J(t)\,\mathrm dt=\frac{vM\beta_0h_0^3}{32}.
\end{aligned}
\]
The symmetric variation bound and $Q=Mh_0^2/8$ yield
\[
\begin{aligned}
&\frac12|F(\bar X+q)-F(\bar X-q)|\ge|DF_{\bar X}[q]|-\frac{C_{\mathrm{sym}}Q^3}{6}\ge\frac{vM\beta_0h_0^3}{32}
-\frac{C_{\mathrm{sym}}M^3h_0^6}{3072}\ge\frac{vM\beta_0h_0^3}{64}>0.
\end{aligned}
\]
The two outputs belong to the same information fiber of the reference transcript. Its radius must therefore be at least the last positive quantity, independently of $N$, contradicting the assumed vanishing worst-case radius. Hence $\delta_N\to0$.
\end{proof}

\subsection{Proof of the bounded-acceleration minimax theorem}
\label{app:acceleration_main_proof}

\begin{proof}[Proof of Theorem~\ref{thm:minimax_acceleration}]
The uniform-grid policy from Lemma~\ref{lem:app_acc_uniform_upper} gives $\mathfrak R_N(F;\mathcal X_{L,M})\le C_{\mathrm{upper}}N^{-2}$. For every $N$ choose a deterministic policy $\pi_N$ within $N^{-3}$ of the minimax infimum, so that
\[
\begin{aligned}
&\sup_{X\in\mathcal X_{L,M}}
\operatorname{rad}F(\mathcal F_{\pi_N}(X;\mathcal X_{L,M}))\le\mathfrak R_N(F;\mathcal X_{L,M})+N^{-3}\le\frac{C_{\mathrm{upper}}}{N^2}+\frac1{N^3}.
\end{aligned}
\]
Such a policy exists by the definition of the infimum. Run $\pi_N$ on the straight reference path $\bar X_t=vtu$. Add $a,b$ only to analyze the gaps within $I$, not as extra measurements available to the policy. Write the sorted, augmented local times as $a=t_0<t_1<\cdots<t_n=b$. Since the policy uses at most $N-1$ internal queries, the augmented partition has $n\le N+2$ gaps. By Lemma~\ref{lem:app_acc_coarse_gap},
\[
\delta_N:=\max_{0\le i<n}(t_{i+1}-t_i)\longrightarrow0.
\]

On every local gap choose the nonnegative clamped function $\eta_i$ of Lemma~\ref{lem:app_acc_optimal_bump}. Since $\lambda_t^\top[V_1,V_2](\bar z_t)$ has a constant sign on $I$, choose $\sigma_i\in\{-1,1\}$ to make the first-variation contributions have the same sign, and define $q_t=\sigma_i\eta_i(t)\nu$ on $[t_i,t_{i+1}]$, extending it by zero outside $I$. The bounds of Lemma~\ref{lem:app_acc_glue} give
\[
\begin{aligned}
\|\dot q\|_{L^\infty}&\le\frac{M\delta_N}{4},\quad
\|\ddot q\|_{L^\infty}\le M,\quad
Q:=\operatorname{Var}(q)=\frac M8
\sum_{i=0}^{n-1}(t_{i+1}-t_i)^2.
\end{aligned}
\]
Because $v<L$ and $\delta_N\to0$, the inequality $v^2+(M\delta_N/4)^2\le L^2$ holds for all sufficiently large $N$. Thus $X^+=\bar X+q$ and $X^-=\bar X-q$ are both in $\mathcal X_{L,M}$.

Every actual query time of the policy's reference transcript belongs to the augmented partition or lies outside $I$, so $q$ vanishes there. Initially $X^+$, $X^-$, and $\bar X$ also agree at the two free endpoints. Suppose the first $k-1$ queried times and their values agree. Determinism then forces the $k$th query time to be identical; $q$ vanishes at that time, so its observed value also agrees. Induction gives equality of the complete ordered transcripts, including the stopping decision:
\[
\mathcal O_{\pi_N}(X^+)
=\mathcal O_{\pi_N}(X^-)
=\mathcal O_{\pi_N}(\bar X).
\]
Consequently, the information fiber of the reference transcript contains both paths, and its output radius satisfies
\[
\begin{aligned}
&\operatorname{rad}F(\mathcal F_{\pi_N}(\bar X;\mathcal X_{L,M}))
\ge\frac12|F(\bar X+q)-F(\bar X-q)|.
\end{aligned}
\]

The adjoint identity and the positivity of the clamped bumps now give the leading term. Since the sensitivity is continuous and bounded away from zero on $I$, define on each existing local interval $\underline\beta_i=\min_{t\in[t_i,t_{i+1}]}\beta_{v,u}(t)\ge\beta_0$. Summing the signed contributions over the gaps gives
\[
\begin{aligned}
|DF_{\bar X}[q]|&=v\biggl|\sum_{i=0}^{n-1}\sigma_i\det(u,\nu)
\int_{t_i}^{t_{i+1}}\eta_i(t)\times\lambda_t^\top[V_1,V_2](\bar z_t)\,\mathrm dt\biggr|\\
&=v\sum_{i=0}^{n-1}
\int_{t_i}^{t_{i+1}}\eta_i(t)
\beta_{v,u}(t)\,\mathrm dt\\
&\ge v\sum_{i=0}^{n-1}\underline\beta_i
\int_{t_i}^{t_{i+1}}\eta_i(t)\,\mathrm dt\\
&=\frac{vM}{32}\sum_{i=0}^{n-1}
\underline\beta_i(t_{i+1}-t_i)^3.
\end{aligned}
\]
Here the second equality uses the fixed sign of the bracket sensitivity on $I$ and the chosen signs $\sigma_i$. Apply the symmetric variation lemma and the lower bound on the fiber radius. Using the exact expression for $Q$, we obtain
\begin{equation}
\label{eq:app_acc_preliminary_lower}
\begin{aligned}
&\operatorname{rad}F(\mathcal F_{\pi_N}(\bar X;\mathcal X_{L,M}))\ge\frac{vM}{32}\sum_{i=0}^{n-1}
\underline\beta_i(t_{i+1}-t_i)^3-\frac{C_{\mathrm{sym}}M^3}{3072}
\left(\sum_{i=0}^{n-1}(t_{i+1}-t_i)^2\right)^3.
\end{aligned}
\end{equation}

To compare the remainder with the leading gap sum before establishing its $N$-rate, apply Cauchy--Schwarz to the powers $1/2$ and $3/2$ of each interval length:
\[
\begin{aligned}
&\left(\sum_{i=0}^{n-1}(t_{i+1}-t_i)^2\right)^2\le\left(\sum_{i=0}^{n-1}(t_{i+1}-t_i)\right)
\left(\sum_{i=0}^{n-1}(t_{i+1}-t_i)^3\right)=(b-a)\sum_{i=0}^{n-1}(t_{i+1}-t_i)^3.
\end{aligned}
\]
Multiplying by the square root of the same bound shows
\[
\begin{aligned}
&\left(\sum_{i=0}^{n-1}(t_{i+1}-t_i)^2\right)^3\le(b-a)^{3/2}
\left(\sum_{i=0}^{n-1}(t_{i+1}-t_i)^3\right)^{3/2}.
\end{aligned}
\]
Moreover,
\[
\begin{aligned}
0\le\sum_{i=0}^{n-1}(t_{i+1}-t_i)^3
&\le\delta_N^2\sum_{i=0}^{n-1}(t_{i+1}-t_i)
=(b-a)\delta_N^2\longrightarrow0.
\end{aligned}
\]
Divide the cubic remainder bound by the positive sum $\sum_i(t_{i+1}-t_i)^3$ to obtain
\[
\begin{aligned}
&\frac{C_{\mathrm{sym}}M^3}{3072}
\frac{\left(\sum_i(t_{i+1}-t_i)^2\right)^3}
{\sum_i(t_{i+1}-t_i)^3}
\le\frac{C_{\mathrm{sym}}M^3}{3072}(b-a)^{3/2}
\left(\sum_i(t_{i+1}-t_i)^3\right)^{1/2}
\le\frac{C_{\mathrm{sym}}M^3}{3072}(b-a)^2\delta_N
\longrightarrow0.
\end{aligned}
\]
Since $\underline\beta_i\ge\beta_0$, eventually the last expression is at most $vM\beta_0/64$. The remainder is then no more than half the positive lower bound in~\eqref{eq:app_acc_preliminary_lower}. Explicitly,
\[
\begin{aligned}
&\operatorname{rad}F(\mathcal F_{\pi_N}(\bar X;\mathcal X_{L,M}))\ge\frac{vM\beta_0}{64}
\sum_{i=0}^{n-1}(t_{i+1}-t_i)^3.
\end{aligned}
\]
The near-minimax upper bound now implies
\[
\begin{aligned}
\sum_{i=0}^{n-1}(t_{i+1}-t_i)^3
&\le\frac{64}{vM\beta_0}
\left(\frac{C_{\mathrm{upper}}}{N^2}+\frac1{N^3}\right)
\le\frac{64(C_{\mathrm{upper}}+1)}{vM\beta_0}\frac1{N^2}.
\end{aligned}
\]
Set $C_{3,I}:=64(C_{\mathrm{upper}}+1)/(vM\beta_0)$. The cubic-gap sum is at most $C_{3,I}/N^2$. Cauchy--Schwarz then gives
\[
\begin{aligned}
\left(\sum_{i=0}^{n-1}(t_{i+1}-t_i)^2\right)^3
&\le\frac{(b-a)^{3/2}C_{3,I}^{3/2}}{N^3}
=O(N^{-3})=o(N^{-2}).
\end{aligned}
\]
In particular, the subtracted output remainder is at most $C_{\mathrm{sym}}M^3(b-a)^{3/2}C_{3,I}^{3/2}/(3072N^3)$; multiplying by $N^2$ leaves a fixed coefficient times $N^{-1}$, which tends to zero. Substituting this into~\eqref{eq:app_acc_preliminary_lower} proves
\[
\begin{aligned}
&\operatorname{rad}F(\mathcal F_{\pi_N}(\bar X;\mathcal X_{L,M}))\ge\frac{vM}{32}\sum_{i=0}^{n-1}
\underline\beta_i(t_{i+1}-t_i)^3-o(N^{-2}).
\end{aligned}
\]

It remains to pass from the weighted gap sum to a continuous lower constant. H\"older's inequality gives
\[
\begin{aligned}
&\sum_{i=0}^{n-1}\underline\beta_i^{1/3}(t_{i+1}-t_i)\le\left(\sum_{i=0}^{n-1}\underline\beta_i(t_{i+1}-t_i)^3\right)^{1/3}
\left(\sum_{i=0}^{n-1}1\right)^{2/3}=n^{2/3}
\left(\sum_{i=0}^{n-1}\underline\beta_i(t_{i+1}-t_i)^3\right)^{1/3}.
\end{aligned}
\]
Cubing and rearranging gives
\[
\sum_{i=0}^{n-1}\underline\beta_i(t_{i+1}-t_i)^3
\ge\frac1{n^2}
\left(\sum_{i=0}^{n-1}\underline\beta_i^{1/3}(t_{i+1}-t_i)\right)^3.
\]
The function $\beta_{v,u}^{1/3}$ is continuous on the compact interval $I$. If $\omega$ is its modulus of continuity, then for every gap
\[
\begin{aligned}
0&\le\int_{t_i}^{t_{i+1}}\beta_{v,u}(t)^{1/3}\,\mathrm dt
-\underline\beta_i^{1/3}(t_{i+1}-t_i)
\le(t_{i+1}-t_i)\,\omega(\delta_N).
\end{aligned}
\]
Summing and using $\sum_i(t_{i+1}-t_i)=b-a$ gives
\[
\begin{aligned}
0&\le\int_a^b\beta_{v,u}(t)^{1/3}\,\mathrm dt
-\sum_{i=0}^{n-1}\underline\beta_i^{1/3}(t_{i+1}-t_i)
\le(b-a)\omega(\delta_N)\longrightarrow0.
\end{aligned}
\]
Since $n\le N+2$, combining the preceding inequalities with the near-minimax property yields
\[
\begin{aligned}
\mathfrak R_N(F;\mathcal X_{L,M})+N^{-3}
&\ge\frac{vM}{32(N+2)^2}
\times\left(
\sum_{i=0}^{n-1}\underline\beta_i^{1/3}(t_{i+1}-t_i)
\right)^3
-o(N^{-2}).
\end{aligned}
\]
Multiply by $N^2$ and take the lower limit. We have $N^2/(N+2)^2\to1$, $N^2N^{-3}\to0$, and the lower Riemann sums tend to $\int_I\beta_{v,u}(t)^{1/3}\,\mathrm dt$. Therefore
\[
\liminf_{N\to\infty}N^2\mathfrak R_N(F;\mathcal X_{L,M})
\ge\frac{vM}{32}
\left(\int_I\beta_{v,u}(t)^{1/3}\,\mathrm dt\right)^3,
\]
which is~\eqref{eq:minimax_acceleration_lower}. The right-hand side is strictly positive because $v,M,\beta_0$ and $b-a$ are positive. Together with the uniform-grid upper bound, it proves $\mathfrak R_N(F;\mathcal X_{L,M})=\Theta(N^{-2})$ and completes Theorem~\ref{thm:minimax_acceleration}.
\end{proof}

\section{Additional Regularity and Active Sensing Results}
\label{app:regularity_active}

\subsection{An outer certificate under bounded acceleration}
\label{app:regularity_certificate}

Fix the observed record $S$ and suppose the prescribed bounds $L,M>0$ are consistent with it, so $\mathcal X_{L,M}(S)\ne\varnothing$. Every path in this class satisfies both the speed-only area bound $A_i^\star$ and the acceleration bound in Proposition~\ref{prop:app_acc_area_bounds}. Define
\begin{equation}
\label{eq:app_acc_outer_area}
\begin{aligned}
\overline A_i=\min\Biggl\{&A_i^\star,
\frac{M\|x_{i+1}-x_i\|(t_{i+1}-t_i)^2}{12}
+\frac{M^2(t_{i+1}-t_i)^4}{80}\Biggr\},
\end{aligned}
\end{equation}
Using the target coefficients $c_i$ from Theorem~\ref{thm:partial_target_certificate}, define the acceleration-aware leading radius by
\[
R_{2,L,M}^{\mathrm{out}}(S)=\sum_{i=0}^{n-1}\overline A_i|c_i|.
\]

\begin{proposition}[Acceleration-aware outer response bound]
\label{prop:app_acc_outer_response}
Fix the observed record $S$ on $[0,T]$ and prescribed
bounds $L,M>0$ for which $\mathcal X_{L,M}(S)\ne\varnothing$.
Let $V_1,V_2\in C_b^4(\mathbb R^m;\mathbb R^m)$ and $g\in C^2(\mathbb R^m;\mathbb R)$ have bounded first two derivatives. Then every $X\in\mathcal X_{L,M}(S)$ satisfies
\begin{equation}
\label{eq:app_acc_outer_response}
|F(X)-F(\bar X)|\le R_{2,L,M}^{\mathrm{out}}(S)+C_Y(S_3+S_2^2).
\end{equation}
Here $C_Y$ is the scalar-remainder coefficient of Lemma~\ref{lem:app_scalar_expansion}, depending only on $V_1,V_2,g,L,T$. Thus $\mathcal Y_T(S;\mathcal X_{L,M})$ lies in the interval centered at $F(\bar X)$ with radius $R_{2,L,M}^{\mathrm{out}}(S)+C_Y(S_3+S_2^2)$, and its diameter is at most twice this radius. With the fixed sensitivity bound $C_{\mathrm{sens}}$ from Appendix~\ref{app:certificate_adjoint}, the leading radius satisfies
\[
R_{2,L,M}^{\mathrm{out}}(S)\le C_{\mathrm{sens}}\left(\frac{LMT}{12}\delta^2+\frac{M^2T}{80}\delta^3\right),\qquad \delta=\max_i(t_{i+1}-t_i).
\]
\end{proposition}

\begin{proof}
The pointwise expansion from Lemma~\ref{lem:app_scalar_expansion} gives, for each speed-admissible completion,
\[
\begin{aligned}
F(X)-F(\bar X)&=\sum_{i=0}^{n-1}c_i a_i(X)+\varepsilon(X),\\
|\varepsilon(X)|&\le C_Y(S_3+S_2^2).
\end{aligned}
\]
Every $X\in\mathcal X_{L,M}(S)$ is speed-admissible. Its area on the $i$th interval satisfies $|a_i(X)|\le A_i^\star$ by the Dido theorem and
\[
|a_i(X)|\le\frac{M\|x_{i+1}-x_i\|(t_{i+1}-t_i)^2}{12}+\frac{M^2(t_{i+1}-t_i)^4}{80}
\]
by Proposition~\ref{prop:app_acc_area_bounds}. Taking the smaller upper bound and applying the triangle inequality yields
\[
\begin{aligned}
|F(X)-F(\bar X)|&\le\sum_i|c_i|\,|a_i(X)|+|\varepsilon(X)|
\le\sum_i|c_i|\overline A_i+C_Y(S_3+S_2^2),
\end{aligned}
\]
which proves~\eqref{eq:app_acc_outer_response}.

The fixed bound $\sup_i|c_i|\le C_{\mathrm{sens}}$ follows from Appendix~\ref{app:certificate_adjoint}. Since $\|x_{i+1}-x_i\|\le L(t_{i+1}-t_i)$,
\[
\begin{aligned}
R_{2,L,M}^{\mathrm{out}}(S)&\le C_{\mathrm{sens}}\sum_i\overline A_i\\
&\le C_{\mathrm{sens}}\left[\frac{LM}{12}\sum_i(t_{i+1}-t_i)^3+\frac{M^2}{80}\sum_i(t_{i+1}-t_i)^4\right]\\
&\le C_{\mathrm{sens}}\left[\frac{LMT}{12}\delta^2+\frac{M^2T}{80}\delta^3\right].
\end{aligned}
\]
The final inequality uses $(t_{i+1}-t_i)^3\le\delta^2(t_{i+1}-t_i)$, $(t_{i+1}-t_i)^4\le\delta^3(t_{i+1}-t_i)$ and $\sum_i(t_{i+1}-t_i)=T$.
\end{proof}

The parabolic witness of Proposition~\ref{prop:app_acc_area_bounds} achieves a cubic local area scale when the speed constraint has sufficient slack. For globally $W^{2,\infty}$ paths, adjacent intervals share their velocities at observation times, so their maximal areas cannot generally be chosen independently. The remainder $S_3+S_2^2$ can also have order $\delta^2$, the order of the leading acceleration-aware sum. The bound therefore does not determine the sharp asymptotic coefficient of the nonlinear response diameter.

\subsection{Conditional area after an additional observation}
\label{app:midpoint_area}

In the bounded-speed class, consider an interval $[t_i,t_{i+1}]$ and fix an additional observation time $\tau\in(t_i,t_{i+1})$. If the new value is $x=X_\tau$, then the original signed area splits into a known triangle contribution and two still-unobserved areas.

\begin{lemma}[Conditional area fiber after one observation]
\label{lem:app_conditional_midpoint_area}
Let $x_i,x_{i+1}\in\mathbb R^2$ be the recorded endpoints, and let $x=X_\tau$ be an additional measurement at $\tau\in(t_i,t_{i+1})$. Suppose the prescribed speed bound is consistent with the three observed values, namely $\|x-x_i\|\le L(\tau-t_i)$ and $\|x_{i+1}-x\|\le L(t_{i+1}-\tau)$. The signed areas attained by paths matching these measurements and satisfying the speed bound form exactly an interval centered at $\tfrac12(x-x_i)\wedge(x_{i+1}-x_i)$, with radius
\[
A_\star(\|x-x_i\|,L(\tau-t_i))+A_\star(\|x_{i+1}-x\|,L(t_{i+1}-\tau)).
\]
\end{lemma}

\begin{proof}
Splitting the defining area integral at $\tau$ and substituting $X_t-x_i=(X_t-x)+(x-x_i)$ on the second interval gives
\[
\begin{aligned}
2a_i(X)&=\int_{t_i}^{\tau}(X_t-x_i)\wedge\mathrm dX_t+\int_{\tau}^{t_{i+1}}(X_t-x_i)\wedge\mathrm dX_t\\
&=\int_{t_i}^{\tau}(X_t-x_i)\wedge\mathrm dX_t+\int_{\tau}^{t_{i+1}}(X_t-x)\wedge\mathrm dX_t
+\int_{\tau}^{t_{i+1}}(x-x_i)\wedge\mathrm dX_t\\
&=2a_{[t_i,\tau]}(X)+2a_{[\tau,t_{i+1}]}(X)+(x-x_i)\wedge(x_{i+1}-x).
\end{aligned}
\]
Since $(x-x_i)\wedge(x-x_i)=0$, the last term equals $(x-x_i)\wedge(x_{i+1}-x_i)$. Dividing by two yields
\[
a_i(X)=\frac12(x-x_i)\wedge(x_{i+1}-x_i)+a_{[t_i,\tau]}(X)+a_{[\tau,t_{i+1}]}(X).
\]
The first term is fixed by the new observation. By Lemma~\ref{lem:app_dido_fiber}, each of the two remaining areas ranges over its full Dido interval. Their choices are independent, because absolutely continuous paths with a common intermediate value can be concatenated without matching their velocities. The sum of the two centered intervals is the full centered interval with radius equal to the sum of their radii.
\end{proof}

\subsection{One-step gap selection with frozen sensitivities}
\label{app:frozen_sensing}

Each candidate observation time is fixed at the midpoint $\tau_i=(t_i+t_{i+1})/2$. Before measurement, the prescribed speed bound restricts the possible value $X_{\tau_i}$ to the intersection of the disks centered at the recorded endpoints $x_i$ and $x_{i+1}$, each of radius $L(t_{i+1}-t_i)/2$. Every point of this intersection is attained by two straight segments with speed at most $L$. Lemma~\ref{lem:app_conditional_midpoint_area} therefore gives the worst-case residual area radius:
\begin{equation}
\label{eq:app_midpoint_residual}
\begin{aligned}
B_i^{\mathrm{mid}}
&=\max_{\substack{\|x-x_i\|\le L(t_{i+1}-t_i)/2\\\|x_{i+1}-x\|\le L(t_{i+1}-t_i)/2}}
\Bigl[\,A_\star(\|x-x_i\|,L(t_{i+1}-t_i)/2)+A_\star(\|x_{i+1}-x\|,L(t_{i+1}-t_i)/2)\,\Bigr].
\end{aligned}
\end{equation}
The maximum exists because the intersection of the two disks is compact and the Dido area function is continuous. For every fixed midpoint value, the conditional area fiber is a subset of the original Dido interval. Therefore $0\le B_i^{\mathrm{mid}}\le A_i^\star$.

Define $R_i=A_i^\star|c_i|$ and $G_i^{\mathrm{mid}}=|c_i|(A_i^\star-B_i^{\mathrm{mid}})$. The coefficient $c_i$ is computed from the current reference path and held fixed when comparing candidate midpoint measurements.

\begin{proposition}[Optimal gap for the fixed-midpoint surrogate]
\label{prop:app_frozen_gap_choice}
For the additive leading radius $R_2(S)=\sum_iR_i$, suppose that only the selected gap is split, its observation time is fixed at the midpoint, and all coefficients $c_i$ are kept unchanged during this one-step decision. Among the candidate gaps, the smallest worst-case residual surrogate radius is $R_2(S)-\max_iG_i^{\mathrm{mid}}$, attained by any choice
\begin{equation}
\label{eq:app_gar_exact_choice}
i^\star\in\operatorname*{arg\,max}_iG_i^{\mathrm{mid}}.
\end{equation}
This proposition concerns gap selection for a fixed observation time; it does not claim that a midpoint is optimal among all possible times, or that the chosen gap minimizes the fully recomputed nonlinear Neural CDE certificate.
\end{proposition}

\begin{proof}
For a candidate gap $i$, all other terms $R_j$ of the frozen surrogate are unchanged. By Lemma~\ref{lem:app_conditional_midpoint_area}, the selected gap contributes at most $|c_i|B_i^{\mathrm{mid}}$ to the residual radius, and the maximizing midpoint value in~\eqref{eq:app_midpoint_residual} attains this worst case. Hence
\[
\begin{aligned}
R_{\mathrm{after}}^{(i)}&=\sum_{j\ne i}R_j+|c_i|B_i^{\mathrm{mid}}
=R_2(S)-|c_i|A_i^\star+|c_i|B_i^{\mathrm{mid}}
=R_2(S)-G_i^{\mathrm{mid}}.
\end{aligned}
\]
The known triangle term in Lemma~\ref{lem:app_conditional_midpoint_area} shifts the conditional area interval, and hence the frozen scalar surrogate, without changing its radius. Since $R_2(S)$ is the same for all candidate gaps, minimizing the residual radius is equivalent to maximizing $G_i^{\mathrm{mid}}$.
\end{proof}

\subsection{Closed-form midpoint-reduction score}
\label{app:midpoint_score}

The acquisition policy uses a closed-form midpoint residual score in place of the maximization in~\eqref{eq:app_midpoint_residual}:
\[
B_i^{\mathrm{score}}=\begin{cases}
\dfrac{L^2(t_{i+1}-t_i)^2}{4\pi},&\dfrac{\|x_{i+1}-x_i\|}{L(t_{i+1}-t_i)}\le\dfrac2\pi,\\[6pt]
\dfrac12 A_i^\star,&\dfrac{\|x_{i+1}-x_i\|}{L(t_{i+1}-t_i)}>\dfrac2\pi,
\end{cases}
\]
The policy ranks gaps by $G_i^{\mathrm{score}}=|c_i|\max\{A_i^\star-B_i^{\mathrm{score}},0\}$. This score uses $B_i^{\mathrm{score}}$, whereas Proposition~\ref{prop:app_frozen_gap_choice} establishes optimal gap selection for the exact residual $B_i^{\mathrm{mid}}$. In Section~\ref{subsec:exp_acquisition}, the full certificate is recomputed after each midpoint measurement.

\section{Experimental Setup and Reproducibility}
\label{app:experimental_protocol}

\subsection{Signals, targets, and observation model}
\label{app:exp_data}

In the certificate, regularity, and sequential-acquisition experiments, the planar drivers on $[0,1]$ have four Fourier harmonics:
\[
X_t=t\,d+\sum_{k=1}^{4}\bigl(a_k\sin(2\pi kt)+b_k\cos(2\pi kt)\bigr),
\]
where $d,a_k,b_k\in\mathbb R^2$. The sine and cosine coefficients are independently Gaussian before frequency-dependent scaling; the drift is sampled separately. For the speed-only certificate and acquisition datasets, the harmonic amplitudes decay as $k^{-2}$, the complete path is normalized to maximum speed approximately one, and $L=1.08$. For the acceleration dataset, amplitudes decay as $k^{-3}$, the oscillatory part is scaled to prescribed speed and acceleration targets of $0.62$ and $1.55$, and a separate drift is added before any necessary final speed rescaling. This dataset uses $L=1$ and $M=4$.

The two path-dependent targets are the signed area of the open curve closed by its endpoint chord and the readout of a fixed three-dimensional nonlinear controlled system. For the latter, the initial state is $(0.05,-0.03,0.02)^\top$, the final readout is $z_3+0.35z_2-0.12z_1$, and the vector fields are
\[
\begin{aligned}
V_1(z)&=\begin{pmatrix}1+0.10\tanh(z_2)\\0.18\tanh(z_3)\\0.25\tanh(z_2)\end{pmatrix},\quad
V_2(z)=\begin{pmatrix}0.08z_2\\1+0.50\tanh(z_1)\\0.62z_1+0.12\sin(z_3)\end{pmatrix}.
\end{aligned}
\]
As a negative control, the endpoint-dependent target is $\sin(1.2(X_T^1-X_0^1))+0.65(X_T^2-X_0^2)+0.15\|X_T-X_0\|^2$. The signed-area target is evaluated on a fine polygonal representation; the nonlinear physical-system target uses one fourth-order Runge--Kutta step on each consecutive segment of that representation. The certificate dataset uses 257 evaluation points, while the acceleration and sequential-acquisition datasets use 513. The principal learned models have two spatial control channels, matching the planar certificate.

For each data seed (7, 19, and 41), each principal dataset contains 360 training, 96 validation, and 96 test paths. The certificate and regularity experiments evaluate all three targets with 5, 9, 17, or 33 equally spaced observations, nested within the stored 33-point input. The acquisition experiment evaluates the two path-dependent targets. Learned outputs are centered and scaled by each dataset's training-target mean and standard deviation.

\subsection{Learned controlled dynamics and numerical evaluation}
\label{app:exp_models}

For the certificate, regularity, and acquisition experiments, the regressor has hidden dimension 14, an input-dependent initialization network of widths $2\to14\to14$, a two-hidden-layer vector-field MLP of width 40 with $\tanh$ activations, and a scalar linear readout. The vector-field network outputs a $14\times2$ matrix at each hidden state.

During training, the 33-point input is integrated with an explicit trapezoidal/Heun update for each observed increment. Training minimizes mean squared error in standardized targets using AdamW ($2\times10^{-3}$ initial learning rate, weight decay $10^{-5}$), gradient clipping at norm 5, and a reduce-on-plateau learning-rate schedule (factor $0.5$, patience 5). Training uses at most 90 epochs, with validation-based early stopping (patience 14); batch size is 64 in the certificate and acquisition experiments and 256 in the acceleration experiment. One model is trained for each task and data seed, then frozen across all sparse resolutions.

For certificate evaluation, straight reference segments are integrated using eight numerical substeps per observation gap. The learned bracket $[V_1,V_2]=DV_2V_1-DV_1V_2$ is computed with Jacobian-vector products; the target coefficients $c_i$ are obtained by a backward adjoint sweep, differentiating each reference gap map with respect to its initial hidden state. As the readout is linear, its terminal gradient is constant.

\subsection{Circular-arc completions and certificate discrepancy}
\label{app:exp_arc_witness}

For the bounded-speed certificate comparison, the reference curve is the piecewise straight interpolation of the observed values. On each gap, we calculate the exact Dido maximal area $A_i^\star$ from the displacement and length budget $L(t_{i+1}-t_i)$. We then construct circular arcs with signed areas $+A_i^\star$ and $-A_i^\star$. The two global paths use the signs $\operatorname{sign}(c_i)$ and $-\operatorname{sign}(c_i)$ on each interval and therefore share the same sampled values. Each arc is evaluated with 12 subdivisions when propagated through the learned CDE. The observed nonlinear spread between these two paths is $D_{\mathrm{arc}}$, and the predicted leading diameter is $D_{\mathrm{cert}}=2R_2(S)$.

The relative discrepancy reported in Section~\ref{subsec:exp_certificate} is computed separately for each test case as
\[
\frac{|D_{\mathrm{arc}}-D_{\mathrm{cert}}|}{\max\{D_{\mathrm{arc}},D_{\mathrm{cert}},10^{-8}\}}.
\]
The certificate comparison comprises three seeds, three tasks, four held-out trajectories per seed, and four sampling resolutions: 144 cases in total, of which 96 belong to the two path-dependent tasks used for the reported rank correlation. For five- and nine-observation skeletons, a coordinate-wise search also explores alternative maximal-arc sign combinations within the same witness family.

Scaling exponents are ordinary least-squares slopes of the logarithm of the median metric at each sampling resolution against $\log\delta$, where $\delta=1/(n_{\mathrm{obs}}-1)$. In particular, the fit to four aggregated resolutions is a finite-resolution diagnostic rather than an independent proof of the asymptotic exponent.

\subsection{Acceleration-constrained completion search}
\label{app:exp_accel}

For smooth inputs, we compare $2R_2(S)$ with the acceleration-aware leading outer approximation $2R_{\mathrm{acc}}^{\mathrm{out}}(S)$, which replaces $A_i^\star$ in the sum by
\[
\min\!\left\{A_i^\star,\frac{M\|x_{i+1}-x_i\|(t_{i+1}-t_i)^2}{12}+\frac{M^2(t_{i+1}-t_i)^4}{80}\right\}.
\]
The coefficient $c_i$ is computed from the same frozen learned CDE and reference path for both quantities. This comparison uses the leading area terms, without adding the remainder $C_Y(S_3+S_2^2)$ from Proposition~\ref{prop:app_acc_outer_response}.

The completion search keeps the observed positions $x_i$ fixed and jointly optimizes knot velocities. Each gap consists of two constant-acceleration halves whose accelerations are determined by the recorded endpoint positions and shared endpoint velocities. The resulting paths are globally $C^1$.

In this family, maximum speed occurs at a knot or midpoint, and maximum acceleration is the norm of one of the constant half-gap accelerations. The search checks these quantities with relative numerical feasibility tolerance $2\times10^{-5}$ and retains only passing iterates when computing the maximum-minus-minimum output spread $D_{\mathrm{adv}}$. It uses 16 parallel restarts, up to 160 gradient steps, and four integration subdivisions per half-interval.

The evaluation covers three held-out trajectories per seed, three tasks, and four resolutions (108 cases, including 72 path-dependent cases). The reported spread is the largest value found by this search, with the stated numerical feasibility tolerance and integration error; it is not a certified continuous-class diameter.

\subsection{Sequential observation design}
\label{app:exp_acquisition}

The acquisition experiment starts with observations at either $(0,0.25,0.50,0.75,1)$ or $(0,0.10,0.31,0.62,1)$. Each policy selects a gap and measures the held-out physical path at its midpoint. The seven selection rules are: largest local radius $A_i^\star|c_i|$; midpoint reduction $|c_i|\max\{A_i^\star-B_i^{\mathrm{score}},0\}$ with frozen sensitivities (Appendix~\ref{app:midpoint_score}); longest gap; largest Dido area budget; largest $|c_i|$; random choice; and a one-step look-ahead oracle.

The oracle evaluates every candidate true midpoint and chooses the smallest recomputed certificate radius. The midpoint-reduction rule uses $B_i^{\mathrm{score}}$, without solving the exact maximization for $B_i^{\mathrm{mid}}$ in Proposition~\ref{prop:app_frozen_gap_choice}. The oracle has access to unobserved midpoint values and minimizes the next radius.

Each policy makes six acquisitions and recomputes the reference trajectory, target sensitivities and full radius after every measurement. The evaluation records the radius $R_2$, the nonlinear circular-arc witness spread, prediction error relative to the dense learned-model output, and error relative to the synthetic target. The reported normalized certificate-radius area under the six acquisitions is
\[
\frac{1}{R_2(S_0)}\sum_{k=0}^{5}\frac{R_2(S_k)+R_2(S_{k+1})}{2},
\]
where $S_k$ denotes the observation record after $k$ acquisitions. Smaller values indicate faster reduction of the leading ambiguity radius at equal measurement budget.

The evaluation has three seeds, two path-dependent tasks, two held-out trajectories per seed, and two starting designs, giving 24 initial states under each policy. The random baseline uses one seeded realization per initial state rather than an average over possible random acquisition sequences.

\subsection{Admissibility checks for numerical completions}
\label{app:exp_scope}

Circular arcs satisfy the continuous speed constraint analytically, although their CDE outputs are evaluated on numerical subdivisions. The piecewise-constant-acceleration completions are checked against the speed and acceleration limits with the numerical tolerance stated above. In the supplementary spline-and-bump searches in Appendix~\ref{app:additional_experiments}, derivative constraints are checked only on finite evaluation grids. All three procedures evaluate selected completion families rather than the full identified set.

\section{Additional Experimental Results}
\label{app:additional_experiments}

\subsection{Certificate accuracy across observation resolutions}
\label{app:exp_certificate_table}

Table~\ref{tab:exp_certificate_resolution} gives certificate diameters and circular-arc-pair spreads at each observation count, pooled over two path-dependent tasks, three data seeds and four held-out trajectories per seed.

\begin{table}[h]
\centering
\caption{Predicted diameter and feasible circular-arc spread (target SD), with per-case relative discrepancy.}
\label{tab:exp_certificate_resolution}
\begin{tabular}{rrrr}
\toprule
Observations & $2R_2(S)$ & $D_{\mathrm{arc}}$ & Median rel. error \\
\midrule
5  & 5.775 & 5.401 & 5.49\% \\
9  & 2.769 & 2.684 & 2.55\% \\
17 & 1.358 & 1.334 & 1.79\% \\
33 & 0.678 & 0.668 & 1.69\% \\
\bottomrule
\end{tabular}
\end{table}

The pooled median ratio $D_{\mathrm{arc}}/(2R_2)$ is $0.981$, and the Spearman rank correlation between the certificate and the arc-pair spread is $0.9987$ on the path-dependent targets. The generating target functional evaluated on the same arc pairs has Spearman correlation $0.9907$ with the learned-model certificate.

The two learned path-dependent regressors have an average dense-resolution test $R^2$ of $0.9990$. The endpoint-dependent negative control has a median leading certificate diameter $0.00070$ times that of the path-dependent regressors. Across the four resolutions, the fitted log--log slopes are $1.005$ for the arc-pair spread, $1.030$ for the predicted diameter, and $1.554$ for their absolute discrepancy.

\subsection{Effect and conservatism of acceleration regularity}
\label{app:exp_accel_table}

The acceleration-aware leading diameter is smaller than the speed-only leading diameter at every observation count in Table~\ref{tab:exp_accel_resolution}. Entries are medians over the two path-dependent targets, three data seeds and three held-out trajectories per seed.

\begin{table}[h]
\centering
\caption{Median leading geometric diameters and numerically screened completion spread (target SD).}
\label{tab:exp_accel_resolution}
\begin{tabular}{rrrr}
\toprule
Observations & Speed only & Acceleration aware & Searched spread \\
\midrule
5  & 1.7844 & 0.4591 & 0.09966 \\
9  & 0.8547 & 0.0869 & 0.00935 \\
17 & 0.4174 & 0.0183 & 0.00090 \\
33 & 0.2061 & 0.0042 & 0.00010 \\
\bottomrule
\end{tabular}
\end{table}

The pooled median acceleration-aware-to-speed-only diameter ratio is $0.0711$. The adversarial search yields a median ratio $D_{\mathrm{adv}}/(2R_{\mathrm{acc}}^{\mathrm{out}})=0.0911$; the Spearman rank correlation between these two quantities is $0.9748$.

Across the four median resolutions, fitted log--log slopes are $1.038$ for the speed-only diameter, $2.261$ for the acceleration-aware leading diameter, and $3.347$ for the searched adversarial spread. The last value describes the restricted optimization family and is not an estimate of the minimax exponent for all admissible drivers.

The dense-resolution learned-model $R^2$ is $0.985$ for the nonlinear CDE target and $0.518$ for signed area.

\subsection{Acquisition policies and remaining ambiguity}
\label{app:exp_acquisition_table}

The certificate-radius AUCs in Table~\ref{tab:exp_acquisition_table} are calculated from 24 paired acquisition trajectories across two path-dependent targets and two initial observation designs. The oracle chooses the midpoint giving the smallest next radius after measurement; it does not optimize the entire six-step sequence.

\begin{table}[h]
\centering
\caption{Median normalized certificate-radius area under six acquisitions (lower is better), and median final radius relative to its initial value.}
\label{tab:exp_acquisition_table}
\begin{tabular}{lrr}
\toprule
Policy & Radius AUC & Final / initial \\
\midrule
One-step look-ahead oracle & 3.626 & 0.391 \\
Largest local ambiguity & 3.633 & 0.387 \\
Predicted midpoint reduction & 3.633 & 0.387 \\
Area budget only & 3.749 & 0.401 \\
Longest gap & 3.823 & 0.400 \\
Random gap & 4.544 & 0.603 \\
Sensitivity only & 4.744 & 0.747 \\
\bottomrule
\end{tabular}
\end{table}

Largest-local-ambiguity and predicted-midpoint-reduction select different initial gaps in four of the 24 paired states. On those four states, the midpoint-reduction policy achieves the smaller fully recomputed radius once. The one-step oracle selects the same first gap as largest-local-ambiguity in 23 of 24 states.

The oracle minimizes the next recomputed radius for the observation record reached by its own policy. Median final-to-initial ratios aggregate different sequential histories, so its median need not be lower than that of every alternative policy.

\subsection{Commutativity as a structural control}
\label{app:exp_controls}

For the commutativity control, the inputs have three Fourier harmonics and irregularly jittered five- or nine-point observation times. At each resolution and for each seed (7, 19, 41), we evaluate ten random noncommuting three-dimensional CDE systems and two commuting constant-field systems without training. The noncommuting fields are $V_j(z)=c_j+A_j\tanh(B_jz+b_j)$ with five hidden units, Gaussian coefficients and nonlinearity strength drawn log-uniformly between $0.08$ and $1$; the commuting fields are constant.

Candidate completions use the natural cubic spline through the recorded samples plus local smooth bumps that vanish at the observation times. For each case, speed and acceleration budgets are set to $1.08$ and $1.12$ times the larger sampled derivative norm of the generating path and baseline spline, respectively. Feasibility is tested on a finite grid. The search uses three smooth bump modes per gap, 12 evaluation subdivisions per gap, and a finite population of candidates.

The output spread is normalized by the maximum of baseline-state path length, terminal-state displacement, and $0.10$.

The median normalized searched spread is $4.28\times10^{-4}$ for the noncommuting systems and $8.55\times10^{-16}$ for the commuting constant-field controls. A leave-one-seed-out regression of searched spread gives $R^2=0.049$ with sampling-only features and $R^2=0.836$ with an analytic sampling-by-bracket proxy. The grid-based derivative checks do not certify admissibility over the full continuous intervals.

\subsection{Response ambiguity in independently trained path-sensitive models}
\label{app:exp_learned_ambiguity}

The independently trained models use four-harmonic planar paths stored at 33 time points, with 320 training, 96 validation, and 96 test trajectories per seed. Each model has hidden dimension 12, vector-field network width 32, and a three-channel control containing time and two spatial coordinates. Training runs for at most 80 epochs with batch size 64 and validation-based early stopping (patience 12). Models are trained on either 33 observations or a nested nine-point subset, then evaluated on five- and nine-point skeletons from the same held-out trajectories.

The targets are signed area, a nonlinear CDE response, and an endpoint-dependent functional. Their labels are computed from the stored 33-point paths instead of the 257-point paths used in the principal experiment. The nonlinear target has the initial state and linear readout from Appendix~\ref{app:exp_data}, with vector fields
\[
\begin{aligned}
V_1(z)&=\begin{pmatrix}1+0.08\tanh(z_2)\\0.18\tanh(z_3)\\0.22\tanh(z_2)\end{pmatrix},\quad
V_2(z)=\begin{pmatrix}0.06z_2\\1+0.55\tanh(z_1)\\0.65z_1+0.12\sin(z_3)\end{pmatrix}.
\end{aligned}
\]
The endpoint target uses coefficients $1.3$, $0.7$, and $0.20$ in the corresponding expression in Appendix~\ref{app:exp_data}.

The adversarial completions use natural cubic splines through the recorded samples with compactly supported smooth basis perturbations that vanish at the observation times. Speed and acceleration are checked at finitely many points. The search maximizes the learned output difference using three restarts, 65 gradient steps, three bump modes per gap, and eight discretization subdivisions per gap.

Table~\ref{tab:exp_learned_ambiguity} gives output spreads for the models trained on 33 observations and the generating targets evaluated on the same selected completion pairs.

\begin{table}[h]
\centering
\caption{Median learned-model spread and true-target spread on the same selected completion pairs, in training-target standard deviations.}
\label{tab:exp_learned_ambiguity}
\begin{tabular}{lrrrr}
\toprule
& \multicolumn{2}{c}{5 observations} & \multicolumn{2}{c}{9 observations} \\
Target & Model & Target & Model & Target \\
\midrule
Signed area & 0.2712 & 0.2903 & 0.0686 & 0.0705 \\
Nonlinear CDE & 0.2034 & 0.2154 & 0.0517 & 0.0557 \\
Endpoint only & 0.0058 & 0.0000 & 0.0012 & 0.0000 \\
\bottomrule
\end{tabular}
\end{table}

The five-to-nine comparison pairs observations from the same held-out trajectory. The endpoint target is unchanged by any path perturbation that fixes the endpoints, whereas the trained model has a small residual output spread.

For the dense-trained path-sensitive models, leave-one-seed-out regression of searched output spread gives cross-validated $R^2=0.687$ with sampling-only features and $R^2=0.845$ with a sampling-by-bracket indicator.

\end{document}